\documentstyle[11pt,covingtn,epsf]{report}
\newtheorem{sent}{}
\newcommand{\numberSentence}[1]{\begin{sent} #1 \end{sent}}
\newcommand{\labelRule}[2]{\begin{center}\begin{tabular}{ll} #1 & (#2) \end{tabular} \end{center}}
\newcommand{\lexical}[2]{\begin{tabular}{l} #1 $\mapsto$ #2 \end{tabular}}

\title{Learning Unification-Based Natural Language  Grammars}

\author{Miles Osborne \\ 
Submitted  for the degree of Doctor of Philosophy \\ 
The Intelligent Systems Group,\\
The Department of Computer Science \\
The University of York.}

\date{September 1994}

\begin{document}
\bibliographystyle{plain}


\maketitle
\pagenumbering{roman}      
\newpage

\begin{quotation}
`It is high time we turned to Grammar now,' said Doctor Cornelius in a loud
voice.  `Will your Royal Highness be pleased to open Pulverulentus Siccus
at the fourth page of his {\it Grammatical Garden or the Arbour of 
Accidence pleasantlie open'd to Tender Wits?}' \\
After that it was all nouns and verbs till lunchtime, but I don't think
Caspian learned much. \\
C. S. Lewis, {\it Prince Caspian}.
\end{quotation}

\newpage
\begin{abstract}
Practical text processing systems need wide covering grammars.  When parsing
unrestricted language, such grammars often fail to generate all of the
sentences that humans would judge to be grammatical.  
This problem undermines successful parsing of the text 
and  is known as {\it undergeneration}.  
There are two main ways of dealing with undergeneration: either by
sentence correction, or by grammar correction.  This thesis concentrates
upon automatic grammar correction (or  machine learning of grammar) 
as a  solution to the problem of undergeneration.  Broadly speaking, grammar
correction approaches can be classified as being either 
 {\it data-driven}, or
{\it model-based}.  Data-driven learners use data-intensive methods
to acquire grammar.  They  typically use 
grammar formalisms unsuited to the needs of practical text processing and
cannot guarantee that the resulting grammar is adequate for subsequent 
semantic interpretation.  That is, data-driven learners acquire grammars
that generate strings that humans would judge to be grammatically ill-formed
(they {\it overgenerate})
and fail to assign linguistically plausible parses.  Model-based learners
are knowledge-intensive and 
are reliant for success upon the completeness of a {\it model of 
grammaticality}.  But in practice, the  model will be incomplete.  Given that
in this thesis
we deal with undergeneration by learning, we hypothesise that the combined
use of data-driven and model-based learning would allow data-driven learning to
compensate for model-based learning's incompleteness, whilst model-based 
learning would compensate for data-driven learning's unsoundness.  We 
describe a system that we have used to test the hypothesis empirically.
The system combines data-driven  and model-based learning to
acquire unification-based grammars that are more suitable for practical
text parsing.  Using the Spoken English Corpus as data, and by quantitatively 
measuring undergeneration, overgeneration and parse plausibility, we show
that this hypothesis is correct. 

\end{abstract}           
\tableofcontents
\listoffigures
\newpage
\section*{Acknowledgements}
 I would like to thank Derek Bridge for supervising my
work.  Without his attention to detail, this thesis would not
exist.  In particular, Derek has helped me on the long road to
clearer writing.  

Thanks  to  Eric Atwell for supplying the
Spoken English Corpus and for being a sparring partner, 
Ted Briscoe for supplying the SEC Grammar, and 
John Carroll for technical help with
the GDE. 

I wish to thank SERC for funding,  and this
Department for providing computing resources and for 
allowing me to go to various conferences.  

On a less serious note, a hearty roar to the members of the Intelligent
Systems Group at York:
Robert Dormer, Alan Frisch, Tony Griffiths, and Craig MacNish.  How can
I sum up three years of coffee, loud tapes, chocolate bars, sympathy,
and shouting?  Beats me.  

Hmm, this is becoming a long cast list.  I'll stop by thanking 
Sr.\ Amadeus Bulger, IBVM, for helping me in times of love
and grief, and finally Amy Davies.  

Good God!

\newpage
\section*{Declarations}
I hereby declare that my thesis entitled ``Learning Unification-Based Natural Language  Grammars'' is not the same as that of any that I have 
submitted for a degree, diploma, or other qualification at any other
University or similar institution.

I further declare that no part of my dissertation has already been or is
being concurrently submitted for any degree, diploma, or other qualification.

I further state that this research and its results are for the most part
my own.  Where I have adopted theories or results of others I state this
clearly in the text and the bibliography.  

Part of this research has been published already \cite{Osbo93a,Osbo93b,Osbo94a,Osbo94b}.

I further confirm that my thesis does not exceed 100,000 words in length.        
\newpage


\pagenumbering{arabic}     
\pagestyle{headings}       

\chapter{Introduction}

\section{Introduction}

The research described in this thesis is an attempt to solve a problem
facing practical {\it text processing systems}, the problem 
of  {\it undergeneration}.  
In the approach taken, 
undergeneration is tackled by {\it machine learning} of {\it grammar}.

Text processing systems typically 
  consist 
of a   {\it grammar} describing the syntax of a natural language, a
 {\it lexicon}
describing lexical information about words, 
a semantic component that is used to construct the meaning of  sentences, and
a pragmatic component which is used to construct the non-literal meaning of
sentences. They also contains  a {\it parser}, which is a program that  
produces phrases structure 
trees  for sentences in the language defined by the grammar.  Text processing
systems can be used in a variety of tasks, such as machine translation, 
text summarising, natural language interfaces to databases, and so on.
An example text processing system is the Grammar Development 
Environment (GDE) \cite{Carr91}.

Allowing practical processing of naturally occurring language places certain
 demands upon the grammar.  Perhaps the most important   of these
demands
is that the grammar is  of wide-coverage. That is, 
the grammar should generate all of those grammatical sentences that
the text processing system encounters.  
Unfortunately, no manually-constructed natural language grammar
can generate all of the sentences that humans would judge to
be grammatical.  The GDE is distributed with
one of the largest grammars of a natural language (in this case
English) and yet fails to generate the uncontroversially well-formed 
noun phrase (NP)  {\it all abbeys and an abbot}, as shown by the following
trace of the GDE:
\begin{verbatim}
69 Parse>> all abbeys and an abbot
28320 msec CPU, 30000 msec elapsed
381 edges generated
No parses
\end{verbatim}
This is not due to a lack of lexical information as shown by successful
parsing of the NPs {\it all abbeys and all abbots} and {\it an abbot and
abbey}:
\begin{verbatim}
68 Parse>> all abbeys and all abbots
50010 msec CPU, 59000 msec elapsed
547 edges generated
1 parse
67 Parse>> an abbey and an abbot
15410 msec CPU, 16000 msec elapsed
238 edges generated
1 parse
\end{verbatim}
This failure to generate (in this case) a NP is an instance of
 {\it undergeneration}.
Undergeneration is a serious problem for natural language processing
(NLP) and its treatment 
is necessary before
NLP can be successful.  This thesis focuses on overcoming the 
problem of undergeneration, 
whilst trying to ensure that the learnt grammar meets the demands made by
the host text processing system.  

In the next section,  undergeneration is  defined more carefully.  The 
causes of
undergeneration are presented, along with criteria for its successful
treatment.  After this follows a discussion of two approaches dealing
with undergeneration (correcting the string,  or correcting the grammar)
which will give an indication of how the problem can be tackled.  This
leads to an overview of the approach presented in this thesis.  Finally,
a description of what is to come ends this chapter.

\section{Undergeneration}

\subsection{Definition \label{ugDef}}
\begin{figure}
\centering
\leavevmode
\epsffile{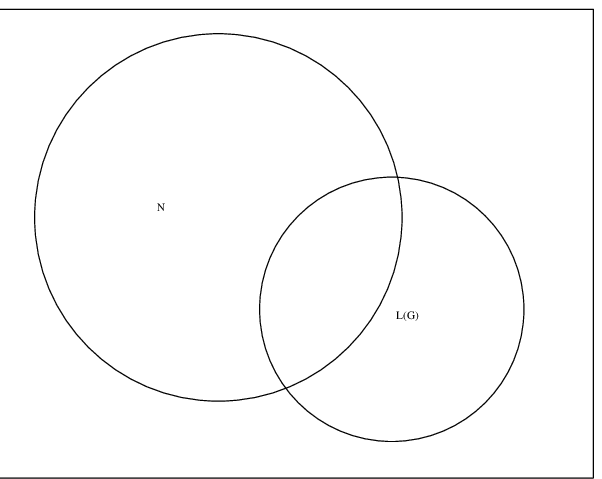}
\caption{\label{venn1} $\cal N$ and $L(G)$}
\end{figure}

Let $\cal N$ be the set of sentences of some natural language.  By
`natural'  is meant any string of words that a human would judge to
be grammatical or acceptable.  An acceptable sentence is one that is
comprehensible, even if it is not grammatically well-formed.  For example,
the sentence:\footnote{Ungrammatical sentences are conventionally marked with
an asterisk.}
\numberSentence{*Sam chase the ball}
is clearly ungrammatical (there is a subject-verb disagreement), but
acceptable (it is obvious what is meant).  Let $G$ be a natural language
grammar that we are supplying.  The 
language generated by $G$ relates to  $\cal N$ as shown 
in figure \ref{venn1}.

\begin{figure}
\centering
\leavevmode
\epsffile{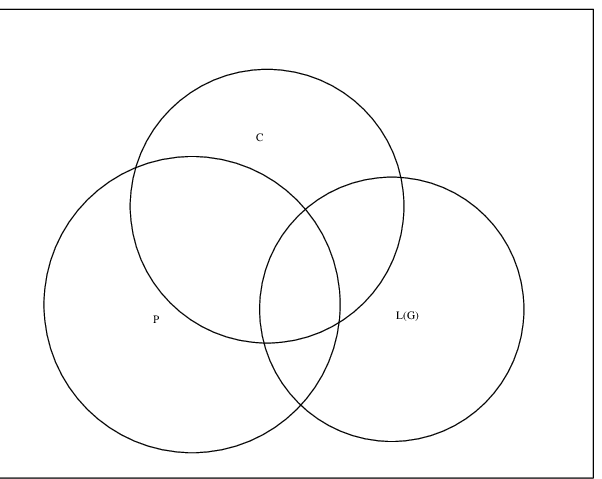}
\caption{\label{venn2} $\cal N$, $C$, $P$ and $L(G)$}
\end{figure}

Here, the universe consists of all possible strings that can be
formed from $T^*$.\footnote{$T^*$ is taken to mean the kleene 
closure of the set of words in question.}  A first attempt at defining undergeneration would
be  the set of sentences in $\cal{N} -L(G)$\footnote{The notation
$L(G)$ means the language generated 
by the grammar $G$.} being not empty.  This treats
all sentences in $\cal N$ equally (they should all be in the language 
generated by the grammar).  Linguists however do not treat natural languages
as a monolithic set of sentences.  Traditionally, language is demarcated in
terms of {\it competence} and {\it performance}.  Competence describes
human knowledge of language and is independent of processes such as 
memory.  Given 
these two types of sentence, the previous
Venn diagram can be refined as shown in figure \ref{venn2}.
Here, $C$ is the set of sentences that a human would judge to be
grammatical, but possibly unacceptable,  and
$P$ is the set of sentences or strings that humans would judge to be
either grammatical or ungrammatical, but still acceptable.  The previous
set of sentences $\cal N$ is now the union  of $C$ and $P$.  There are many
interesting regions in this diagram.  For example, sentences in $C$ but not
in $P$ or $L(G)$, such as :
\numberSentence{The war the general the president appoints starts ends the world} 
 will be grammatical, but unacceptable.\footnote{This center-embedded 
example is taken  from Johnson-Laird \cite[p.271]{John83}.}
Such sentences are  difficult to process by humans.
Any sentence not in $C$ is ungrammatical and a grammar generating these
is said to {\it overgenerate}.   Defining undergeneration now must take
into account which region in the Venn diagram $G$ should expand into.  This 
is controversial.  For example, Atwell considers undergeneration
to be the existence of 
 sentences in $P$ but not in $L(G)$ (personal communication).  Instead of 
Atwell's definition, undergeneration is 
defined as being the existence of  sentences in $C$ but not
in $L(G)$.    We believe that the task of a grammar is to
generate sentences that, as judged by humans, are  `grammatical', and 
leave ungrammatical
strings to a sentence correction approach.  Our definition of undergeneration
means that a grammar is a theory of competence, and not, if using Atwell's 
definition, a theory of whatever
strings or sentences the system encounters (or performance).  From now 
on, we shall refer to
grammars that try to only generate sentences in  $C$ as a {\it competence 
grammar} and grammars that try to generate sentences in $P$ as a 
{\it performance grammar} \cite[p.1]{Sout92}. 

\subsection{Causes}

The natural language grammars  used in NLP systems undergenerate for two
reasons:
\begin{center}
\begin{itemize}
\item Theoretical.  Not all researchers believe that a competence grammar
can be constructed, even in theory \cite{Samp87}.
Sampson syntactically analysed approximately 10,000 noun phrases 
(NPs) \cite{Samp87}
taken from the Lancaster-Oslo-Bergen (LOB) Corpus \cite{Joha78}.  He 
expected to find
most NPs to be of a few  types, and the remainder, corresponding to
ungrammatical or idiosyncratic NPs, to be of types with far fewer members.
Distributions of this kind are known as being {\it Zipfian} and are common 
of other aspects of language \cite{Zipf36}.  For example, function words (``the'', ``a'')
are
far more frequent than other words, whilst  words such as ``travail'' 
appear only once (for example) in this dissertation.  Sampson 
did not find a Zipfian
distribution and concluded that as no such demarcation was found 
in his experiment between grammaticality and ungrammaticality, no demarcating grammar would be able to
account for all of a natural language.  
 Sampson's conclusion, if correct, 
profoundly undermines competence  grammar construction and so 
Taylor {\it et al}
re-constructed Sampson's argument using a wide-covering 
grammar containing rules of greater generality than those used by 
Sampson \cite{Bris87,Tayl89}.  As Church notes, they
 found a Zipfian distribution,
 thereby suggesting that Sampson
failed to find an appropriate grammar \cite{Chur91}.   Hence, Taylor {\it et al's}
experiment weakens Sampson's argument that a competence grammar can never
be constructed that will not undergenerate. It does not totally refute
the argument as Taylor {\it et al's} 
 wide-covering grammar did still  undergenerate.

Assuming competence grammars can be constructed, another 
theoretical cause of undergeneration  is language change.  That is, languages 
evolve over time, and eventually, new constructs will emerge that a
NLP systems's grammar, being static, will be 
unable to generate.  

\item Logistical.  Natural language grammars are large: the descriptive
grammar of Quirk {\it et al.\ }is over 1500 pages in length \cite{Quirk85},
the Alvey Tools Grammar \cite{Grov92} contains just under a thousand 
generalised rules\footnote{These rules are {\it unification-based}.  See 
\S \ref{formalism} for 
a description of unification-based grammars.}, the CLARE grammar of the 
CORE Language Engine \cite{Alsh92} and the IBM Grammar \cite{Shar88,Blac93}
contain a similar number of rules to the Alvey Tools Grammar. 
Natural language grammars thus
represent a major investment of skilled labour.  Some researchers
go as far as saying that, in practice, the task of constructing
a competence grammar is so large as to be
 impractical \cite{Sout89,Sout90,Sout92}.
  Atwell, 
O'Donoghue and 
Souter comment that \cite[p.3]{Sout91}:
\begin{quote}
Since the syntax of a natural language such as English is extremely
complex, large corpora of texts will continue to throw up sentences which
are not dealt with adequately by current generative grammars.
\end{quote}
Similarly, Souter and O'Donoghue conjecture \cite[p.2]{Sout90}:
\begin{quote}
We are, then, somewhat cautious as to the ultimate value of manually building
a large rule-based grammar, as there will always be some new sentences which
contain structures not catered for in the grammar, so any parser using such
a grammar will hardly be robust.
\end{quote}
That is, their conclusion is that the
logistical problems involved with manually constructing a
wide covering grammar will be so daunting that grammars will always 
undergenerate.  As an example  
Taylor {\it et al's} \cite{Tayl89} use of the Alvey Natural 
Language Tools (ANLT) Grammar \cite{Grov92} 
still   failed to account for $3.12\%$ of the 10,000 NPs found in the LOB 
Corpus.  It is difficult to refute the logistical argument, given the fact 
that, to the author's knowledge, all manually constructed wide-covering
competence grammars constructed to date undergenerate.  
\end{itemize}
\end{center}
The theoretical arguments are controversial. If they do not hold, then 
there is no {\it inherent} reason why the logistical arguments cannot be
overcome by devoting even greater resources to grammar development 
projects.  However, the logistical problems still need to be solved before
wide covering grammars can be constructed.  This thesis considers how
these problems might be overcome by automated support of the grammar
engineering process.  
\subsection{Criteria for successful treatment of undergeneration \label{criteria}}
An approach successfully dealing with a grammar $G$'s undergeneration 
should extend $G$, giving the grammar $G'$, such that $G'$ satisfies
the following criteria:
\begin{center}
\begin{enumerate}
\item Ideally, $L(G') = C$.  In practice, $G$ may generate sentences not
in $C$, so we have $L(G') \cap C \supseteq L(G) \cap C $.
\item Ideally, $L(G') \cap  \overline{C}  = \emptyset$.  However, 
increasing the coverage of $G$ might lead to overgeneration.  In practice,
$(L(G') - C) \subseteq  (L(G) - C)$
\item  Ideally, $G'$ should assign the `correct' set of parses to those
sentences that it does generate.  In practice, $G'$ may overgenerate and
so the best that that we can hope for is that 
 $G'$ should assign {\it plausible} parses to those sentences that it
generates.
\item $G'$ should be in a form suitable for NLP.
\end{enumerate}
\end{center}
$C$ is the set of sentences generated by a competence grammar mentioned
in \S \ref{ugDef}.

The first criterion follows from the definition of undergeneration
introduced in the previous section.  That is, the extended grammar
should generate all the sentences in $C$, along with those sentences
that the grammar could originally generate.

The second criterion means that
$G'$ should not overgenerate.  As previously mentioned, overgeneration
is to be avoided.  
Minimising overgeneration is important as overgeneration undermines
semantic interpretation and produces spurious parses, thereby 
increasing the
chance of the NLP system being swamped.

The third criterion is due to the fact that   
contemporary theories of semantics and pragmatics rely upon constituents 
being identified.  A grammar that simply recognised a sentence as being 
within its language, yet failed to assign a plausible parse, makes the task
of semantic interpretation harder.

The final criterion follows from demands made by NLP systems upon the
formalism used to encode the grammar.  These demands
include the grammar being declarative, formal, computationally 
tractable, and explicit.  For example, it would be possible
to extend the grammar implicitly.   Project April has a `grammar' that
consists of a set of parse trees and a matching algorithm  \cite{Haig88}.  As
such, the grammar is implicit in the parse trees and the matching algorithm.  
However, implicit 
grammars are difficult to
reconcile with contemporary theories of semantic 
interpretation.\footnote{Contemporary semantic interpretation tends to
pair semantic rules with syntactic rules.}  It is not clear how if
the grammar rules
are implicit, semantic rules can be paired with corresponding syntactic
rules.  Furthermore, 
implicit grammars require ad-hoc parsing algorithms, are of an
uncertain coverage and cannot be used to generate sentences.  Although
this criterion is obvious, not all treatments of undergeneration adhere to it.

\section{Dealing with undergeneration}

The problem of undergeneration is a special case of the problem of
{\it robust} text parsing.  A robust text parser will be able to deal
 with {\it any} sentence encountered in 
some text.  Broadly speaking, there are two ways NLP system designers
can achieve robustness:
\begin{center}
\begin{itemize}
\item Correct the sentence 
 such that the NLP system considers the sentence now to be
within $L(G)$.  This encompasses a set of ad-hoc approaches that are used
in applications such as interpreting short, ragged messages and in grammar
checking.  Undergeneration  is  reduced, at the 
expense of not being  permanent.  The NLP
 system will always need the correcting
approach in order to deal with undergeneration.  A number of applications 
using sentence correction are presented in chapter two.
\item Correct the grammar $G$, giving grammar $G'$, 
such that some sentence W which is not in $L(G)$ is within $L(G')$.
This encompasses a set of machine learning approaches, some of which
also appear in chapter two.  Unlike sentence correction, grammar correction
deals permanently with cases of undergeneration.  
\end{itemize}
\end{center}
Approaches that just use sentence correction might be able to deal with 
strings in $P$, but they erroneously treat sentences in $C$ as if they
were strings in $P$.  Approaches that just use grammar correction err the
other way round.  That is, they tend to acquire a performance grammar, and
cannot determine when a string in $P$ is encountered.
The optimal approach to the problem of robust text parsing would be one
that used sentence correction for strings in $P$, and grammar correction for
sentences in $C$.  No system uses both of these approaches, given the
complexity of both tasks.  The grammar learner presented in this thesis
is so designed that it can be used in conjunction with a sentence corrector.

\section{A novel approach to dealing with undergeneration}
In this section, an approach to dealing with undergeneration by
machine learning is presented.  Novel aspects of the learner include:
\begin{center}
\begin{itemize}
\item The combined use of data-driven and model-based learning to acquire
plausible natural language grammars.
\item Learning competence grammars, and not performance grammars.
\item Using a unification-based formalism.
\end{itemize}
\end{center}
\subsection{Overview of the approach}

When presented with an input string, W, an attempt is made to parse
W using $G$. If this fails, the learning system is invoked. First, the learning system
tries to generate rules that, had they been members of $G$, would have enabled
a derivation sequence for W to be found. This is done by trying to extend
incomplete derivations using what are called  {\it super rules}. Super rules
enable  at least one derivation 
sequence to be found for W.

Many instantiations of the super rules may be produced by the parse completion
process that  was described above. Linguistically implausible
instantiations must be rejected and   this rejection process is interleaved
with
the parse completion process.  Rejection of rules is carried out by 
model-driven and data-driven learning processes.

If no instantiation is plausible, then the input string W is deemed
ungrammatical. Otherwise, the surviving instantiations of the super rule
are linguistically plausible and may be added to $G$ for future use.

As will become clear from the rest of the thesis, this approach  
extends a grammar in a manner that meets the success
criteria for dealing with undergeneration.  
\subsection{Assumptions}

The final chapter considers the  assumptions made  in greater detail.
These assumptions include:
\begin{center}
\begin{itemize}
\item Natural language competence is (at least) strictly context 
free \cite{Pul84}.   
This 
(engineering) assumption is usually made in NLP systems.  However, it should
be noted that some languages, for example Swiss German, contain constructs
that  appear not to be context free \cite{Shie85}.  These constructs are rare 
and are usually ignored in most NLP systems \cite[p.147]{Gazd89}.  
\item Syntax is important for NLP.  
A position of syntax being distinct from semantics and
the other knowledge sources that there might be in a natural language
system has been adopted.  The reasons for 
this differentiation include modularity
(it is far easier to change the semantic formalism without having
to change the syntax if the two are not intimately  entwined), and the
freedom to select a far wider range of parsing algorithms (for example
it is difficult to see how an efficient parser could be used 
with Small's approach to natural language processing\footnote{The 
syntax of Small's
approach is implicit.  Efficient parsing algorithms such as
the LR family require an explicit grammar to compute the
LR-characteristic machine.} \cite{Smal80}).
\item Binary and unary rules are sufficient for plausible syntactic
analyses.  As most rules in manually written grammars such as 
the ANLT example are either unary or binary, this is a plausible 
assumption to make.  
\item The lexicon is complete.  This is the strongest assumption
and means that the system assumes that each word encountered is
already tagged with its part-of-speech. Given the advent of robust 
lexical taggers (e.\ g. \ \cite{Chur88,Cutt92,Brill93}), this 
assumption is commonly
made by other workers.\footnote{Work on automating lexical acquisition
(for example \cite{Knig91,Russ93,Finc93,Hugh94}) further supports the 
complete lexicon assumption.}
\item Competence and not performance grammars are learnt.  Most other
researchers learning grammar acquire performance and not competence
grammars.  This is because such researchers use inductive methods, with
no model of grammaticality, and so treat any sentence encountered as
being grammatical, regardless of how deviant that sentence may be.
Granted the fact that overgeneration undermines successful NLP and that
performance grammars (by definition) overgenerate, the research reported
in this thesis attempts to learn competence grammars.  Performance is assumed
to be dealt with by  rule-based  psycholinguistic  processes.
\end{itemize}
\end{center}
\section{Overview of the thesis}

There are six other chapters in the thesis:
\begin{center}
\begin{itemize}
 \item Chapter two presents related work: methods that  deal
 with undergeneration by
correcting  strings into   member of  $L(G)$, methods that
overcome undergeneration  by 
correcting  the  grammar, and finally,  related machine learning work
on theory (grammar) correction.

\item Chapter three presents a novel method of overcoming 
undergeneration.  Firstly  the approach is described,
secondly a worked example is presented, and thirdly, the
implementation of this theory is presented.  Finally, properties of 
the learner are discussed.

\item Chapter four expands upon the concept of model-driven (deductive)
learning, and then goes on  to both  show how model-driven learning 
can help to learn grammatical rules and  to show the failings
of model-driven learning.

\item Chapter five presents data-driven (inductive) learning, which 
complements model-driven learning, and then goes on to both show how
 data-driven learning
can help to learn grammatical rules and also to show failings
of data-driven learning.

\item Chapter six links the previous three chapters by evaluating the entire
system.  Evaluation shows the contribution that model-driven and
data-driven learning make upon the quality of the grammars 
learnt.  Principally, the chapter considers the hypothesis that using both
learning styles is better than using either learning style in isolation.
The results of the evaluation show that this hypothesis is correct.

\item Chapter seven, the final chapter, summarises the research, reconsiders
the 
assumptions made, shows ways of extending this work, and ends by
making some general conclusions.
\end{itemize}
\end{center}

\chapter{Related Work}

\section{Introduction}

As mentioned in chapter one, approaches deal with undergeneration either
by correcting the sentence, or by correcting the grammar.  Both of
these approaches will now be considered in more detail.  			

\subsection{Sentence correction}

Correction is defined as mapping a string not in $L(G)$ into
one or more  sentences that are in $L(G)$.  There 
are a variety of ways of implementing 
this relation, some of which relate to formal languages and not necessarily 
to natural languages.

The simplest method would be to skip over the point of difficulty
in the hope that the parser could be restarted.
This is known appropriately enough as {\it panic mode} \cite{Grah73}.
To be of any use, the parser needs to have an idea of what constitutes
a point of synchronisation in the sentence.  In a block structured
language such as Pascal, such a point might be a {\it begin} statement.
In the natural language context, this might be a phrasal head.  As
may be seen, the synchronisation point needs to be specified.  Panic mode
can be seen as a (limited) form of correction in that part of the sentence
is deleted (ignored) \cite{Hamm84}.  However, the deletion may also
remove an arbitrary amount of the sentence.  Consider the example of:
\numberSentence{\label{panic}
*The ants eats everything in the kitchen.}
Here we have a subject-verb disagreement.  A left to right parser would 
consider the word {\it eats} and fail.  Panicking might involve trying to
look for a verb that agreed further in sentence \ref{panic}.  Such a
 search would
exhaust the sentence here, but might work for a restarted sentence.  A 
restarted sentence is one in which the speaker can be seen to change
her plan (as realised in the sentence) whilst in the process of 
constructing the sentence in question, for example:
\numberSentence{
*The ants eats everything eat everything in the kitchen \ldots}

Correction can also change the sentence.  The idea here is that a
 sentence not in the language generated by the grammar 
is  a distorted version of a grammatical
sentence and if the underlying content of the grammatical sentence is
predictable then there can be an informed mapping from the former to
the latter.  This is the optimal method for dealing with undergeneration
with respect to corrective ability, though computationally difficult
to achieve.  For example, the  sentence in $P$
\numberSentence{* Sam chase the cat \label{before}}
could be corrected into the  sentence in $C$:
\numberSentence{Sam chases the cat \label{after}}
Correcting sentence \ref{before} into \ref{after} assumes that the 
verb does not agree with the subject (and not vice versa).  In general,
there will be many possible corrections that can be made.

\subsection{Grammar correction}

Correction of the grammar can be defined as an attempt to extend the
grammar $G$, giving grammar $G'$,  such that some sentence not in
$L(G)$ is within $L(G')$ and $L(G) \subset L(G')$.  
Grammar correction is an example of {\it machine learning}.

Carbonell and Langley comment that attempting to define {\it machine learning}
formally is always open to controversy \cite{Lang87}.  Nevertheless,
the learning task can be stated as: given one or more instances of a class, find a description that predicts which class a new instance is a member
of.  In the context of grammar learning, learning would consist of creating
a description that (amongst other things) predicts whether a sentence is
grammatical or ungrammatical.

Approaches to machine learning of grammar can be grouped broadly into
whether they are {\it inductive} (or {\it data-driven)} or {\it deductive}
(or {\it model-driven or explanation-based}).

In inductive learning
 the task is to construct a description that covers  a set of positive
examples and does not cover any of the set of negative examples.
There are many implementations of this
scheme, including the use of version spaces \cite{Mitc78} (in which
  a set of possible
descriptions is maintained), decision trees \cite{Quin86} 
(a series of questions, the answers to 
 which characterises the concept) and the early example of Winston's
Arch, which refined a single concept of the arch from near 
misses \cite{Wins75}.
Inductive learning is important in many problem solving applications,
for example in 
speech recognition and  in learning a language.  To learn
the classification, a set of features needs to be defined that
demarcate the space of possible examples.  So, for the speech example,
we may have features that correspond to those found in phonology
(segment, nasal, sonorant, etc.).  Each instance then becomes
a vector of such features.  Note that inductive
learning is not sound: the learnt concept that
has evolved may at some later stage need to be revised, given
a counter-example.  The usual example  is that of trying to
devise a law that characterises the colour of sheep.  By seeing
a flock of white sheep on the hill, it could be concluded that
sheep are  white, but the black sheep that is out of
sight will cause this theory to be revised.

{\it Deductive learning} is 
less data-driven  as only a few examples
of the concept to be acquired are needed \cite{Ellm89}.  The idea 
 is to determine why a  given example is an example.  This is
achieved with
 domain-specific
knowledge. Once the explanation is constructed, generalisation
can then take place.  Rich gives an example
of a fork in chess \cite[p.472]{Rich91}.  The learning program
would explain  why this is a bad position
and then try to generalise this explanation by
discarding unimportant aspects of the explanation. So, from an instance
of a particular fork, the learner would be able to apply this
knowledge to learn about forks in general. Constructing an explanation
is similar to constructing a proof of the example being deducible from
the domain-theory and hence deductive learning is both sound and also
can be viewed as a search over the domain theory.  Note that
if the domain theory is incomplete, the deductive learner
will be unable to learn any of the examples that cannot be proved.
Again, there are many examples of deductive learning systems, as surveyed
by Ellman \cite{Ellm89} and Mitchell, Keller and Kedar-Cabelli \cite{Mitc86}.

In sum, sentence correction can deal with undergeneration, but the
solution is temporary, and arguably, does not address the issue of
undergeneration as being due to a deficient grammar, and not being due to
the string being ungrammatical.  Grammar correction also deals with
undergeneration, but permanently, and  directly.  For the sake of 
completeness
however, this chapter considers both approaches, as outlined in the next
section.

\subsection{Systems reviewed}

The rest of this chapter presents both sentence correction and also 
grammar correction approaches dealing with undergeneration.  Machine
learning researchers also deal with undergeneration under the guise of
the {\it incomplete theory problem} and hence it is useful to consider
such related work.  The chapter ends with a discussion of how well 
undergeneration is dealt with by the previously mentioned approaches.

The systems reviewed in this chapter can be broadly classified as
follows:
\begin{center} \small
\begin{tabular}{|l|l|l|l|l|l|}  \hline
System & Boundary recognition & Corrects & Learns & Plausible & Explicit \\ \hline
NOMAD & No & Yes & No & ? & Yes \\
Weischedel & Yes & Yes & No & No & Yes \\
Parse fitting & No & Yes & No & No & Yes \\
Carbonell & No & Yes & No & ? & Yes \\ 
Mellish & Yes & Yes & No & ? & Yes \\ \hline
Berwick & Yes & No & Yes & Yes & Yes \\
Vanlehn and Ball & No & No & Yes & No & Yes \\
DACS & No & No & Yes & No & Yes \\
Inside-Outside algorithm & No & No & Yes & No & Yes \\
MIP & No & No & Yes & No & No \\ \hline
\end{tabular}
\end{center}
The first five systems deal with undergeneration by sentence correction, 
whilst the other five deal with undergeneration by grammar learning.

In the table, 
{\it boundary recognition} means that the approach recognises the
difference between performance, competence and word salad.
{\it Corrects} means that
the system changes sentences not in the language generated by 
the grammar into one or more sentences that are generated by 
the grammar.  {\it Learn} means that the system corrects the grammar.  
{\it Plausible} is meant to suggest that the system produces 
linguistically plausible parse trees for sentences in the language 
generated by the grammar.
Plausibility is difficult to define precisely \cite[p.5]{Blac93}, but
roughly stated, can be thought of as saying that constituents are 
identified correctly in the parse.  Identification is stronger than
simply bracketing a sentence, which is a common criterion for parse
plausibility \cite{Blac93,Harr91}.  {\it Explicit} is meant to say that the NLP system's
concept of grammaticality is expressed as a distinct, symbolic grammar.

\section{Sentence correction}

Here, a variety of sentence correction approaches are presented. As
previously mentioned, there is little over-arching theory of correction
and hence all of these systems adopt a variety of strategies.

\subsection {The NOMAD System \label{nomad}}

Granger's NOMAD system  converts extremely ragged messages into a
database readable form \cite{Gran83}.  The messages are 
from a restricted domain
of Naval ship-to-shore dialogue, for example: 

\numberSentence{
Locked on open fired destroyed}
This message lacks punctuation, and lacks subjects and objects for the
verb phrases.
NOMAD would produce for the above message: 

\numberSentence{
	 We aimed at an unknown object. We fired at the object. The 
	object was destroyed. }

The system attempts to process texts in a left to right manner, and uses
scripts \cite{schank75} to give a 
predictive element.  Each word that is processed suggests
new expectations.  These expectations contribute to the meaning of the text,
and when they are not met, result in `surface-text' alerts.
The alerts can be due to unknown words, missing subjects or objects, 
missing clause boundaries, ambiguous word usage, or a lack of tense agreement.

The algorithm that NOMAD uses is: 
\begin{tabbing} 
	 {\tt  parse the message} \\
	 {\tt if a} \= {\tt blockage occurs} \\
		\> {\tt set an alert flag} \\
		\> {\tt try to continue to parse} \\
	{\tt if the text cannot be fully interpreted} \\
		\> {\tt use the failure in interpretation to resolve the alert flag}
\end{tabbing}
NOMAD encodes linguistic knowledge in terms of word-level routines (similar to 
Small's Word Experts \cite{Smal80}) and relies on user interaction to resolve 
potential solutions.  

As Granger himself notes, NOMAD is difficult to extend, due to
the use of word-level routines.  The use of scripts as the predictive element
is worrying, given their well known problems (i.e.\ difficulty
 in matching, and lack
of flexibility).  NOMAD uses a failure
to understand as a guide to locating the cause of the grammatical error, a
form of blame assignment.  To do this well, the semantics component must
be correct (as a failure here will limit any attempt at dealing with
undergeneration) and so the approach is only as good as the expectation
mechanism.  

If the parser cannot make sense of a fragment, yet the text is still
comprehensible, then according to Granger, the fragment will be 
ignored.  For example, faced
with

\numberSentence{
Toby ate the lobster and Jane crab.} 
NOMAD, if armed with a restaurant script, might generate:
\numberSentence{
Toby ate the lobster.}
Here, an expectation has been found (a person eating in a restaurant) and
so the other material would be incorrectly ignored.

There is no description of the effectiveness of the approach, other than
the size of the texts being up to 17 sentences in length.  It is  suspected
 that
the approach will not scale-up easily to that of an open domain, where
the connection between a failure to understand and a solution to 
undergeneration
may not be so obvious.  This is  an example in natural
language processing of a problem that
is tackled through constraining the domain, and thus making the solution
of less general value.

\subsection{Weischedel's meta-rules \label{mr}}

Weischedel advocates relaxing grammatical constraints when dealing
with undergeneration \cite{Weis83}.  Relaxation can be considered
as a form of sentence correction.  For example, if the grammar lacked rules dealing
with transitive verbs and had rules dealing with (say) ditransitive 
verbs, encountering a transitive verb would result in the system treating
the transitive verb as being ditrantive, albeit with an NP implicitly 
introduced into the sentence.  Such relaxation is carried-out
 by the application
of meta-rules to either the productions of the grammar, or to the parsing
configuration.  Meta-rules are said to correspond to different types of 
error.  Each meta-rule is of the form: 
\begin{eqnarray*}
C_1 C_2 \ldots C_n \rightarrow A_1 A_2 \ldots A_n
\end{eqnarray*}
where $C_i$ is the $i^{th}$ condition that has to be met, and $A_j$ 
is the $j^{th}$ action that is made if all of the conditions on the
left hand side of the rule are met.

An example meta-rule that deals with a failure for the subject and the 
verb to agree is:
\begin{verbatim}
(failed-test? (subject-verb-agree? ?x ?y))
	--> (new-configuration 
                (failed-constraint (subject-verb-agree ?x ?y)
                (substitute-in-arc (subject-verb-agree ?x ?y) T)))
\end{verbatim}
That is, the test \verb+(subject-verb-agree ?x ?y)+ is replaced 
by the Lisp atom
\verb+T+, which always evaluates to true.  Here the parser configuration is 
being changed
to allow processing to continue.  In this case, we are dealing 
with an ATN formalism\footnote{An {\it ATN} (Augmented Transition Network)
is one of many implementations of a grammar and parser \cite{Wood70}.}
but the author of the paper suggests that the approach is applicable to 
any parser that is capable of being expressed in terms of a configuration,
an obvious example being an LR parser.

The approach is to apply a meta-rule when parsing is blocked.  
The meta-rule is said to diagnose the error, relax
the violated rule, add a `deviancy note'  and allow processing to continue.

The errors that are said to be dealt with include: failed grammatical
tests, word confusions, spelling errors, unknown words, restarts and
contextual ellipses.  

The true nature of an
(apparent) error is not always obvious.  Consider the sentence:
\numberSentence{
Toby kicked the red ball and Jane the blue one.}
If the system had a simple view of coordination, with rules of the form
$\alpha \rightarrow \alpha~\mbox{conj}~\alpha$, where $\alpha$ is 
a string of terminals or nonterminals, and $conj$ is a nonterminal
symbol for a coordinating lexical category, then the above sentence might
be either a failure to coordinate sentences, or may be a case of ellipsis.
Weischedel  ranks such hypotheses by the amount of material that 
has been processed and by a partial ordering on the meta-rules themselves.
The details of the ordering are not given in the paper.
Of course, it is not clear  what this ordering should be.  The hypothesis
that accounts for most of the sentence is preferred over those that
account for less. 

If the parser cannot continue, due to the meta-rules not accounting for
the `error', then processing is abandoned.  The parser will still allow
undergeneration to take place therefore, and this will be as a result 
of oversights
in the coverage of error classes by the meta-rules and inadequacies of the 
approach.

Using meta-rules is interesting, but the performance of the scheme is
bounded by the ability to account for errors.  Also, as 
Weischedel notes, ``Significant effort is required of the grammar writer 
to devise the condition-action pairs.''\cite[p.95]{Weis80}.
    
\subsection{Parse fitting \label{pf}}

EPISTLE is an attempt at providing a grammatical and stylistic critique
of technical English texts \cite{Jen83}.  The parser uses  an augmented 
phrase structure
grammar along with a lexicon of 130,000 entries.  Parsing is divided into
three parts: simple parsing using the (`core') grammar that defines the uncontroversially
grammatical structures, a set of procedures to handle ambiguity, and finally
the fitting procedure.  The core grammar consists of about 300 rules, and
these (are reported to) account for $70 \%$ of all sentences that are to be
analysed.  Ambiguity is resolved by a metric that ranks alternative parses.

In cases where the parser fails to find an analysis 
for a sentence, the fitting
procedure tries to fit the parse
fragments  (in a global manner)  into a
 tree whose root is the start symbol of the grammar.  The 
algorithm for fitting is in two stages.  The first
stage attempts to find a head constituent\footnote{The head of a rule
is a category in the rule's right hand side that characterises 
the rule.  For example, the head of a noun phrase is 
a noun.  The head of a constituent is therefore the head of a rule
used in a parse tree.  See \S \ref{constructor} for 
a more detailed discussion of heads.} from an ordered list of 
candidates, the candidates being the fragments that are
built by the parser.  The second stage then tries to `glue' the remaining
 pre and
post positional fragments to the head constituent, such that all of the
material is accounted for.

The head constituents are searched for in decreasing order of preference:
VPs with tense and subject, VPs with tense and no subject, phrases without
verbs (NPs, PPs), non-finite VPs and finally `others'.  If more than one 
candidate is found then the candidate that accounts for more of the
material is selected.

Should the head constituent not cover the entire sentence, then the remaining
constituents are added on to either side of the head constituent, with
the following order of preference: non-VP fragments, untensed VPs and
finally tensed VPs.  The overall effect of fitting is to select the 
largest chunk of
sentence-like material and to attach left over chunks in some reasonable
manner.  Both the set of head constituents and `fillers' are
considered to be principles of syntactic well-formedness.

Here is an example from the paper.  Consider the following input string: 
\numberSentence{
Example: Your percentage of \$ 250.00 is \$ 187.50.}
This string is considered to be a sentence inasmuch as it begins with
a capital and ends with a full stop.  However, the core grammar does
not consider it to be grammatical and so will fail to assign a complete
analysis.  The set of parse fragments  constitute the basis
of subsequent repair.  

The first step is to look for a head constituent amongst the set of 
fragments.  There is an ordering
of candidates for being a head constituent, and the algorithm 
initially looks for VPs with tense
and subject.  Candidates include:
\begin{quotation}
\noindent \$250.00 is \\
percentage of \$250.00 is \\
\$250.00 is \$187.50 \\
Your percentage of \$250.00 is \$ 187.50
\end{quotation}

\noindent The final candidate accounts for most of the material of the
sentence and is therefore chosen.  Note that if no tensed VP with subject
was found, the algorithm would then look for a tensed VP with no
subject, and so on.

\begin{figure} \label{fitted}
{\tiny
\input{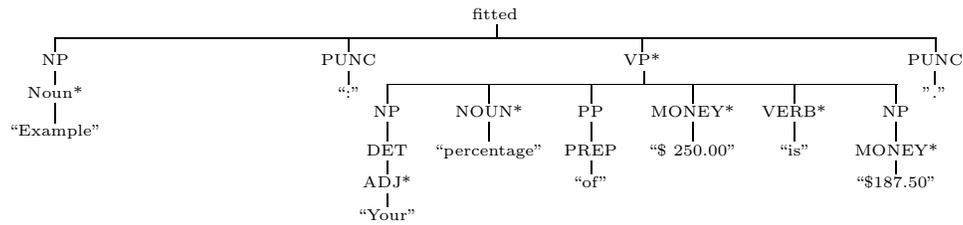}}
\caption{A fitted parse tree.}
\end{figure}

Fitting then continues by adding material that is to be found to 
the left and to the right of the chosen constituent.  Here, this results
in ``Example:'' being found prior to the head constituent and the full
stop posterior to the constituent.  These are therefore added to the fitted
parse tree, resulting in a tree that spans the entire sentence, as shown
in figure \ref{fitted}.
The starred nodes are nodes that are fitted together.

Parse fitting will always produce a tree that spans any sentence, no 
matter how badly formed that sentence might be.
 At the worst case, the
tree will consist of a root node immediately dominating the lexical
items.
Success follows from parse fitting's liberal definition of a 
well-formed tree:  for a
tree to be well-formed it does not have to have (say) an S node dominating
the entire sentence; all that is required is that the fragments
be joined up.  This therefore tries to make sense of what has
been tried and does little to correct the sentence or the parse
tree.  There is an idea of what a sentence should be composed of
in the ordering of material to look for when searching for 
candidates.  This idea of sententiality is however a little mysterious
in that it is not obvious why the authors have chosen this particular
set, and why this particular ordering.  In their paper, they do not
motivate their choice at all. 

\subsection{Carbonell et al.}
Researchers at Carnegie Mellon have developed a series
 of robust parsers around a
common theme of using case frame instantiation in a restricted 
domain.  A case frame is a structure that makes explicit the
semantic roles of (for example) subjects and objects in sentences.
These parsers include: FlexP \cite{Haye81}, CASPAR \cite{Haye81b},
 DYPAR \cite{Bogg83}, DYPAR II \cite{Carb83} and Multipar \cite{Fain85}.
  As
these are all broadly similar,   CASPAR and 
DYPAR II shall be concentrated  upon (these are reasonably well documented).

CASPAR tries to deal with simple imperative VPs in a robust manner.
For example,
\numberSentence{
Cancel math 247}
\numberSentence{
Enrol Jim Campbell in English 324}
These can be seen as examples of case constructions in that the verb is
the central concept and the attendant NPs are the arguments.  Parsing
is designed to exploit this restricted syntax.  The algorithm is simply
to parse from left to right, applying patterns corresponding to the 
verb phrase, and if a match is found, look for case fillers.

This approach will usually produce a command and a (possibly) incomplete
set of arguments, and is insensitive to extraneous input.  Arguably,
success depends upon the ability of the case-frame matching process and
any unmatched material that is lost could conceivably be relevant.

DYPAR II again uses case frames and tries to combine multiple parsing
strategies within a single framework.  These strategies include pattern
matching, semantic grammars and syntactic transformations.  This 
parser is used in XCALIBUR \cite{Carb83},
 which allows natural
language access to the XSEL expert system. It deals with spelling corrections,
ignores garbled or spurious phrases in otherwise acceptable material, 
recognises constituents when they occur in unexpected order and has a
simple ellipsis resolution mechanism.  The robustness again stems from the
simple minded parsing scheme, which is similar to CASPAR's approach.  
What is new is the use of XCALIBUR's discourse structure to resolve
unfilled case fillers.  When XCALIBUR generates a query to the user, 
there can  (within this small domain) be only a limited set of
felicitous responses by the user.  This set of responses suggest
the particular case frames to look for.  Discourse is considered
 to be stack based. If a case frame is only
partially filled, then  the first (completed) case
 frame that matches the current (incomplete)
frame is used to supply the missing material.  Expectation is also
used to constrain the search when dealing with spelling corrections.
For example, the sentence: 
\numberSentence{
*Add a dual prot disk}
is corrected to
\numberSentence{
Add a dual port disk}
as a disc descriptor is expected \cite[p.4]{Carb83}.  This is
a reduction over the 13 possibilities that a spelling program
suggested.

Parsing here may be robust, but this is by no means a solution to
undergeneration.  If the domain of discourse was not so highly
specified then trying to skim the sentence in the manner used here
would fail.  This is simply because the unspecified material of the
sentence would be ignored, thus treating undergeneration by pretending
it does not exist.

\subsection{Mellish}
Mellish presents an approach that tries to `fit' parse fragments into
a tree dominated by the start symbol of the grammar \cite{Mell89}.  However, 
unlike
parse fitting (\S \ref{pf}), the fragment 
consolidation process is carried out by
relaxing grammar rules in a manner similar to the application of
meta-rules (\S \ref{mr}).  That is, correction is defined not in terms of 
linguistic notions such as headness, but in terms of possible errors
such as omitted words in a sentence.  Mellish's approach also differs
from either meta-rules or parse fitting in that he uses heuristic search
to decide which of the many possible parses the parser should choose at any
one stage.  For example, in the 
ungrammatical sentence:
\numberSentence{*The gardener collected manure if the autumn}
the word {\it if} should be a preposition.  Hence, a rule such as
\psr{NP}{P~NP}, necessary to complete the parse, cannot be applied.
Relaxing this rule  both treats the word {\it if} as a 
preposition and also allows the parse to be completed.  

Mellish  reports that his approach can deal with single errors, but
that due to the large size of the search space, may not scale-up to 
dealing with either multiple errors or to using a large grammar.  Because
Mellish relaxes the grammar and hence presupposes that the necessary grammar
is there in the first place, any case of undergeneration will undermine this
solution to the problem of correcting errors.

\section{Grammar correction}

There are a number of results outlining the conditions
for successful  grammar  learning.  Learning successfully  means that the
system eventually acquires a grammar that meets the success criteria 
outlined in chapter one.  That is, the acquired grammar should not
undergenerate, overgenerate, or assign implausible parses to those 
sentences that it does generate.  
The system is then said to have {\it identified in the limit the
language} \cite{Gold67}.  Gold showed that context free languages can be identified
in the limit if the system has an {\it informant}.  An informant is a
knowledge source that can be called upon by the system when determining how
to revise the grammar.  Example informants include, but are not limited to, 
 people (the system would
query the user), some function which could be called upon to present
the learner with sentences of ever increasing syntactic complexity (ordering
the text), negative information, or a 
model of grammaticality (the system would use the model to
prove a sentence's  derivation sequence).  If the learner does not have
an informant, or the informant makes mistakes, 
context free languages cannot be identified in the limit.  Pinker notes
that in these two cases, the success criterion for language identification
must be relaxed \cite{Pink79}.  That is, context free 
languages will not be identified
in the limit and the learnt grammar will only be an approximation of the
(hypothetical) grammar that has generated the context free language being
identified.  Learning systems surveyed in this chapter use all of these
versions of an informant.  Clearly, not all of these version are equally
suitable for all NLP tasks.  For example, non-interactive 
systems cannot query the user 
 and interactive systems operating in real time cannot
 order the text.  However, an informant implemented as a domain theory is suitable for
all NLP applications, including text parsing.

The chapter now goes on to present a series of grammar learners.
\subsection{Berwick \label{Berwick}}

Robert Berwick bases his grammar acquisition work on 
Chomskyan ideas of Universal Grammar (which uses a model-based
informant) and on 
Marcus' Parsifal parser \cite{Marc80} \cite{Berw85}, which 
 is (generally held to be) deterministic.
Parsifal consists of two data structures: a pushdown
stack  and a three cell lookahead
buffer.  The first cell may be filled with either a word or
a phrasal category and is the item under consideration when
parsing.  

Acquisition is simple.  If the parser cannot continue
then at the point of failure an attempt is made to construct
a new grammar rule to allow parsing to continue.  If parsing
cannot continue with the addition of this rule then the 
sentence is rejected.  

Berwick starts with no grammar.

Upon encountering a case of undergeneration,  four possible
actions  are tried (this is therefore a generate and test method):
\begin{itemize}
\item Attach
\item Switch
\item Insert lexical item
\item Insert trace
\end{itemize}

{\it Attach} takes the item in the left-most cell and attaches this
to the node currently on top of the stack.  This therefore corresponds
to a late closure.  The action is constrained by
$ \overline{X}$ Syntax.\footnote{See \S \ref{constructor}
 for an explanation of $ \overline{X}$ Syntax.} If 
this rule fails, then {\it switch} interchanges
the contents of the left-most cell with the cell next to it.  This
rule  succeeds only if the rule following this new rule
itself succeeds.\footnote{The reasoning behind  this 
seems to be that if a switch is needed, 
then the sentence being parsed is in reality
a transformation of some other syntactic structure.}
 The next action is to {\it insert a
lexical item}.  This adds into the surface structure a
specified closed class word, for example `you' which may
have been deleted in the surface structure (p.146).  The final 
option is to add
a {\it trace}.  Note that if all of
these potential actions fails to allow parsing to continue 
 then the sentence is
rejected. The actions are ordered as shown in the above list
and  Berwick claims
that this ordering results in the narrowest grammar being acquired
(p.113).  

Sentence order presentation matters in that the acquisition
procedure depends upon an assumption of one rule per error (p.109).
  Therefore,
a (relatively) complicated sentence will contain grammatical structures
which need to be learnt from exposure to a simpler sentence.  Berwick's
approach, for example, needs to learn about prepositional phrases
before it can learn about preposed prepositional phrases.  There is a 
question  of how local this {\it finite error condition} 
(Berwick's  terminology) is  to be.  Berwick is forced
into his position by the choice of Marcus's Parsifal, with its
narrow idea of sentential context.  For long-distance constructs
such as gaps, the top of the stack may not relate to the part
introducing the gap and so gapping cannot be learnt.

Berwick himself admits that his approach cannot learn coordination (p.38)
and will not learn ambiguity (given the strong determinism of the
Marcus Parser).  There is no explicit analysis of the performance
of the system, for example how  the system fails, or  how
plausible  the parses produced are.  The ordering 
requirement (of the sentences)
means that the parser will have to continually reparse a training set, if
a text is being used.

\subsection{Vanlehn and Ball}

Vanlehn and Ball present a {\it version space} approach to learning
grammars \cite{Vanl87}.  A version space is a set of all generalizations consistent
with a given set of instances. In the context of learning language
this means that for a set of sentences, the version space is a set
of grammars that covers the set of well-formed sentences but excludes
the set of ill-formed sentences.  These generalisations can be given
a partial ordering of generality, although this cannot be done for
the general case of context free grammars.\footnote{The test to see
if the language generated by a 
 CFG $A$ contains the language generated by a CFG  $B$ is 
undecidable \cite{Hopc79}. 
Version spaces can be used to learn context free grammars if the
lattice is not ordered by language inclusion.}

It is a well known result that for any language $\ell$ there is an
infinite number of weakly equivalent grammars.  This means that
the version space for context free grammars is infinite.  To avoid
this problem, Vanlehn and Ball restrict the grammars by not allowing
rules of the form $A \rightarrow \lambda$, $A \rightarrow B$, where
$\lambda$ is the empty string and $A$ and $B$ are non-terminals, and 
disallowing  useless non-terminals.\footnote{A {\it useless} non-terminal
 is one that cannot be reached from the start symbol in a derivation,
or if reached, does not match the right hand side of any production in the grammar.}
Furthermore, they introduce the idea of a {\it reduced grammar}.  A
reduced grammar is one that given a set of sentences P cannot be reduced
in size by removing a production and still parse P.

Given these conditions, the version space is said to be finite.

The details of the algorithm are best illustrated by an example adapted
from the paper (p.63) in which a command language is acquired.  Sentences
and non-sentences of this language include:
\numberSentence
 {delete all}
\numberSentence
{*all delete \label{bad}}
\numberSentence
 {delete it}

Suppose the first of these sentences was presented.  This would
result in a set of grammars being produced, the form of the
productions being limited to a simple format which allows all possible
rule permutations that cover the sentence to be constructed.  These
grammars might be: 
\begin{center}
\begin{tabular}{|l|l|}  \hline
Grammar & Rules \\ \hline
G1 & \psr{S}{\mbox{delete all}} \\
G2 & \psr{S}{\mbox{delete}~S}, \psr{S}{\mbox{all}} \\
G3 & \psr{S}{S~\mbox{all}}, \psr{S}{\mbox{delete}} \\
G4 & \psr{S}{S~S}, \psr{S}{\mbox{delete}}, \psr{S}{\mbox{all}} \\ \hline
\end{tabular}
\end{center}
The most specific grammar is 
G1 and the most general is G4.

If the non-sentence \ref{bad}  was presented next then the algorithm
would try to exclude this negative example from the languages
generated by the grammars.  This sentence is only generated
by G4 and so G4 is `split' in an attempt
at exclusion.  Splitting must generate a grammar that
both generates the positive sentences and also fails to
generate the negative sentences, and is achieved by making
simple changes to the grammar.  The splitting of G4
results in three extra grammars and of these,  one
still generates a negative  sentence and so this grammar is further split
into two grammars.  One grammar generates a negative sentence and is
rejected.  The other grammar 
thus generated is
`covered' by an existing grammar and is abandoned.\footnote{The paper
does not give the details of these rejected grammars.}

The two new grammars that are produced by the splitting of G4 are:
\begin{center}
\begin{tabular}{|l|l|}  \hline
Grammar & Rules \\ \hline
G5 & \psr{S}{S~A}, \psr{S}{\mbox{delete}}, \psr{A}{\mbox{all}} \\
G6 & \psr{S}{A~S}, \psr{A}{\mbox{delete}}, \psr{S}{\mbox{all}} \\ \hline
\end{tabular}
\end{center}

This algorithm suffers from a combinatorial explosion
evident in the splitting of grammars.  For the learning of
natural language grammars, it is unclear if this work can
be directly used.  Firstly natural language grammars seem to violate
the restrictions placed upon grammars here (they have unit productions
and productions introducing gaps) and so the version space is
infinite.  A partial ordering is therefore not possible in the general
case.  Secondly, the result of this particular form of induction is a
space of grammars and so the question arises as to which 
of the grammars is to be used.  The most specific grammar 
is in effect an enumeration of the well-formed sentences of the 
training set and as
such is useless.  The most general grammar  may overgenerate too much to
be useful.  Thirdly, to use
this approach requires a set of negative sentences.  Whilst such
a set could be constructed, the need for this appears to be
unnatural.  What is really required is a method of acquiring the
syntactic structure directly from the sentence, and not as a
composite from the ill and well-formed sentences.  Fourthly, 
the training set
of sentences needs to be kept in order to evaluate new grammars
and this set could be arbitrarily large if an extensive
grammar of natural language is to be induced.  Clearly this is
unsatisfactory as one can imagine that the training set would be rather
large.  An advantage of using version spaces is that it is clear
how far  the learner has to go to before halting. The search space
is indicated by the number of grammars in the lattice and when this
contains only a single grammar, the learner can halt.

\subsection{DACS}

DACS (`Data driven ACquisition of Syntactic knowledge') \cite{Naum93} is 
an inductive,
symbolic system that infers HPSG-style grammars \cite{Poll87}.  Prior 
to parsing,
the words of the strings of the corpus are assigned their parts of speech
(this is known as {\it tagging}) and then the strings are 
sorted by length.  Sorting the corpus
is an approximation of an informant.    Once
the corpus is sorted, the system then selects the  first {\it k}  sentences and
attempts to parse these.  For cases of undergeneration, DACS
 first generates  all possible substrings and then, using
heuristic pruning, joins these substrings together.
The number of joining operations (a join is called confusingly a `gap') 
is noted for 
these sentences and the new rules corresponding to the tree  with the
lowest number of joining operations are added to the grammar.  The corpus
is then re-parsed and the process continues until the corpus  is generated
by the grammar.  There is no attempt to learn recursive rules
 or to impose linguistic plausibility constraints upon what
can and cannot be used to form the left hand side of a rule.  DACS has no
concept of headedness and so learns rules such as \psr{NP}{Adj~Adv}.  Rules
are learnt in a monotonic manner and so bad rules are never rejected.  Because
the system only joins substrings together, DACS cannot
learn  ambiguous attachments.  DACS obeys   the {\it finite error
condition}.

This particular system is interesting because of its use of a 
parsimonious (unification-based) 
formalism  and because it uses an ordering informant.

\subsection{The Inside-Outside algorithm \label{bw} } 

Baker's Inside-Outside algorithm is a popular way of inducing a stochastic
context free grammar directly from a text \cite{Bake79}.  The algorithm
has been used by workers at IBM \cite{Shar88,Blac93}, at 
Cambridge University  \cite{Lari90,Bris92,Youn94} and at Brown University \cite{Carr92}.  Prior to training, the 
set of all possible rules  (modulo some length of the number of symbols in the right
hand side) are given initial probabilities.  Then  
these values are iteratively re-estimated, in a manner 
similar to training Hidden
Markov Models, until convergence is attained.  There is no guarantee that
the algorithm will converge towards a global optimum: the algorithm hill
climbs and hence is not admissible \cite{Nils80}.
This lack of convergence is due to induction's  unsoundness in that
this particular  method picks a solution that is optimal
in a local sense, but not in a global sense.  For example, Carroll and 
Charniak,
using the Inside-Outside algorithm to estimate rules for a generated
corpus, assigned random initial values
to all the rules in the grammar \cite{Carr92}.  They reported that for each
of $300$ different random starting points, a different local minimum was
reached and that none of the grammars corresponding to these local minima was
correct,  pointing out the need for selecting sensible initial values.

Imposing constraints upon the space of rules is one way of helping
the Inside-Outside algorithm converge upon a plausible
grammar.  This insight (in the context of learning natural
language grammars)  has been exploited both by Briscoe and
Waegner \cite{Bris92} and by Carroll and Charniak \cite{Carr92}. 

Briscoe and Waegner place three restrictions upon rules.  Firstly, they
restrict rules to be in Chomsky Normal Form (CNF).  A grammar in CNF
has rules of the form \psr{A}{a} and \psr{A}{B~C}, where $A,B,C$ are
 non-terminals and $a$ is a terminal symbol.  This places a bound upon the
number of possible rules in the grammar.  Secondly, a headedness constraint
is imposed.  This constraint only licenses rules whose left hand side (LHS)
 is a {\it projection} of the rule's head category.  A projection (roughly
speaking) relates the phrase as a whole to the rule's head. 
 Note that this
constraint is a little draconian in that some rules, for example those
to do with possessives, arguably do not project the head.  The 
third constraint forces rules whose right hand side (RHS) immediately
dominates lexical items to have a second category in the RHS that is
nominal or adverbial.
This constraint seems less well justified than the other two.  Briscoe and
Waegner compile out the  feature-based formalism\footnote{A 
{\it feature-based formalism} replaces the atomic categories of a 
PSG with  complex categories consisting of  sets of {\it feature-value
pairs}.  See \S \ref{formalism} for a full description of such a formalism.} used into a CNF grammar.  The
resulting grammar contains 3786 rules and parses about $75\%$ of the Spoken
English Corpus.  The grammar also hugely overgenerates.  The following
sentence taken from the same corpus:
\numberSentence{Next week a delegation of nine Protestant ministers from
Argentina visits the Autumn assembly of the British Council of Churches.}
receives 2186624992778031036  parses.  The parse ranked as most 
likely, however, 
is close to the desired parse for such a sentence.  Note that there is no
 rule retraction: the constraints are presumed to
be complete.  Also, to be successful, such an approach is reliant upon
the parse selection mechanism.

Charniak and Carroll constrain the Inside-Outside algorithm by firstly using
a dependency grammar (which places similar constraints upon the space
of possible rules as does a CNF grammar) with a length bound upon the
number of symbols in a rule's RHS and secondly by removing rules from the
grammar that have a probability lower than some threshold.  Charniak and 
Carroll do not evaluate their system. They do present  the learnt 
grammar,  which 
appears to be plausible from a casual inspection, but also is 
extremely small, and thus presumably continues to undergenerate.

Apart from problems of convergence, the Inside-Outside algorithm has a 
computational complexity of O($n^{3}$),  in terms of 
the number of non-terminals in the grammar.  Briscoe and Waegner comment that
this complexity means that the algorithm 
cannot be used practically  to estimate a 
wide covering grammar with many non-terminals such as the Alvey Tools
Grammar \cite{Bris92}.  
Lari and Young also note in order to achieve convergence, the algorithm
needs more symbols in the grammar than is strictly necessary, thereby
exacerbating this inefficiency \cite{Lari90}.  The algorithm is also unlikely to learn
a grammar that is linguistically plausible, given the vast number of
competing, linguistically implausible grammars that could also be induced.

\subsection{MIP}

Brill, Magerman, Marcus and Santorini have developed an inductive, stochastic
system called the {\it Mutual Information Parser} (MIP) \cite{Bril92}.  MIP,
 unlike the
Inside-Outside algorithm, does not use a conventional grammar at all: the
system locates constituency by using a {\it distributional analysis} and hence
is radically non-symbolic.
Sequences of word tags are analysed and the hypothesis is that certain
subsequences occur sufficiently frequently to be  
identified as being  constituents.  MIP
uses {\it mutual information} statistics between tag sequences of
a certain length (also called {\it n-grams}).  Mutual information
is a measure of the interdependence between these {\it n}-grams. Parsing
using mutual information is a search for a bracketed structure
that maximises the partitioning into {\it n}-grams of the sentence in
question.  Parser training is done by counting frequencies of tags
pairs found in a tagged version of the Brown Corpus \cite{fran79}.
On testing, MIP made about 2 errors per sentence for 
sentences of less than 15 words and between 5 and 6 errors for sentences
between 16 and 30 words in length.  A major problem with MIP
is that distributional analysis does not produce labelled parses, which are
necessary
 in order to carry out
semantic interpretation. 
Leeds University's {\it Constituency-likelihood grammar} \cite{Gars87}, and 
 speech recognition systems (for example  \cite{Wrig93,Jone93})   
are other examples of using {\it n}-grams.

\section{Incomplete theories and machine learning}

As mentioned in the introduction to this chapter, machine learning researchers
have also looked at the problem of undergeneration  and there are clear
links between machine learning and parsing.  
{\it Explanation-based learning}  (EBL) can be viewed as being similar to parsing:
in EBL, the domain theory is used  to prove that an example is indeed
an example.  Likewise, in parsing, the grammar is used to prove
 a string  is a sentence.
Researchers in machine learning face similar problems 
as do researchers in NLP and  call undergeneration the 
{\it incomplete theory problem} (errors of omission) 
and overgeneration the {\it inconsistent
theory problem} (errors of commission) \cite{Mitc86}.  It is 
therefore also  useful to consider
related machine learning work.  Rajamoney and DeJong classify 
 and Ourston and Mooney give an overview of approaches
dealing with these problems \cite{Raja87,Ours90}.  Here, two sample
approaches are described with a view to seeing how well they can be
used to overcome undergeneration.

\subsection{EITHER}
Ourston and Mooney's EITHER system (Explanation-based and Inductive THeory
Extension and Revision) deals with the problems  of inconsistency
and incompleteness \cite{Raja87}.  EITHER
 uses
Horn clauses as its knowledge representation and makes uses of both
positive and negative examples.  The incomplete theory problem is dealt
with by trying to prove, using EBL, the positive example.  When the
proof fails, the resulting partial explanation is then examined and 
assumptions are tentatively removed.  If no negative examples are
proven after assumption removal, the assumptions are permanently removed.
If negative examples are also proved, inductive methods are used to learn
rules that make this discrimination between positive and negative
examples.
The authors also give a similar method for dealing with
inconsistent theories.    Relating this work to
approaches dealing with undergeneration suggests that the 
main difference is the use
of negative examples to determine the utility of the newly learnt rules.
If we recall, NLP systems generally do not encounter non-sentences that
are marked as such and hence EITHER cannot be directly applied.  To
apply EITHER directly would be equivalent to grammar relaxation, thereby
increasing overgeneration.

\subsection{Fawcett}
Fawcett's approach does not rest upon negative examples \cite{Fawc89}.  Like
 EITHER,
his system generates explanations and when a concept cannot be proved, 
the resulting partial explanations form the basis of the search for 
plausible explanations.  The search is limited by two constraints.  
Firstly, certain relations in the domain theory are considered to be
necessary to be proved (unlike EITHER which can remove any assumption)
and secondly, heuristics are used to arbitrate between competing
explanations.  These heuristics include succinctness, minimising the use
of unverified assumptions, preferring specific rules over less specific
rules, and so on.  Heuristics are composed in an
ad-hoc manner to give explanations a score.  Rules are generated that allow
the failing rule's antecedent to match the concept. Fawcett gives no 
evaluation of his system.  To be useful for
overcoming natural language 
undergeneration, Fawcett's approach will need a different
set of ordering heuristics  appropriate for natural language.
The approach seems to assume that only a single rule is necessary for each
unproved antecedent to complete the otherwise incomplete explanation.
Such an approach for grammar rule learning would create flat trees, which
are arguably implausible.  Fawcett's system, like EITHER, assumes that 
examples are recognised as being positive (or negative).  NLP systems do
not have this information and must at times reject word salad.  If 
Fawcett's system were used directly, 
the system would overgenerate by learning rules for bad sentences.

\section{Discussion}

In chapter one, it was argued that an approach dealing with a grammar's
undergeneration should correct sentences that need correction, learn
grammar to generate  sentences that do not need correcting, minimise both
a grammar's  undergeneration and overgeneration, assign 
plausible parses to sentences
not in $L(G)$ and finally should produce an explicit grammar.  A system
meeting these criteria would have a sentence correcting  and a 
grammar learning module.  None of the systems surveyed fully met these
criteria or allowed the possibility of these criteria being met.

NOMAD corrects sentences that need correcting,  but is reliant upon a domain theory
for success.  It would be a stronger (corrective) approach to
undergeneration if NOMAD had a performance theory (that is, why
people make performance errors).  By not differentiating between the
sentence types,\footnote{That is, whether the sentence is in $C$, $P$, or
elsewhere.} NOMAD cannot be used in conjunction with a learning
module and so fails to minimise future under- and overgeneration.

Weischedel corrects correctly and allows the possibility of being
used with a learning module.  The parses produced will not necessarily be
plausible as sentence correction is achieved by correcting the parse tree
in such a manner that a parse is produced, irrespective of what the 
parse actually says about the grammatical structure of the sentence in
question.  The use of meta-rules is the most interesting aspect of his
work.

Parse fitting is close architecturally to the ideal system dealing
with undergeneration in that the core grammar can be considered a
competence grammar and the fitting procedure can be considered a
correcting approach.  However, the approach assumes that the core grammar
does not undergenerate and treats all cases of undergeneration as errors.
Likewise, the Carbonell family of approaches do not
recognise this  distinction between knowing when to learn and 
knowing when to correct and so cannot be used in
conjunction with a learning module.  Mellish's approach suffers from the
same problem.  However, his reconstruction addresses issues such as
the need to carry out search, using declarative `meta-rules', and 
using a unification-based formalism.  As such, Mellish's method would be
the choice of a system architect designing a sentence correction method.

Berwick learns grammar correctly (with respect to the criteria for
successful treatment of undergeneration), but his approach is
purely deductive and so does not compensate for informant incompleteness.
Unlike the inductive grammar learners, he does take seriously the 
idea of language identification.

Vanlehn and Ball do not recognise differing sentence types and so
learn grammars that will overgenerate.  Because their system 
places strong restrictions upon the format of
rules,  it cannot be considered as a serious solution
to learning plausible natural language grammars.

The DACS system does not recognise the boundaries between 
sentence types and so does not know when to   correct sentences.
The learning algorithm attempts to overcome induction's unsoundness
by ordering the text by length, which is a crude approximation to
ordering by syntactic complexity.  DACS's informant will therefore be
incomplete and will not guarantee that natural languages are 
identified in the limit.  Furthermore, ordering precludes dealing
with interactive language processing tasks.  However, DACS does use 
a unification-based formalism, which sets it apart from many other grammar
learners.  

The Inside-Outside algorithm produces a grammar that, whilst dealing
with undergeneration, will tend not to minimise overgeneration
and so cannot be used easily in conjunction with a sentence correction
component.  Plausible parses are arguably not produced for sentences.
Because the approach is not incremental, it cannot be used in an 
interactive application.  Computationally, the algorithm is too
expensive to be able to learn a grammar with a large number of
non-terminals and so cannot easily be used to learn a grammar that
would assign fine syntactic analyses to sentences.  Again, there
is no informant and so the approach cannot identify in the limit any language.
However, the Inside-Outside algorithm is arguably the most attractive  of
the inductive grammar learners, given that it produces an explicit grammar
and that it has a clear notion of convergence.

None of the
incomplete theory approaches can be used directly.  This is because they
either use negative examples (which most NLP systems do not have access
to), or are unconcerned with the plausibility of the resulting
theory.  As should be clear, plausibility is an important aspect of
a natural language grammar.

In conclusion, most of the sentence correction approaches preclude
grammar correction, and conversely, most of the grammar correction
approaches preclude sentence correction.  This means that the criteria
for successful treatment of undergeneration  are not met by any of the
systems survey in this chapter.  
The next chapter presents an approach that meets the criteria for successful
treatment of undergeneration.

\chapter{The Grammar Garden}

\section{Introduction}

This chapter describes  the learning system (whose 
implementation is called {\it The Grammar Garden}).\footnote{Early
versions of the learner have been described in two papers \cite{Osbo93a,Osbo93b}.}  Conceptually,
the system consists of the following components:
\begin{center}
\begin{itemize}
\item A set of knowledge sources.  This consists of various aspects of 
syntactic and lexical information that the system makes use of.
\item A rule construction mechanism. The rule constructor extends the grammar
by building missing rules required to complete a parse for some sentence.
\item A control strategy.  Deciding which rules are to be constructed
is equally important as  the actual construction of the rules and this
is carried out by the control strategy.
\end{itemize}
\end{center}
Each of these components will be described in detail 
in section \ref{arch}.  The section following this, \ref{example},
 gives a worked example
that helps to explain how grammars are learnt.  
Section \ref{imp} outlines the implementation of the learner.  
Section \ref{conc}
discusses properties of the learner and concludes the chapter.  

\section{System overview \label{arch}}
Here, the various components of the learner are described.  
\subsection{The set of knowledge sources}
Like most {\it knowledge-based systems}, the grammar learner contains
various sources of knowledge.  In this case, they relate to the syntax
of a natural language.  Without these sources of knowledge, the system
would not be able to learn grammar.  The knowledge sources 
consists of:
\begin{center}
\begin{itemize}
\item A grammar $G$.  
\item A lexicon.
\item A language model LM.
\item A model of grammaticality MG.
\item A set of rule templates called {\it super rules}.
\end{itemize}
\end{center}
Each of these elements will now be described in turn.
\subsubsection{The grammar $G$ \label{formalism}}
 The  learning system
is designed to overcome the undergeneration of grammar $G$.  
Although $G$ may be empty, usually, it will contain some
rules.  Apart from
consideration of the set of strings generated by $G$, another aspect of $G$
is the {\it formalism} used to encode such a grammar.  Grammar formalisms
are languages to describe grammars.  Shieber presents the following three
criteria of grammar formalisms \cite{Shie86}:
\begin{center}
\begin{itemize}
\item Linguistic felicity.  How closely linguistic phenomena can be stated
as linguists would want to state them.
\item Expressiveness.  The ability to state linguistic phenomena.
\item Computational effectiveness.  Whether grammars expressed in some
formalism can be used to generate sentences, and any computational
limitations that such a formalism might present.
\end{itemize}
\end{center}
A useful formalism would allow grammars to be written in a manner that 
is natural, allow all constructs to be described, and finally, would
be computationally tractable.   {\it Unification-based} formalisms 
are widely used in computational
linguistics and arguably  meet  the above criteria well.  This 
is in contrast to other formalisms such as a
context free grammar (CFG) which might be computationally 
effective, but do not capture all generalisations
a linguist might want to make, as will be shortly demonstrated.
The grammars learnt by the Grammar Garden are therefore unification-based.

Unification-based formalisms replace the  atomic non-terminal category set
of formalisms such as a CFG 
with a set of complex {\it feature-structures}.  A feature structure is
a partial function from features to their values.  For example, a
mapping from a {\it Person} feature   might be
to the value  {\it 3}.  This mapping is commonly written as follows:
\begin{center}
\fs{Person 3}
\end{center}
Feature structures  can be recursively nested, with a feature taking a feature structure
as a value:
\begin{center}
\fs{Cat \fs{Person 3}}
\end{center}
Furthermore, two or more features within 
a feature structure can share values.  This is known as
{\it reentrancy} and is shown by a numbered square box.  For example:
\begin{center}
\fs{Person \fbox{1} \\ Cat \fs{ Person \fbox{1}}}
\end{center}
This feature structure states that the value of the {\it Person}
feature must be the same in both cases.

Within the grammar formalism used in this thesis, rules are of the
form \psr{A}{$\alpha$}, where $A$ is a feature structure and $\alpha$ is
a string of feature structures of any length.  The set of features used
that defines the set of possible feature structures 
is fixed. Fixing the set of features is an assumption made in this
thesis.  This, and other assumptions, are discussed in chapter
seven.   A typical rule
might be:
\begin{center}
\fs{Number \fbox{1} \\ Cat NP}$\rightarrow$ \fs{Cat Det}~\fs{Number \fbox{1} \\ Cat N1}
\end{center}
This is intended to say that a NP consists of a determiner followed by a
nominal phrase and that the nominal category in the
rule's right hand side (RHS)  must agree in number with the rule's
left hand side (LHS) category.  To
achieve agreement (for example) in a CFG requires that the agreement
be stated explicitly:
\begin{center}
\psr{NP1sing}{Det~N1sing} \\
\psr{NP2plu}{Det~N2plu} \\
\end{center}
 Note the
parsimonious nature of the single unification-based rule compared
with the profligate nature of  the CFG rules.

Feature structures can be ordered by how informative they are.  The
feature structure:
\begin{center}
\fs{Person 3}
\end{center}
is more informative than the feature structure:
\begin{center}
\fs{ \ }
\end{center}
and this ordering is known as {\it subsumption}. Before defining subsumption
more formally,\footnote{This definition of subsumption and unification
is taken, almost directly, from Shieber \cite[p.15]{Shie86}.}
let the notation $D(f)$ mean the value of feature $f$ in feature
structure $D$ and let $dom(D)$ be the domain of feature structure 
$D$ (i.\ e.\ its set of features).  Let a {\it path} in a feature structure be a sequence of features
that terminates in a feature value.  Given these definitions, feature
structure $D$ subsumes feature structure $D'$ 
if and only if $D(l) \subseteq D'(l)$ for all $l \in dom(D)$ and $D'(p) = D'(q)$
for all paths $p$ and $q$ such that $D(p) = D(q)$.  Atomic feature structures
only subsume identical atomic feature structures and reentrant features  can
subsume any
feature structure.  $D$ subsumes $D'$  is written as  $D \sqsubseteq D'$.

The {\it unification} of feature structures $D'$ and $D''$ is the most general
feature structure $D$ such that $D' \sqsubseteq D$ and $D'' \sqsubseteq D$
and is written as $D = D' \sqcup D''$.  Unification  fails
if the two feature structures contain inconsistent information and a
failed unification produces the inconsistent category $\bot$.  $\bot \sqcup D =
\bot$ for any category $D$.  The empty category $[~]$ unifies with any
other category.

Simple examples of unification include:
\begin{center}
\begin{tabular}{ll}
\fs{Cat NP \\ Person 3} $\sqcup$ \fs{Cat NP \\ Person 3}& $=$ \fs{Cat NP \\ Person 3} \\ \\
\fs{Cat NP \\ Person 3} $\sqcup$ \fs{Cat NP \\ Person \fbox{1} }& $=$ \fs{Cat NP \\ Person 3} \\ \\
\fs{Cat NP \\ Person 3} $\sqcup$ \fs{Cat NP \\ Person 2}& $= \bot$
\end{tabular}
\end{center}

As a notational convenience, feature structures will sometimes be represented
by atomic categories.  These conventions will be given when necessary.

A useful (notational) extension to the formalism is {\it disjunction}.
Braces are conventionally used to indicate disjunction in feature
structures. For example,
\begin{eqnarray*}
\left\{ \fs{Person 3 \\ Cat NP}, \fs{Person 3 \\ Cat N1} \right \}
\end{eqnarray*}
is the disjunction of the two categories within the braces.  This notation
can  be extended to deal with value disjunction:
\begin{center}
\fs{Person 3 \\ Cat \{NP,N1\}}
\end{center}
This category is another way of stating the previous disjunctive
category.  Disjunction really is just a notational extension as it
is possible to multiply-out the disjuncts into a set of conjuncts.  The
following list  of disjunctive feature structures properties
makes this clear:
\begin{center}
\begin{tabular}{ll}
$A \wedge (B \vee C) = (A \wedge B) \vee (A \wedge C)$ & (distribution) \\
$A \vee A = A$ & (idempotency) \\
$\bot \vee A = A$ & (bottom) \\
$ [~] \vee A  = [~]$ & (top) \\
$A \sqsubseteq B$ if and only if $A \vee B = B$ & (interdefinability of
subsumption and disjunction)
\end{tabular}
\end{center}
where $A,B$ and $C$ are possibly disjunctive feature structures.  This
set of properties is taken from Pollard and Sag \cite{Poll87}.
The disjunctive unification of some $(A \vee B) \sqcup ( C \vee D)$ is
$(A \sqcup C) \vee (A \sqcup D) \vee (B \sqcup C) \vee (B \sqcup D)$.
Disjunctive subsumption of some $(A \vee B) \sqsubseteq ( C \vee D)$ is
$(A \sqsubseteq C) \wedge (A \sqsubseteq D) \wedge (B \sqsubseteq C)
 \wedge (B \sqsubseteq D)$.
Disjunction is to be interpreted
strongly in the sense that $A~\vee~B$ is considered true only in the
case when neither $A$ nor $B$ can be inferred, but the disjunction
$A~\vee~B$ is true \cite{Lu93}.  
Ordinary unification is almost linear in time 
complexity \cite{Mart82}, whilst
disjunctive unification is NP-complete \cite{Kasp86}.

The formalism presented is (apart from the notational use of
disjunction) a variant of the PATR-II formalism \cite{Shie83}.  PATR-II is
intended to be a tool formalism and so can be used to simulate most
other unification-based formalisms.  

\subsubsection{The lexicon}
The learner does not acquire lexical information and presumes a complete
lexicon.  That is, all words encountered have a lexical entry.  As outlined
in chapter one, this is another assumption made by this research.

\subsubsection{The language model LM}
 The language
model is the basis of grammaticality used by the data-driven learner, as
explained in chapter five.  
\subsubsection{The  model of grammaticality MG}	
The  model of grammaticality forms the basis of grammaticality for
the model-based learner, as explained in chapter four.
\subsubsection{Super rules}
{\it Super rules} are templates that correspond (roughly) 
to various kinds of missing rules from G. In the Grammar Garden, they
 consist of the following unification-based rules:
\begin{center}
\begin{tabular}{ll}
$[~]\rightarrow[~]~[~]$ & (binary) \\
$[~]\rightarrow[~]$ & (unary)
\end{tabular}
\end{center}
The binary super rule says that any category can be written as 
any two categories.  The unary super rule says that any category
can be re-written as another category.  
The super rules, being completely vacuous, subsume all unary and binary
rules that the grammar can express.  Rules learnt are refinements of these
super rules.    The super rules exploit the expressive power
of a unification-based grammar, implicitly representing  all unary and
binary rules, and hence are complete.  Note that 
using just unary and binary super rules is sufficient to be   able to 
learn rules to generate any sentence.\footnote{Any string of terminals can
be generated using just unary and binary rules.  For example, using 
some binary rule would rewrite a string of 
length $n, n>2$ to a string of length $n-1$.
Repeating this rewriting process would eventually 
result in a string of length $1$.
Strings of length $1$ can be rewritten using some unary rule.}
    Unary and binary super rules
are not necessarily sufficient to produce rules that always 
assign plausible parses
for sentences.  Linguists typically use rules without a  RHS category
 to introduce
gaps into parse trees for constructions such
as preposed constituents.  For the learner  to acquire such rules requires
using 
 a 0-ary super.  However, this would greatly increase the
search space size.  Hence, gapping rules are not learnt.  Likewise, 
linguists typically use rules with a RHS containing three categories 
when generating constructs such as ditransitive VPs.  Again, this would
imply that the learner used a trinary super rule, thereby increasing the
search space.  
This assumption, of only learning unary and binary rules,
is discussed in chapter seven.

Super rules are similar to Weischedel's meta-rules (which correspond to 
various kinds of syntactic error) in that they correspond to various
rules missing from the grammar.  However, unlike meta-rules (which
explicitly enumerate the type of error that the system deals with),
super rules implicitly represent all the missing (unary and binary)
rules.  Hence, the super
rules are complete.  By comparison, the set of meta-rules will most
likely be incomplete.  The super rules are also similar to Mellish's
generalised rules.  However, Mellish's rules  simply join local trees
together.  The super rules on the other hand can be used to generate
local trees.  Hence, the super rules have no reliance upon the 
initial grammar.  

\subsection{The rule construction mechanism \label{constructor}}
Rule construction within the learner consists of determining what the
LHS of the rule is, given a RHS.  Such a determination requires a theory
of how rule LHSs relate to rule RHSs.  One such theory, which is
popular in many linguistic theories,  is 
{\it X-bar Syntax} \cite{Jack77}.  There are many variations of
X-bar Syntax.  Broadly speaking, X-bar syntax constrains rules 
 either to be of the form:
\begin{eqnarray*}
H\fs{\mbox{BAR}~N} \rightarrow \alpha~H\fs{\mbox{BAR}~N}~\beta
\end{eqnarray*}
(which is recursive), or of the form:
\begin{eqnarray*}
H\fs{\mbox{BAR}~N+1}  \rightarrow \alpha~H\fs{\mbox{BAR}~N}~\beta
\end{eqnarray*}
(which is non-recursive).  Here, the notation 
$H\fs{\mbox{BAR}~N}$ indicates any category with a 
BAR level of $N$.  The $H$ category within the RHS is called the
{\it head} of the phrase.  The head characterises that phrase.  For example,
the head of a noun phrase is a noun.  The LHS category is called the 
{\it projection} of the head.   Lexical categories usually
have a bar level of zero and intermediate phrases have projections that 
raise this bar level.  Bar levels can only be raised up to some limit
 and
a category whose bar level is equal to this limit is called the {\it maximal
projection} of the phrase.    For example, the maximal projection of a noun
is a noun phrase.  X-bar syntax therefore helps determine what the LHS of the
rule should be and places a limit upon the number of rules that might be
learnt.  So, the rule \psr{VP}{N1} would not be created as it
has a projection that is not related to the head.  X-bar syntax is therefore
a powerful constraint upon rules.

X-bar syntax constitutes a restriction over the space of possible rules
that are linguistically plausible.   From a learning perspective, this 
restriction can be thought of as an {\it inductive bias} \cite{Haus88}.  An
inductive bias is a constraint upon the space of possible hypotheses that the
learner might consider.  We saw in \S \ref{bw} that Briscoe and Waegner 
used X-bar syntax as an inductive bias upon the Inside-Outside Algorithm.
Likewise, and independently \cite[p.40]{Osbo92},  we also use X-bar 
syntax as an inductive bias.

X-bar syntax relies upon locating the head in the RHS and determining if the
rule is recursive, or finite.  This constitutes knowledge of syntax.  
  As such, implementations of
this theory of syntax will in practice be incomplete.  That is, they 
will not always 
be able to determine if a rule is finite or recursive, or what the head of
the rule is.  Problems of determining what the LHS should be 
are tackled by creating the {\it disjunction} of the possible categories
that might form the LHS category.\footnote{This is  similar to  
 the INDICO system's approach to LHS construction \cite{Stah93}.}  Later, 
when sufficient evidence has built
up, the LHS can be refined to be that disjunct that has proved itself to
be the projection of the rule.  Chapter five explains how rule LHSs are 
refined.

In the learner, unary rules are constructed as follows.  Given some RHS of the form
$A\fs{\mbox{BAR} X}$, the rule that the learner will initially construct is:
\begin{eqnarray*}
 A_i\fs{\mbox{BAR} X+1} \rightarrow A\fs{\mbox{BAR} X}
\end{eqnarray*}  
If the bar level of category $A$
exceeds the maximal bar level, then it is not used in the rule LHS.  If no 
category can be used in the LHS then the rule is not created.  Note that 
unary rule construction only creates non-recursive rules.
This is because recursive unary rules lead to parser nontermination.

For some RHS of the form:
\begin{eqnarray*}
A_1\fs{\mbox{BAR} X} A_2\fs{\mbox{BAR} Y}
\end{eqnarray*}
 the learner initially constructs
 the rule:
\begin{eqnarray*}
\{ A_1\fs{\mbox{BAR} \{X, X+1 \}},A_2\fs{\mbox{BAR} \{Y, Y+1\}} \}\rightarrow A_1\fs{\mbox{BAR} X} A_2\fs{\mbox{BAR} Y}
\end{eqnarray*}  
Again, if the bar level of category in the LHS
exceeds the maximal bar level, then it is not disjoined with the
other categories.  

Minor categories (such as \fs{Det +}) 
do not have bar levels and take no part in the rule
construction process.  However, some rules, for example those treating
possessives, arguably have  a LHS which is a minor category.  An example
rule might be \psr{Det}{NP~POSS} (where POSS is the lexical
entry for the item ``'s'').  The extra
computational expense of allowing  minor categories to be part of the
LHS is such that this incompleteness with respect to possessives can
be overlooked.    If the LHS cannot be created, then the rule is judged
to be implausible and so rejected.  

Note that there is no reason why the rule constructor could not use 
theories of phrase structure other than X-bar syntax.  

\subsection{The control strategy}
The task of the control mechanism is to allow the rule constructor to
create rules that allow at least one parse to be created for the sentence
$W$ that the system is currently dealing with.  The sentence $W$ guides the 
learning process and, in a sense, learning could be said to be incremental.  
Compare this with
batch-orientated approaches such as the Inside-Outside algorithm, which
learn all possible rules at once.  By being incremental, the learning 
approach only learns as much as it
needs to.  

The control strategy  needs to be exhaustive
(it should allow all ambiguous attachments to be learnt), goal directed
(it should consider strings of categories that have a chance of allowing
the parse to be completed) and efficient.  {\it Chart parsers} \cite{Kay86}
 meet these 
demands well and when adapted, can be used as the learning control strategy.
Furthermore, they are very
flexible, leaving the choice of processing strategy open,
do not require expensive pre-processing of the grammar and use
a form of book-keeping that avoids the need to re-compute previously
constructed derivation sequences.  These last two advantages make
chart parsing especially suitable for interleaved learning and parsing.
The chief disadvantage of chart parsers is  that their speed  is
theoretically slower in the worst case than for other algorithms,
notably the LR family of parsers.  But in practice, the parse time between
a state-of-the-art chart parser and an LR-style parser is not that 
great \cite[p.128]{Carr93}.  

\begin{figure}
\centering
\leavevmode
\epsffile{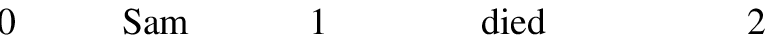}
\caption{\label{c1} An empty chart}
\end{figure}

A chart parser accepts an input string of length $n$ and builds a
data-structure known as a {\it chart}.  A chart consists of a set of
{\it vertices} that label the input string, starting from 0 to
$n$.  For example, the sentence:
\numberSentence{Sam died}
would have an initial chart as shown in figure \ref{c1}.

The parser now proceeds to build structures called {\it edges} that
span vertices.  An edge can either represent the hypothesis of a phrase
starting from vertex $i$, or can represent the fact that a phrase
has been found, spanning from vertex $i$ to vertex $j$, where $i \leq j$.
The former sort of edge is called {\it active} and the latter edge
is called {\it inactive}.  An  edge is of the form:
\begin{eqnarray*}
\langle \mbox{{\it start}},\mbox{{\it end}},\mbox{{\it found}},\mbox{{\it needed}} \rangle
\end{eqnarray*}
 {\it Start} indicates the left hand side of the edge's span, 
{\it end} indicates the right hand side of the edge's span,
{\it found} is  (roughly) a list of non-terminals that have been seen
as evidence of this being a particular phrase, and {\it needed} is
a list of categories needed before the active edge becomes inactive.
An active edge has a non-empty {\it needed} list, whilst an inactive edge has an empty {\it needed} list of categories.

\begin{figure}
\centering
\leavevmode
\epsffile{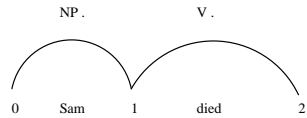}
\caption{\label{c2} A chart with lexical edges}
\end{figure}

Initially, the parser adds {\it lexical edges} to the chart.  A lexical edge
is an inactive edge that only spans a single word.  
Intuitively, adding lexical edges can be thought of as `seeding' the chart
with edges that may ultimately represent phrases.  Assuming the grammar:
\labelRule{\psr{S}{NP~VP}}{S}
\labelRule{\psr{VP}{V}}{VP1} 
and the lexicon:
\begin{center}
\begin{tabular}{l}
\lexical{{\it Sam}}{NP} \\
\lexical{{\it died}}{V} \\ \
\end{tabular}
\end{center}
 Adding  lexical edges would give 
the chart   shown in figure \ref{c2}.

Conventionally, an inactive edge is depicted as an arc labelled with the
phrasal category represented by the edge, and an active edge is depicted
as a dotted pair whose left hand component is the {\it found} list and 
right hand component
is the {\it needed} list.  Active edges are normally also shown with their
potential category that they might eventually `become'.  For simplicity, this
extra information is omitted.  

The next step is to {\it propose} new active 
edges from inactive edges in the chart.  New edges are proposed for rules
whose first daughter matches with the category of an inactive edge.  

\begin{figure}
\centering
\leavevmode
\epsffile{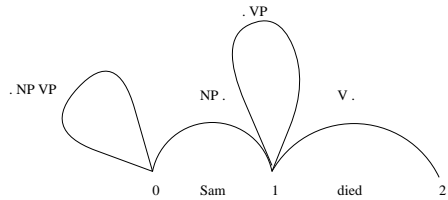}
\caption{\label{c3} A  chart after proposing edges }
\end{figure}

Proposing edges in this way gives a bottom-up parsing algorithm.
Proposing new edges to the chart 
so far  constructed results in the  chart shown in figure \ref{c3}.

\begin{figure}
\centering
\leavevmode
\epsffile{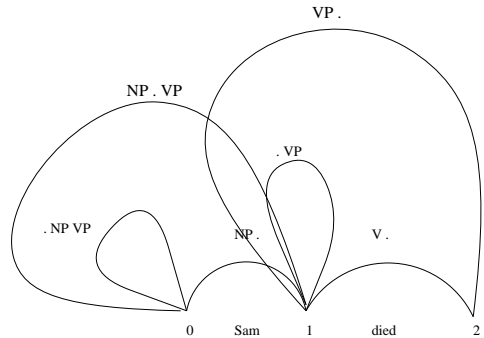}
\caption{\label{c4} A chart after extending edges}
\end{figure}

After adding edges, the final step is to {\it extend} active edges present
in the chart.  An active edge of the
 form $\langle i,~j,~\alpha,~B~\beta \rangle$ can be extended if there is
an inactive
 edge of the form $\langle j,~k,~B,~\mbox{nil} \rangle$
(for some $i \leq j \leq k, \alpha, \beta \in N^*, B \in N$).  Extending
such an edge adds the edge $\langle i,~k,~\alpha B,~\beta \rangle$
to the chart.  Extending the edges in the chart results in figure \ref{c4}.
An inactive edge has been added, showing that a VP has been found, along
with an active edge.  

For completeness, both {\it propose} and {\it extend} need to be called 
repeatedly until no further 
edges can be either added to the chart, and no edges
in the chart can be further extended. 

This has informally described how a chart parser works.  In order to adapt
the chart parser for learning, the basic chart parsing algorithm needs to
be changed.  
During interleaved parsing and learning, the grammar is 
partitioned into the grammar $G$ and the grammar $Q$.  
$G$ is the original grammar and $Q$ is the learnt grammar.  
This is motivated on efficiency 
grounds (the super rules, being complete, will lead to previously learnt
rules being re-learnt and hence there is no need to propose previously
learnt rules). During normal parsing, both $G$ and $Q$ are used.

After parsing some sentence $W$ has {\it failed}, the resulting chart 
is then `seeded' with super rules at each vertex.  That is, super rules
are proposed.  Seeding the chart allows the possibility of any unary or
binary branching parse to be completed, irrespective of the original grammar
$G$.  Parsing then continues as before, except this time proposing 
new edges using the grammar
$G$ and the super rules.  Upon instantiating  the RHS of a super rule, that 
instantiated super rule is then passed to a {\it critic}

The critic is composed of a data-driven and model-driven component and is
used to {\it refine} (or {\it reject})
instantiated super rules.  Super rule refinement 
consists of using the rule constructor to create the LHS, and super
rule rejection  consists of discarding rules that contain 
categories that cannot co-exist as part of a rule's RHS.  Chapters four 
and five describe in detail how RHSs are refined and rejected.  If the LHS 
cannot be
constructed, or no categories can co-exist within the RHS of the rule, 
then the edge containing the instantiated super rule is marked
as being {\it bad}.  Edges marked as being bad are not proposed from, and
are not used in parse tree.  By marking edges in this way, the learner
does not pursue possible parses that arise from linguistically implausible
rules.  If the edge is not marked as being bad, then that edge is treated
as other edges, and the newly constructed rule is retained.  

Parsing continues until no edges can be extended.  Any rules used in a
parse for W that are not subsumed by any other rule in the grammars $G$ or
$Q$ 
(ignoring the super rules) are kept for later use.

After interleaved parsing and learning has concluded, the system can
either be used to deal with another sentence,  can be used to
post-process the learnt rules in an attempt to reduce overgeneration, or
can be used to merge the learnt rules with the original grammar, thereby
delivering a grammar which can be used in another NLP application.

In conclusion, the control strategy is based upon a chart parser, and makes
use of a critic to determine if some string of categories should be passed
to the rule constructor.  The critic, as well as refining edges, helps
prevent the learner from considering unfruitful branches of the search
space.  

\section{A worked example \label{example}}
\begin{figure}
\centering
\leavevmode
\epsffile{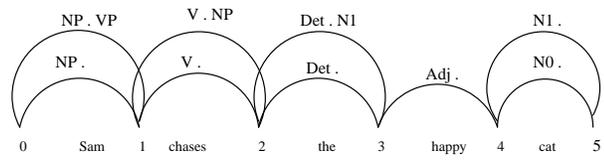}
\caption{\label{chart1} The resulting chart after a failure to parse }
\end{figure}

\begin{figure}
\centering
\leavevmode
\epsffile{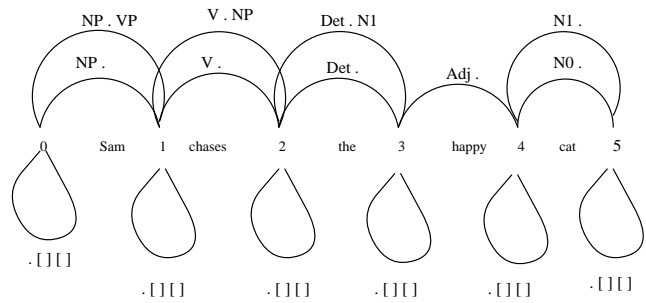}
\caption{\label{chart2} The  chart after seeding with binary 
super rules }
\end{figure}

\begin{figure}
\centering
\leavevmode
\epsffile{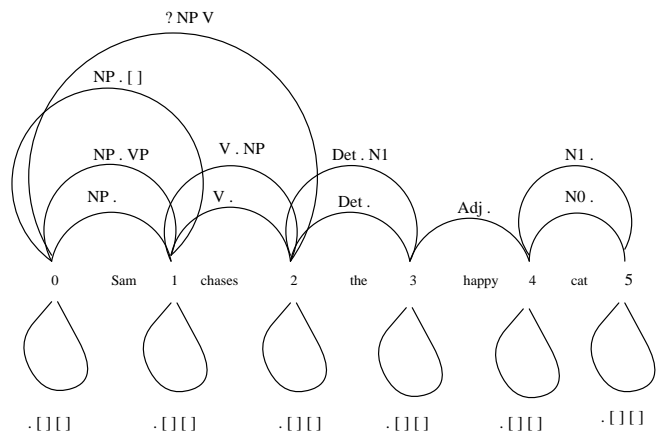}
\caption{\label{chart3} The chart after extending a super rule instantiation}
\end{figure}

Here, an example will be worked through.
  Note that although the grammar is 
unification-based,
 atomic symbols are used instead as a notational
 convenience.\footnote{This grammar is meant to be  demonstrational
 only and not intended to make any linguistic claims.}
Suppose the system started with the following grammar:
\labelRule{\psr{S}{NP~VP}}{S1}
\labelRule{\psr{NP}{Det~N1}}{NP1}
\labelRule{ \psr{VP}{V0~NP}}{VP1}
and with the lexicon:
\begin{center}
\begin{tabular}{l}
\lexical{the}{Det} \\
\lexical{cat}{N1} \\ 
\lexical{happy}{Adj} \\
\lexical{chases}{V0} \\
\lexical{Sam}{NP} \\
\end{tabular}
\end{center}
Parsing the sentence:
\numberSentence{Sam chases the happy cat \label{exampleSentence}}
will fail, producing the chart as shown in figure \ref{chart1}.\footnote{In this, and in the other diagrams of various charts, a few edges are missing.  These
are unimportant for the  exposition.}
Enabling learning and 
adding new edges (`seeding') 
using the binary super rule,\footnote{For simplicity, only the
 binary rule will be used in this example.  It should be clear how unary rules are learnt.}
produces the chart as shown in \ref{chart2}.  
That is, each inactive edge has been used to propose a new active edge that
can join with an inactive edge in the next vertex.  Clearly, parsing can now
be resumed using these new active edges.  Concentrating just on vertex 0,
extending these edges produces the chart shown in figure \ref{chart3}.
The extended edge (marked with a question mark) represents the possible 
application of a rule whose RHS is NP V.  The task now is to decide if this
RHS can be used as part of a plausible rule.  Let us assume that criticising 
the edge containing these categories results in the edge being judged as 
being implausible.  This will both not lead to a new rule being constructed,
and will also not lead to further edges being proposed from this edge.

\begin{figure}
\centering
\leavevmode
\epsffile{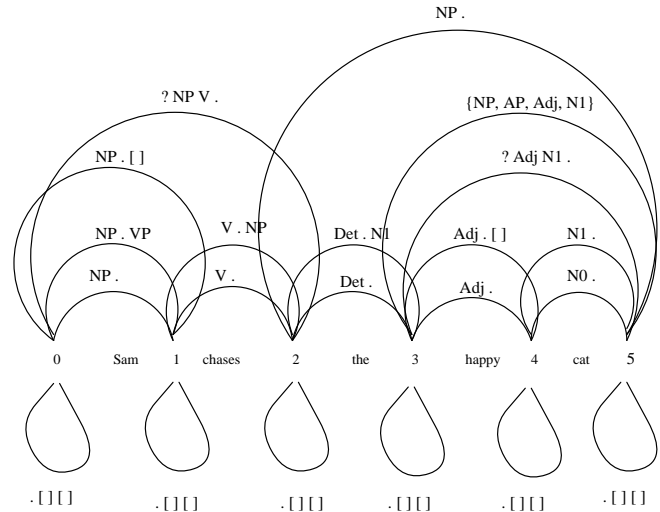}
\caption{\label{chart4} The chart after extending another super rule instantiation}
\end{figure}

\begin{figure*}
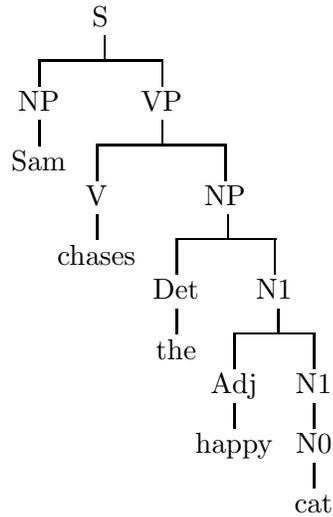
				
{\parskip=0pt\offinterlineskip%
\hskip 18.63em\hbox to 0.50em{\hss{S}\hss}%
\vrule width0em height1.972ex depth0.812ex\par\penalty10000
\hskip 16.75em\vrule width2.25em height-0.407ex depth0.500ex%
\vrule width.04em\vrule width2.00em height-0.407ex depth0.500ex%
\vrule width0em height1.500ex depth0.500ex\par\penalty10000
\hskip 16.75em\vrule width.04em%
\hskip 4.21em\vrule width.04em%
\vrule width0em height1.500ex depth0.500ex\par\penalty10000
\hskip 16.25em\hbox to 1.00em{\hss{NP}\hss}%
\hskip 3.25em\hbox to 1.00em{\hss{VP}\hss}%
\vrule width0em height1.972ex depth0.812ex\par\penalty10000
\hskip 16.75em\vrule width.04em%
\hskip 1.96em\vrule width2.25em height-0.407ex depth0.500ex%
\vrule width.04em\vrule width2.19em height-0.407ex depth0.500ex%
\vrule width0em height1.500ex depth0.500ex\par\penalty10000
\hskip 16.00em\hbox to 1.50em{\hss{Sam}\hss}%
\hskip 1.25em\vrule width.04em%
\hskip 4.40em\vrule width.04em%
\vrule width0em height1.972ex depth0.812ex\par\penalty10000
\hskip 18.50em\hbox to 0.50em{\hss{V}\hss}%
\hskip 3.69em\hbox to 1.00em{\hss{NP}\hss}%
\vrule width0em height1.972ex depth0.812ex\par\penalty10000
\hskip 18.75em\vrule width.04em%
\hskip 2.71em\vrule width1.75em height-0.407ex depth0.500ex%
\vrule width.04em\vrule width1.63em height-0.407ex depth0.500ex%
\vrule width0em height1.500ex depth0.500ex\par\penalty10000
\hskip 17.25em\hbox to 3.00em{\hss{chases}\hss}%
\hskip 1.25em\vrule width.04em%
\hskip 3.34em\vrule width.04em%
\vrule width0em height1.972ex depth0.812ex\par\penalty10000
\hskip 20.75em\hbox to 1.50em{\hss{Det}\hss}%
\hskip 2.13em\hbox to 1.00em{\hss{N1}\hss}%
\vrule width0em height1.972ex depth0.812ex\par\penalty10000
\hskip 21.50em\vrule width.04em%
\hskip 1.96em\vrule width1.50em height-0.407ex depth0.500ex%
\vrule width.04em\vrule width1.25em height-0.407ex depth0.500ex%
\vrule width0em height1.500ex depth0.500ex\par\penalty10000
\hskip 20.75em\hbox to 1.50em{\hss{the}\hss}%
\hskip 1.25em\vrule width.04em%
\hskip 2.71em\vrule width.04em%
\vrule width0em height1.972ex depth0.812ex\par\penalty10000
\hskip 22.75em\hbox to 1.50em{\hss{Adj}\hss}%
\hskip 1.50em\hbox to 1.00em{\hss{N1}\hss}%
\vrule width0em height1.972ex depth0.812ex\par\penalty10000
\hskip 23.50em\vrule width.04em%
\hskip 2.71em\vrule width.04em%
\vrule width0em height1.500ex depth0.500ex\par\penalty10000
\hskip 22.25em\hbox to 2.50em{\hss{happy}\hss}%
\hskip 1.00em\hbox to 1.00em{\hss{N0}\hss}%
\vrule width0em height1.972ex depth0.812ex\par\penalty10000
\hskip 26.25em\vrule width.04em%
\vrule width0em height1.500ex depth0.500ex\par\penalty10000
\hskip 25.50em\hbox to 1.50em{\hss{cat}\hss}%
\vrule width0em height1.972ex depth0.812ex\par}
\caption{The final parse tree \label{example3}}
\end{figure*}

Suppose now the parser extended the active edge 
starting at vertex 3, and then  tried to extend this active edge 
Adj . $[~]$  starting at vertex 3
and ending at vertex 4, with the inactive edge labelled N1 spanning 
from vertex 4 to 5.
This results in an inactive edge (marked with a question mark) 
being added to the chart (as shown in figure \ref{chart4}).  Because the edge
originated from a super rule, it is passed to the critic.  Assuming that 
the critic judges this edge to be plausible, this edge is then passed to the
rule constructor.  The constructor then builds the rule:
\begin{center} \label{expo}
$\{AP,NP,Adj,N1\} \rightarrow Adj~N1$
\end{center}
Now, the active edge Det . N1 can be extended with this criticised edge,
resulting in the chart shown in figure \ref{chart4}.
Now, interleaved parsing and learning can continue, until finally, the
parse as shown in figure \ref{example3} is produced.  That is, the active
edge spanning from vertex $1$ to vertex $2$ has been extended with the
inactive NP edge, giving an inactive VP edge spanning from vertex $1$ to
vertex $5$.  Now, the active edge spanning from vertex $0$ to 1 can
be extended with the VP edge, giving an inactive edge that spans the entire
chart.  Note that the disjunctive LHS of the learnt rule has, through 
unification, become a single, non-disjunctive category in the parse tree.
That is, $\{AP,NP,Adj,N1\} \sqcup N1 = N1$, where $N1$ is the first category
within the RHS of the rule NP1.  In general, parse trees will contain
disjunctive categories.
\section{Implementation \label{imp}}

The system consists of 3300 lines of Common Lisp (AKCL) and has been   
run on a Sun 3/50, a Silicon Graphics Indigo,
 a Sparc Workstation  and an IBM RS6000.  It is
embedded in the Grammar Development Environment (GDE), 
version 1.33 \cite{Grov92}.  Although
the
GDE is a mature and useful tool for manipulating unification-based grammars,
its chart parser had to be re-written for the following reasons:
\begin{center}
\begin{itemize}
\item It uses  non-disjunctive unification, whereas the Grammar Garden uses 
disjunctive unification.
\item The GDE's parser does not carry scoring information necessary for the
data-driven learner (see chapter five for an explanation).
\item  The GDE's parser uses a  monolithic grammar, but the learner uses
a partitioned grammar of original rules and learnt rules.
\item It only calls  {\it propose} once for each inactive edge (therefore introducing
incompleteness with respect to a dynamic grammar).  
\end{itemize}
\end{center}
The learning system has a number of flags that allow the system to be parameterised.
Here is a typical setting which shows that the system will learn, using a 
critic composed of a number of components (type checking,  LP rules 
and  a Head Feature convention).  As it turns out, these components make-up
the model-based aspect of learning, which is explained in the next chapter.
Parameterising the system allows experimentation between
various learning styles and produces  distinct learning configurations.
\begin{verbatim}
3 Gde> flags
Current flag settings:

Learning                : ON
Type checking           : ON
LP rules                : ON
HFC                     : ON
SBL                     : OFF
Training                : OFF
\end{verbatim}
Here, the implementation is shown learning  the same rule
as was learnt in the worked example, using the same grammar and
lexicon as before.
\begin{verbatim}
9 Parse+>> Sam chases the happy cat
1 rule(s) acquired.
1 parse(s)
10 Parse+>> !*parses*
((``S1''
  ((|Sam|)
   (``VP''
    ((|chases|) 
    (``NP1'' ((|the|) 
              (``*binary1583'' ((|happy|) 
                                (|cat|))))))))))
11 Parse+>>
\end{verbatim}
The new rule is used in the parse and prefixed by a star.  This 
is the same    rule as learnt in \S \ref{expo} (namely $\{AP,NP,Adj,N1\} \rightarrow Adj~N1$).

Note that  the chart parser is bottom-up.  Because the
chart parser drives learning, learning also takes places bottom-up.  With some
work, learning could take place using a parser that adopted a different 
strategy.  
\section{Conclusion \label{conc}}
This chapter presented a grammar learning system that can be used to deal with
a grammar's undergeneration.  Of the systems presented in chapter two, the
learner is closest to Briscoe and Waegner's batch-orientated 
approach (see \S \ref{bw}).
They use a unification-based formalism, a data-driven component, and also
a model-based component.  Our incremental 
learner differs in that it uses a disjunctive
unification-based formalism, makes greater use of the model-based component,
and
attempts, unlike Briscoe and Waegner,  to deal with overgeneration.  

The rest of this section discusses 
properties of the grammar learner.  These properties include:
\begin{center}
\begin{itemize}
\item Meeting the success criteria.
\item Completeness.
\item Termination.
\item Complexity.
\end{itemize}
\end{center}
The first property is concerned with how well the learner deals with 
undergeneration.  The other properties are concerned with computational
aspects of the grammar learner.
\subsection{Meeting the success criteria}
If we recall from chapter one, the criteria for success
 consists of attempting to reduce a grammar's undergeneration,
trying not to overgenerate, learning a grammar that assigns plausible
parses for sentences, and finally using a grammar in a form suitable
for NLP.

Undergeneration is  clearly reduced by learning 
 (even if just for the sentence that the extended grammar can now
generate,  but the unextended grammar could not generate). 

Overgeneration is controlled  through the rejection of implausible rules
by the critic.  For
example, if  when learning rules for the example 
sentence \ref{exampleSentence}, the critic allowed rules such as 
\psr{NP}{Det~NP} and
\psr{NP}{Adj~N},  
to be acquired,  then these rules
would lead to overgeneration;the ungrammatical string \ref{thing},for example,
would be generated by the extended grammar:
\numberSentence{*Sam chases happy cat \label{thing}}

Plausibility is closely related to overgeneration.  For example, the 
previous rules do not assign a plausible parse to 
sentence \ref{exampleSentence}.  Hence, rejecting such rules using the critic
 helps produce plausible parses.

  By using a unification-based grammar, the learner
easily meets the final criteria of using a grammar in a form suitable
for NLP.

In conclusion, the grammar learner (in theory) is a solution to the
problem of undergeneration.  
Chapter six quantitatively evaluates just how well an implementation 
of the theory  meets
the success criteria.

\subsection{Completeness}
The super rules express all unary and binary rules.  As was shown, these are
sufficient to allow rules to be learnt that will generate any sentence.  
Furthermore, if the critic is always able to identify when a rule is
plausible, or otherwise, then that critic can be seen as a model-based
informant \cite{Gold67}.  Given such an informant, natural languages can be identified
in the limit.  Hence, if the model is complete, the learner is also
complete with respect to learnability. In practice, the model is usually
incomplete and so the learner will be incomplete.  This incompleteness can
be reduced by the contribution of the data-driven component.  

\subsection{Termination}
The learner will terminate when learning grammar for any sentence.  This
is because the learner does not acquire rules that allow cyclic derivation
sequences to be constructed.  For example, the learner does not create
unary rules that are recursive. Allowing cyclic derivations would lead to
parse trees that are infinitely deep.

\subsection{Complexity \label{complexity}}

Grammar learning takes at least exponential space with respect to 
sentence length.  This is because grammar learning implies that the
parser, in the worst case, 
 finds all binary and unary parses for any given sentence.  As is well known,
parsing with ambiguous grammars is of exponential space complexity.
For example, compound nouns are usually generated using a rule 
such as \psr{N}{N~N}.  This generates, with respect to sentence length, 
 a number of parses equal to
the Catalan series \cite{Chur82}.  To overcome this
intractability implies that the control strategy is made incomplete.
Interesting ways of introducing incompleteness
would be to introduce what Fodor, Bever and Garrett call {\it performance
strategies} \cite{Fodo74}.  Berwick comments that \cite[p.166]{Berw87}:
\begin{quotation}
Thus, if ordinary processing methods are tailored to the most difficult
situation, we can easily miss the forest for the trees -preparing for the
worst cases, but missing streamlined strategies that would work in the
usual case.
\end{quotation}
Performance strategies are
 also similar to the idea of {\it heuristic search} used in 
knowledge-based systems.  In the context of the control strategy, 
 incompleteness might be introduced by halting the learning process after
constructing $n$ parses, or halting after creating $m$ edges.  Both of these
resource bounds are used in chapter six.

The next chapter explains what is meant by model-based learning in this
thesis and shows how the critic can be used to learn plausible grammars.
The following chapter explains what is meant by data-driven learning 
and shows how data-driven learning  interacts with model-based learning.

\chapter{Model-Based Learning}

\section{Introduction}
The purpose of this chapter is to explain what a model of grammaticality
is, to show how a grammar can be extended with such a model, and finally, to
discuss characteristics of the model-based learner.  

\section{Model-based learning in machine learning and in linguistics}
Recently, machine learning researchers have looked at 
{\it model-based} methods of learning.    Instead of relying upon many 
training examples to
constrain the search for some generalisation, model-based methods constrain 
the search by using knowledge of the task domain.  This knowledge is called
a {\it model}.  After using the model to analyse an example, these methods
 produce a valid generalisation of the example, along with a 
deductive justification of the generalisation in terms of the 
model \cite{Mitc86}.  Because of this, 
 model-based methods are sound in the sense that
the generalisation acquired deductively follows from the model.  
In the context of grammar
learning, the training examples will be sentences, the model
will be a high-level theory of syntax, and the generalisation will
be a grammar generating those sentences.

  It could be argued that the model
could be compiled-out to produce any generalisation required without
recourse to training data.  However, such a compilation process  is usually
computationally far too expensive to perform without guidance from  
training examples.  For example, the grammar presented in chapter six
can express $621495189504^{3}$ binary rules, and $621495189504^{2}$ 
unary rules.\footnote{A unification-based grammar with $n$ features, where 
each feature $f_i$ has
$\mid v_i~\mid$ values, can express:
\begin{equation}
(\prod_{j=1}^{n}\nu_j)
\end{equation}
distinct categories.  This assumes that features do not contain categories
as values (in which case the space would be infinite).}  Clearly, all of these rules cannot be compiled-out and
tested against the model to determine if they are plausible.   Furthermore,
model incompleteness will mean that implausible rules will also be
compiled-out.  Therefore,
it is valid to say that 
model-based learning  involves learning, but  the learning of 
search heuristics over the
concept space defined by the model \cite{Lang87}, and not of learning
concepts themselves.    We shall consider what the linguistic
equivalent of the model is next.

Linguists interested in human language acquisition have proposed that 
successful language learning is possible if the human language learner has an
innate knowledge of language \cite[p.51]{Chom65}.  This
innate knowledge is usually called {\it Universal Grammar} 
(UG)  \cite[p.29]{Chom75}.  UG is motivated on the grounds that the training
data available to children underdetermines 
the final grammar (the training data is
too impoverished to explain why a full competence grammar is acquired),
the training data may contain performance errors (`noise') and finally
the data does not contain negative evidence (errors are not
corrected) \cite{Whit89}.  The last motivation is especially interesting,
given the popularity of using negative examples in machine learning
algorithms and the fact that, with negative examples, almost all languages
are learnable in the limit.  An example of this can be seen with 
the machine learning
algorithm  ID3  \cite{Quin86},  which learns a
classifier after being trained with positive and negative examples.  The 
author  
trained ID3 using 60 grammatical sentences, and 100 ungrammatical sentences.
After training, it was tested on 60 unseen sentences, and 100 unseen 
ungrammatical
strings.  The results showed that with negative examples, the classifier
could label all the ungrammatical test strings as being ungrammatical, 
and most of the test sentences as being grammatical.  When
trained without the negative (ungrammatical) 
examples, the classifier considered all 
of the test sentences and strings to be grammatically well-formed.  Negative 
examples are therefore a powerful source of grammatical information.  
However, as was stated, children are not told when they use ungrammatical
sentences and so this method of identifying languages 
in the limit is not used by
language acquisition theorists.  Instead, UG is usually formulated as a
{\it model of grammaticality}, consisting of a set of {\it principles}
and {\it parameters}.  Principles capture generalisations across
particular constructions and parameters set in the principles to operate 
in various ways.  An example principle of UG (called {\it subjacency} in
Government and Binding Theory (GB) \cite{Chom81})
 might  say that
a filler  can only be `moved' from a trace across a limited number of
constituent boundaries.  The parameter for such a principle would be the
set of constituent boundaries.  In English, this set includes S and NP.
For example, the following ungrammatical sentences can be explained as 
violations of this parameter setting for English \cite[p.23]{Whit89}:
\numberSentence{*What did Mary wonder whether John bought e}
\numberSentence{*What did Mary believe the claim that John saw e}
In the first case, the filler {\it what} has moved across two
S categories from {\it e} and in the second case, the filler
{\it what} has moved across two S categories and an NP category.
Language acquisition using UG  consists of `triggering' a
particular setting of these parameters.  Parameters are triggered by
the learner being exposed to a training set of sentences.  

Several linguistic theories are influenced by  UG considerations.  The 
most obvious
example is Government and Binding Theory (GB) \cite{Chom81}.  Other 
approaches, which are
less obviously motivated  by acquisition 
 theory, include Lexical Functional Grammar (LFG) \cite{Bres78}, 
Generalised
Phrase Structure Grammar (GPSG) \cite{gkps85} and 
Head Driven Phrase Structure
 Grammar (HPSG) \cite{Poll87}.\footnote{Fodor 
has shown how GPSG can be seen in terms of
language acquisition from a 
principles and parameters perspective \cite{Fodo90}.} 
 An implementation of a model of grammaticality would
therefore draw upon these linguistic theories.

The model-based component of the critic (of the Grammar Garden)
draws upon  GPSG's realisation of UG.
Most of the parameters within the model 
are already set in advance 
(to expect English).   The parameters  set during learning 
are the individual phrase 
structure rules (which Fodor considers to be the parameters of GPSG).  
Therefore, because the model happens to 
draw largely upon GPSG, it would be true to say that the grammar
`learnt' is a GPSG grammar.  However, the system is not limited just to 
learning GPSG-style grammars.  With some work, the model could use principles 
drawn from other linguistic theories.  For example, the system could use
subjacency.  This would then mean that a GPSG grammar is not being learnt.
Model-based learning is capable of learning any style of grammar, so long as
the relevant principles can be formalised appropriately.

The rest of this chapter is as follows.  Section \ref{gram} presents 
a unification-based grammar $G$.  Section \ref{model} describes the 
components of the model of grammaticality.  Section \ref{modelExamples} 
then gives various examples of grammar $G$ being extended using model-based
learning.  Finally, section \ref{modelConc} is a concluding discussion.

\section{The initial grammar and lexicon \label{gram}}
The following grammar 
is adapted from the example grammar
``ex-sem'' given 
in the third release of the Alvey Tools Grammar and is intended to
be  for demonstrational purposes.  That is, the exact features
structures used are unimportant.  Note that the grammar contains
a rule (PP) that cannot be used in any derivation sequence constructed
using $G$.  This rule is intentionally in $G$ and is used in one of the 
learning examples.

The features of the categories  
are:
\begin{center}
\small
\begin{tabular}{|l|l| } \hline 
Feature & Values \\ \hline
N & + - \\
V & + - \\
BAR & 0 1 2 3 \\
DET & + - \\
PER & 1 2 3 \\
PLU & + - \\
PRD & + - \\
NTYPE & NAME PRO COUNT MASS \\
DEF & + - \\
SUBCAT & INTRANS TRANS DITRANS \\
PAST & + - \\
CASE & NOM ACC \\
VFORM & FIN PASS BSE \\
ADV & + - \\
PFORM & TO BY  \\
AUX & DO BE - \\
INV & + -  \\
NULL & + - \\
EMPTY & + - \\
CONJ & AND BOTH BUT NEITHER NOR OR NULL \\ \hline
\end{tabular}
\end{center}
The rules are:
\labelRule{\psr{\fs{N - \\ V + \\ BAR 2 \\ DET - \\
	  		   PER \fbox{1} \\ PLU \fbox{2} \\PAST \fbox{3} \\
		           VFORM FIN \\ AUX \fbox{4} \\ INV \fbox{5}}}
		      {\fs{N + \\ V - \\ BAR 2 \\ DET - \\ PLU \fbox{2} \\
			   PRD - \\ CASE Nom  \\ PER \fbox{1} }~
		      \fs{N - \\ V + \\ BAR 1 \\ DET -  \\ 
			  PER \fbox{1} \\ PLU \fbox{2} \\ PAST \fbox{3} \\
                           VFORM FIN \\ AUX \fbox{4} \\ INV \fbox{5}}}}{S1}
(paraphrase: \psr{S}{NP~VP})
\labelRule{\psr{\fs{N - \\ V + \\ BAR 1 \\ DET -  \\ 
			  PER \fbox{1} \\ PLU \fbox{2} \\ PAST \fbox{3} \\
                           VFORM \fbox{6} \\ AUX \fbox{4} \\ INV \fbox{5}}}
	     {\fs{N - \\ V + \\ BAR 0 \\ DET - \\ PER \fbox{1} \\ 
	      PLU \fbox{2} \\ SUBCAT Intrans \\ PAST \fbox{3} \\
	      VFORM \fbox{6} \\ AUX \fbox{4} \\ INV \fbox{5} \\ Conj Null }}}{VP1}
(paraphrase: \psr{VP}{V0})
\labelRule{\psr{\fs{N - \\ V + \\ BAR 1 \\ DET -  \\ 
			  PER \fbox{1} \\ PLU \fbox{2} \\ PAST \fbox{3} \\
                           VFORM \fbox{6} \\ AUX \fbox{4} \\ INV \fbox{5}}}
	     {\fs{N - \\ V + \\ BAR 0 \\ DET - \\ PER \fbox{1} \\ 
	      PLU \fbox{2} \\ SUBCAT Trans \\ PAST \fbox{3} \\
	      VFORM \fbox{6} \\ AUX \fbox{4} \\ INV \fbox{5} \\ Conj Null }~
	     \fs{N + \\ V - \\ BAR 2 \\  DET - \\ PRD -  \\ CASE Acc}}}{VP2}
(paraphrase: \psr{VP}{V0~NP})
\labelRule{\psr{\fs{N + \\ V - \\ BAR 2 \\ DET -  \\ 
			  PER \fbox{1}}}
	     {\fs{DET + \\ Conj Null }~
	     \fs{N + \\ V - \\ BAR 1 \\  DET - \\ PER \fbox{1}}}}{NP1}
(paraphrase: \psr{NP}{Det~N1})
\labelRule{ \psr{\fs{N + \\ V - \\ BAR 1 \\ DET -  \\ 
			  PER \fbox{1} \\ PLU \fbox{2} \\ PRD \fbox{3} \\
		Ntype \fbox{4} \\ CASE \fbox{5}}}
	{\fs{N + \\ V - \\ BAR 0 \\ DET -  \\
                          PER \fbox{1} \\ PLU \fbox{2} \\ PRD \fbox{3} \\
                Ntype \fbox{4} \\ CASE \fbox{5} \\ Conj Null}}}{N1}
(paraphrase: \psr{N1}{N0})
\labelRule{\psr{ \fs{N - \\ V - \\ BAR 2 \\ DET -}}{\fs{N - \\ V - \\ BAR 0 \\ DET - \\ Conj Null }
	~\fs{N + \\ V - \\ BAR 2 \\ DET -}}}{PP}
(paraphrase: \psr{PP}{P~NP}) \\
The lexical entries are:
\begin{center} \small
\begin{tabular}{l}
\lexical{\it cat}{\fs{N +, V -, BAR 1, DET -, PER 3, PLU -, NTYPE COUNT}} \\
\lexical{{\it chases}}{ \fs{N -, V +, BAR 0, DET -, PER 3, PLU -, SUBCAT TRANS \\ PAST -, VFORM FIN,
      AUX -, INV -, CONJ NULL}}\\
\lexical{{\it down}}{ \fs{N -, V -, BAR 0, DET -, SUBCAT NP, CONJ NULL}} \\
\lexical{{\it happy}}{\fs{N +, V +, BAR 1, DET -, ADV -}} \\
\lexical{{\it road}}{\fs{N +, V -, BAR 0, DET -, PER 3, PLU -, NTYPE COUNT, CONJ NULL}}\\
\lexical{{\it Sam}}{\fs{N +, V -, BAR 2, DET -, PER 3, PLU -, PRD -, NTYPE NAME}}\\
\lexical{{\it the}}{\fs{DET +,  DEF +, CONJ NULL}} \\
\end{tabular}
\end{center}
Note that the entries with BAR levels of one arguably should have a BAR 
level of zero.  However, having a BAR level of one simplifies the 
exposition.  One could interpret BAR one categories in the lexicon as
representing a BAR level one category re-written implicitly by a rule that
raises the BAR level by one.  Note also, the entry for the word {\it happy} cannot
be generated by the grammar $G$.  Again, this is intentional and
 rules will be learnt that enable
$G$ to generate sentences that contain this word.
\section{The model \label{model}}

The model consists of the set of principles $P_1 P_2 \ldots P_n$. Each 
principle $P_i$ is a predicate over a rule $\alpha$ implicitly used in an edge
 and is true if $P(\alpha)$ cannot be proved to be false:
\begin{eqnarray}
\bigwedge_{i=1}^n P_i(\alpha)
\end{eqnarray}
If some 
$P_j$ cannot prove $\alpha$ to be implausible, $P_j(\alpha)$ is
judged as being plausible.   This is motivated on the
grounds that, in practise, the principles will be incomplete with respect to
being able to prove if $\alpha$ is linguistically plausible, or otherwise.
However, the data-driven component of the critic, which is explained in the
next chapter, is complete, and hence can deal with some $\alpha$ passed-on
by the model-based component of the critic.  If the principles rejected 
$\alpha$ because  they failed to  prove it to be linguistically plausible, 
the data-driven component of the learner would not have a chance to compensate
for incompleteness in the model of grammaticality.  The actual choice of
the principle set depends upon decisions such as 
capturing generalisations that are thought to hold
across languages.  In this thesis, the choice of principles is
ad hoc, and not supposed to be a statement about being  the `best'
such set.   The principle set in this thesis consists of:
\begin{center}
\begin{itemize}
\item Linear precedence rules.
\item Types.
\item Feature-passing conventions
\end{itemize}
\end{center}
These are all drawn from GPSG.  Other principles, from other linguistic 
theories, could be used.  Indeed, an extension of this work, as
discussed in the final chapter,  would be to
extend the principle set.  The system is so designed that the principle set
used in learning can be any combination of  these principles 
(allowing flexibility in experimentation).  There
are other, inbuilt principles used in the learner, such as X-bar syntax.  However,
these cannot be varied in the same manner.  For example, learning without
X-Bar syntax would result in very low quality rules being constructed.
Interestingly enough, Gazdar {\it et al} argue for inbuilt principles, and
not for variable principles \cite[p.3]{gkps85}.  
From an experimental perspective however, it is preferable to have
principles that can be varied.  Like axioms, they can be changed, and the
effects then noted.

The principles that can be varied will now be explained in turn.  

\subsection{Linear precedence rules}

A PSG may contain rules such \psr{A}{B~C} and \psr{A}{B~D} which are used
to generate trees such as (A (B C)) and (A (B D)).
Within the tree (A (B C)), the daughter B is said to {\it linearly precede}
daughter C.  Likewise in tree (A (B D)), daughter B linearly precedes
daughter D.  A linear precedence (LP) constraint 
between two daughters A and B is 
described by  the 
notation $A \prec B$.  This means that B cannot linearly precede A in a
tree and hence a rule cannot be constructed that would license such a
tree.  Given such a  LP rule,  rules such as  \psr{A}{B~A} would be ill-formed.
LP rules have a history in linguistics (described   by 
Gazdar {\it et al} \cite[p.47]{gkps85}) and are used to express
intra-language generalisations.  For example, within GPSG, lexical categories
subcategorise, whilst phrasal categories do not subcategorise and,  according 
to GPSG,  lexical categories linearly 
precede phrasal categories in rules.  The LP
rule:
\begin{eqnarray*}
[\mbox{SUBCAT}] \prec~ \sim [\mbox{SUBCAT}]
\end{eqnarray*}
expresses this fact.  For this rule to be violated, a daughter must both
not subcategorise and also linearly precede a subcategorising daughter.
For example, such an LP rule would be satisfied by a local tree 
(NP (DET the) (N1 cat)), but not by the local tree (NP (N1 cat) (DET the)).

The model contains the following LP rules:
\begin{center}
\begin{tabular}{|l|l|} \hline
Rule name & rule \\ \hline
LP1 & $[\mbox{SUBCAT}] \prec~ \sim [\mbox{SUBCAT}]$ \\
LP2 & \fs{N +} $\prec$ \fs{N -, V -, BAR 2} \\
LP3 & \fs{N +} $\prec$ \fs{N -, V +, BAR 2} \\
LP4 &   \fs{N -, V -, BAR 2} $\prec$  \fs{N -, V +, BAR 2} \\ \hline 
\end{tabular}
\end{center}
LP1 has been previously described.  LP2 says that prepositional phrases
follow nominal, adjectival, or adverbial  phrases.  For example, the sentence:
\numberSentence{*the in the park boy laughed}
is ungrammatical as a prepositional phrase cannot precede a noun.
LP3 says that nominal, adjectival, or adverbial phrases 
precede VPs or sentences.  This accounts for the
ungrammaticality of sentences such as :
\numberSentence{* laughed in the park the boy}
LP4 says that PPs precede VPs or sentences, but not the other way around.  
\subsection{Types}
As well as restricting the linear ordering of categories in trees (and hence
restricting the ordering of categories in rules), it is also useful to
restrict the co-occurrence of categories within rules.  For example, one may
want to state that determiners can co-occur in the RHS of a rule with a
nominal, adverbial, or adjectival  category, but not with a NP:
\numberSentence{Sam chases the cat}
\numberSentence{*Sam chases the Sam}
Although this is possible to achieve using LP rules:
\begin{center}
\begin{tabular}{l} 
\fs{DET +}  $\prec$ \fs{N +, V -, BAR 1} \\
\fs{DET +}  $\prec$ \fs{N +, V -, BAR 2} \\
 \fs{N +, V -, BAR 1}  $\prec$  \fs{DET +} \\
\end{tabular}
\end{center}
(that is, the first of these LP rules allows determiners to combine with a
nominal category, the second LP rule 
says that determiners cannot be preceded by a
NP, whilst the third LP rule says that NPs cannot be preceded 
by a determiner.)  it is also possible  to state this co-occurrence more
compactly by associating {\it types}\footnote{Types in computational
linguistics have a primary use in semantic interpretation. Enforcing 
co-occurrence restrictions using types is the way we use them.}  with each 
syntactic category and 
determining if these types can be (functionally) applied together.  This is
similar to type checking in programming languages.  To perform this checking 
requires a language to express these types \cite[p.88]{Dowt81}.  Such a
language is the extensionally typed lambda calculus \cite[p.89]{Dowt81}, 
which is defined
as follows:
\begin{center}
\begin{itemize}
\item {\it e} is a type.
\item {\it t} is a type.
\item If {\it a} and {\it b} are types, then $\langle a,b \rangle$ is a type.
\item Nothing else is a type.
\end{itemize}
\end{center}
Drawing upon GPSG, this typed language ``represents the semantic role
of the various syntactic categories in the grammar'' \cite[p.185]{gkps85}.
Hence, a failure to apply the types corresponding to some pair of syntactic
categories implies that these syntactic categories cannot co-occur in the
same rule.  For example, if the type of a determiner is:
\begin{center}
$\langle \langle e, t \rangle, \langle \langle e, t \rangle, t \rangle \rangle$
\end{center}
the type of a nominal category is:
\begin{center}
$\langle e, t \rangle$
\end{center}
and the type of a NP is:
\begin{center}
$\langle \langle e, t \rangle , t \rangle$
\end{center}
then the type of determiners can be applied with the type of 
nominal categories:
\begin{center}
$\langle \langle e, t \rangle, \langle \langle e, t \rangle, t \rangle \rangle 
\circ \langle e, t \rangle =  \langle \langle e, t \rangle , t \rangle$
\end{center}
(where $\circ$ represents the functional application operator).  However, 
the type of determiners cannot be applied to the type of NPs.  Hence,
given just these types, it is possible to restrict the co-occurrence of
determiners.  This is achieved more compactly than when using the previous
LP rules.  Note that types complement, and do not replace, 
 the LP rules in that linear
precedence cannot be enforced with type checking.

Type checking is implemented as follows.  Let $TYP$ be a partial function
from feature structures to types.  If $A$ and $B$ are feature structures
within the RHS of a rule, then if 
\begin{center}
$TYP(A) \circ TYP(B)$
\end{center}
is defined, then categories $A$ and $B$ can co-occur within the RHS of a 
rule.

The set of category-type pairs extensionally defining the $TYP$
function  in the model are:
\begin{center}
\begin{tabular}{|l|l|} \hline 
category & type \\ \hline
\fs{V +, N -, BAR 2, DET -} & $t$ \\
\fs{V +, N -, BAR 1, DET -} & $\langle \langle \langle e, t \rangle ,t 
\rangle ,t \rangle$ \\
\fs{V +, N -, BAR 0, DET -, SUBCAT INTRANS} & $\langle \langle \langle e, t \rangle , t \rangle , t \rangle$ \\
\fs{V +, N - , BAR 0, DET -, SUBCAT TRANS} & $\langle \langle \langle e, t \rangle , t \rangle ,\langle \langle \langle e, t \rangle ,t \rangle , t \rangle  \rangle$ \\
\fs{N +, V -, BAR 2, DET -, PRD -} & $ \langle \langle e, t \rangle ,t
\rangle$ \\
\fs{N +, V -, BAR 1, DET -} & $\langle e, t \rangle$ \\
\fs{N +, V -, BAR 0, DET -} & $\langle e, t \rangle$ \\
\fs{N +, V +, DET -} & $\langle \langle e, t \rangle,  \langle e, t \rangle \rangle$ \\
\fs{DET +} & $\langle \langle e, t \rangle, \langle \langle e, t \rangle, t \rangle \rangle$ \\ \hline
\end{tabular}
\end{center}
These types are for demonstrational purposes only and
not meant to be taken as a formal claim of the co-occurrence patterns of 
categories in rules for English syntax.  

In practice, within the system, not every syntactic category will 
have a corresponding
type (due to incompleteness) and hence, if $TYP(A)$ is undefined, then 
$TYP(A) \circ TYP(B)$ is taken as being defined.  It would then be the task
of other aspects of the model to reject a super rule instantiation that 
satisified type checking through incompleteness.
\subsection{ Feature-passing conventions \label{hfc}}

Many of the features of a grammar are {\it head features}, whose distribution
involves the head of a rule.  For example, a local tree generated using a 
rule with a nominal head would have a mother containing the same PLU feature
as the nominal head daughter.  This is expressed using a {\it head feature
convention} (HFC) \cite[p.94]{gkps85}, which places a restriction upon local trees 
as follows:
\begin{eqnarray}
\phi (C_0) \mid \mbox{{\it Head}} = \phi (C_H) \mid \mbox{{\it Head}} 
\end{eqnarray}
$C_0$ is the LHS category of some rule, $C_H$ is the head category of that
rule, $\phi$ is a mapping from categories in rules to nodes in trees, 
$\mid$ is function restriction and {\it Head} is  the set of head features.
Note that this is the simplest possible head feature convention.  There
are many other, more realistic conventions possible (for example those involving
multiple heads).  However, this  HFC is sufficient for demonstrational
purposes.  

Within the model of grammaticality, the head features are all the previously
mentioned features  (in \S \ref{gram}) except for the set of 
features 
\begin{eqnarray*}
\{ \mbox{NTYPE, CASE, CONJ, NULL, BAR} \}
\end{eqnarray*}
  Features not in the set
of head features do not obey the HFC.  In particular, as outlined
in \S \ref{constructor}, the BAR feature 
receives a special treatment. 

 The HFC is implemented as follows.  Instead
of being a distinct device from the grammar (as in GPSG and HPSG) and 
applying the HFC to local trees, the HFC is compiled directly into rules.
Those features that are in the head set are shared between the LHS and
the head category, whilst those features that are not 
in the head set (with the exception of the
BAR feature) are removed from the LHS.  Compiling the HFC into rules is
an efficiency measure and is used by other systems, such as the Grammar
Development Environment (GDE) \cite{Carr91}.  A rule
obeys the HFC if the head features are shared between the head and the 
projection and the non-head features are not shared between the head and the
projection.  For example, the rule:
\begin{center}
\psr{\fs{N + \\ V - \\ BAR 2}}~{\fs{N - \\ V + \\ BAR 1}}
\end{center}
violates the HFC as the N feature is not shared between the head and the
projection.  

This concludes the description of the model of grammaticality.  The next
section shows how this model can be used to learn plausible rules.  
\section{Model-based rule learning \label{modelExamples}}

This section demonstrates the capabilities of model-based learning by 
firstly re-presenting the learning example given in chapter three, 
 showing the effect of gradually increasing the extent of
the model's coverage, secondly by giving an
example of ambiguous attachment, thirdly by showing how 
ungrammatical sentences can be detected, and finally 
by presenting an example
where both unary and binary rules are acquired.  

Suppose the Grammar Garden's parser 
tried to parse the sentence:
\numberSentence{Sam chases the happy cat}
using just the initial grammar and with the learning facility turned off:
\begin{verbatim}
: Entering Grammar Garden Parser  ... (level 2)
6 Parse+>> Sam chases the happy cat
0 parse(s)
850 msec CPU, 1000 msec elapsed
\end{verbatim}
As the rule \psr{N1}{AdjP~N1} is missing from the initial grammar, 
the sentence cannot be parsed.  We therefore need to deal with this.

Now, if the sentence is reparsed using just the binary super rule, without
model-based learning, then we have the following result:
\begin{verbatim}
21 Parse+>> Sam chases the happy cat
learning
15 rule(s) acquired.
244 parse(s)
91863690 msec CPU, 98998000 msec elapsed
\end{verbatim}
No instantiations of the super rule are rejected by the model
(since the model of the critic has been turned-off), and 
  the system acquires
15 rules, which is presumably more than necessary.  Using atomic symbols for
feature structures,\footnote{V1 stands for a bar one verbal feature structure
and V0 for a bar zero verbal feature structure.} an example spurious 
rule learnt is \psr{\{V1, V0\}}{V0~Det}.
This rule is both implausible, and also overgenerates.  For example, it
would allow the ungrammatical sentence:
\numberSentence{*Sam chases the the the cat}
to be generated.

If the sentence is reparsed as before,  using a model consisting of
just  the LP rules,
then we have a reduction in the number of rules acquired:
\begin{verbatim}
39 Parse+>> Sam chases the happy cat
learning
9 rule(s) acquired.
85 parse(s)
11777140 msec CPU, 12798000 msec elapsed
\end{verbatim}
Similarly, if the sentence is re-parsed as before but 
using a model consisting of  just the semantic types, 
an even greater reduction in rules acquired occurs:
\begin{verbatim}
45 Parse+>> Sam chases the happy cat
learning
6 rule(s) acquired.
8 parse(s)
165420 msec CPU, 181000 msec elapsed
\end{verbatim}
In both cases, the model is not constraining enough and implausible 
rules are acquired.  If both LP rules and semantic types 
are used together,  the following result is found:
\begin{verbatim}
29 Parse+>> Sam chases the happy cat
learning 
1 rule(s) acquired.
1 parse(s) 
5940 msec CPU, 7000 msec elapsed
\end{verbatim}
The parse generated is:
\begin{verbatim}
30 Parse+>> !*parses*
((``S1''
  ((|Sam|)
   (``VP''
    ((|chases|) 
    (``NP1'' ((|the|) 
             (``*binary25158'' ((|happy|) 
                               (|cat|))))))))))
\end{verbatim}
The names of the rules used in the parse  label the
nodes of the tree and the newly acquired rule (*binary25158) in full is:
\begin{eqnarray*}
\left\{ \fs{ N + \\ V - \\ BAR \{1,2\} \\ DET -  \\PER 3  \\ PLU - \\  NTYPE COUNT
	},
	\fs{N  + \\ V  + \\ BAR \{1,2\} \\ DET  - \\ ADV  -} \right\} \rightarrow 
	 \fs{N  + \\ V  + \\ BAR  1 \\ DET  - \\ ADV  -}
	 \fs{N + \\ V - \\ BAR \{1,2\} \\ DET -  \\PER 3  \\ PLU - \\  NTYPE  COUNT }
\end{eqnarray*}
This is paraphrased into an atomic phrase structure rule as:
\begin{center}
$\{AP,NP,Adj,N1\} \rightarrow Adj~N1$
\end{center}

This rule leads to overgeneration as the model cannot refine the disjunctive
LHS into a single non-disjunctive category.  Chapter five shows how data
driven learning can refine such a rule.
Note  that the search space explored (as roughly measured by the time elapsed)
is much less than that of the previous examples. 

If the previous sentence is reparsed as before, using both LP
rules and semantic types, but this time also
using the Head Feature Convention, then the rule acquired has a 
slightly different LHS:
\begin{eqnarray*}
\left\{ \fs{ N + \\ V - \\ BAR \{1,2\} \\ DET -  \\PER 3  \\ PLU -},
	\fs{N  + \\ V  + \\ BAR \{1,2\} \\ DET  - \\ ADV  -} \right\} \rightarrow 
	 \fs{N  + \\ V  + \\ BAR  1 \\ DET  - \\ ADV  -}
	 \fs{N + \\ V - \\ BAR \{1,2\} \\ DET -  \\PER 3  \\ PLU - \\  NTYPE  COUNT }
\end{eqnarray*}
The NTYPE feature is not in the set of head features, and hence it does not
obey the HFC and so no longer appears in the rule's LHS.

The system is capable of learning ambiguous attachments.  For example,
prepositional phrases are traditionally thought to attach ambiguously.  
The original grammar $G$ lacks rules to attach the rule PP and these rules
will need to be learnt.  If the 
system tries to learn rules for the  sentence:
\numberSentence{Sam chases the cat down the road}
 using the full model (LP rules, 
types, and the HFC), the binary super rule and the original
grammar (which contains no rules for PP-attachment), the following
result occurs:
\begin{verbatim}
17 Parse+>> Sam chases the cat down the road
learning
8 rule(s) acquired.
13 parse(s)
189290 msec CPU, 226000 msec elapsed
\end{verbatim}
One of these rules attaches the PP to the N1, another to the 
NP, and another to the  VP.  The PP is not
 attached to the S due to LP4 (which says that prepositional phrases do not
attach to sentences).  The other 5 rules\footnote{Some of these
implausible rules are shown on page \pageref{ppProblem}.} are implausible and
caused by   the model being
underconstraining.    Chapter five shows
how model incompleteness can be overcome inductively.  Note that if
the original grammar had a rule that attached PPs, then no undergeneration
would take place.  Consequently, other ambiguous attachments would not
be learnt.  Likewise, if the system learnt about PP-attachment in the
context of a sentence such as:
\numberSentence{Sam   chases down the road the cat}
then the system would not be able to learn about PP-attachment to NPs.  
This therefore shows that for learning about ambiguity, the system
is dependent upon the original grammar and 
the ordering of the training material.

An advantage of model-based learning is that ungrammatical
sentences 
can be recognised and not lead to a performance grammar being
acquired.  For example, using the full model and the binary super rule:
\begin{verbatim}
52 Parse+>> Sam chases happy the cat
learning
0 parse(s)
3010 msec CPU, 3000 msec elapsed
\end{verbatim}
No rules are acquired (due to semantic 
type checking).  Model incompleteness will, however, 
lead to some ungrammatical  sentences  having parses using learnt rules that 
the model could not reject:
\begin{verbatim}
68 Parse+>> Sam chases the cat happy
learning
1 rule(s) acquired.
1 parse(s)
69 Parse+>> !*parses*
((``S1''
  ((|Sam|)
   (``VP''
    ((|chases|) 
    (``NP1'' ((|the|) 
             (``*binary12888'' ((|cat|) 
                               (|happy|))))))))))

\end{verbatim}
This implausible rule ($\{N1,NP, Adj, AP \} \rightarrow N1~Adj$)
 is learnt due to a lack of linear precedence information regarding
the ordering of adjectives and nominals.

Throughout this section, all of the examples used just the binary super
rule.  The system is also capable of learning unary rules.  Here the 
model is as before (LP rules, types and Head Feature Convention), but this
time both binary and unary super rules are used:
\begin{verbatim}
83 Parse+>> the happy cat
learning
8 rule(s) acquired.
15 parse(s)
318060 msec CPU, 377000 msec elapsed
\end{verbatim}
As expected, the search space has greatly expanded.  
Most of the extra rules learnt 
are chains of unary rules, up to the maximum bar level in depth (as, for 
example \psr{AP}{Adj}, \psr{A3}{AP}).  
When using just the binary super rule (with the same learning
configuration as before), only the single rule is learnt:
\begin{verbatim}
10 Parse+>> the happy cat
learning
1 rule(s) acquired.
1 parse(s)
1950 msec CPU, 3000 msec elapsed
11 Parse+>> !*parses*
((``NP1'' ((|the|) (``*binary3072'' ((|happy|) (|cat|))))))
\end{verbatim}
This is the same  as rule *binary25158 previously learnt.
\section{Discussion \label{modelConc}}

This chapter explained what a model of grammaticality is, and went on
to show how a grammar could be extended using such a model.  There are
several advantages to using a model when learning grammar:
\begin{center}
\begin{itemize}
\item Plausible rules can be learnt.
\item Ungrammatical strings can be recognised for what they are.
\item Expressing linguistic generalisations concisely within a model
removes the need to discover such generalisations inductively, using 
large quantities of data.
\item Languages can be identified in the limit using a model.
\end{itemize}
\end{center}
We saw from the examples how plausible rules can be learnt.  As well as
helping acquire plausible rules, the model also supplied a justification
of the rules.  That is, each rule learnt satisfied the model.  Inductive
approaches provide no such support for the existence of any given rule.

Linked with the previous point is the fact that if the model rejects
all of the rules for a string,  that string is ungrammatical.  The model
justifies the rejection of a string and this justification can then
serve as the basis for string
 correction.  Hence, the model doubles-up as both a yardstick of grammaticality, and also as a yardstick of 
ungrammaticality.

Model-based learning is knowledge-intensive, and not data-intensive.  This
means that large quantities of data are not necessary to learn a 
grammar using a model-based learner.  Hence, all things being equal, 
a model-based learner will learn a `reasonable' grammar faster than
a data-driven learner.  

As has been previously mentioned, a model of grammaticality acts as
an informant and enables the learner to eventually learn the language
in question.  Inductive approaches, without an informant, 
 cannot identify the language and
hence will always be limited solutions to the problem of language learning.

Various problems face model-based learners:
\begin{center}
\begin{itemize}
\item Model incompleteness undermines learning.
\item The model may be inconsistent and hence may lead to an implausible
grammar being learnt.
\item The model may contain intractable principles that take exponential or
more time.
\end{itemize}
\end{center}

We saw in this chapter how model incompleteness undermines learning.  That
is, the system learnt more rules than was necessary for the PP-example and
acquired implausible rules for an ungrammatical sentence.  In the next
chapter a solution to the problem of 
 model incompleteness is considered.

Ensuring consistency faces any designer of a 
model-based grammar learners.  For
example,   Fong found that his (GB-based) model of 
grammaticality would sometimes unexpectedly fail to prove a previously 
proved sentence after a principle was changed \cite[p.209]{Fong91}.  
There is no absolute solution to this problem, other than using
intuition and hoping that testing uncovers inconsistencies.

Intractable principles can be dealt with  either 
 by placing a resource bound upon the 
principle, or by removing the principle from the model.  For example, Fong 
 found that some GB principles of grammaticality (such as VP-adjunction) 
led to parser non-termination \cite[p.49]{Fong91}.  He placed resource
bounds upon these principles, thereby introducing incompleteness into his
model.  Within the Grammar Garden,  either solution to the 
intractable principle problem can be adopted.  The
data-driven learner would then be expected to compensate for the
resulting incompleteness.

In the next chapter, data-driven learning will be presented.

\chapter{Data-Driven Learning}

\section{Introduction}
This chapter  explains what is meant by {\it data-driven}
(or inductive, or stochastic) grammar learning, presents an example of a 
data-driven grammar learner, demonstrates some of the
characteristics of this learner, and finally discusses data-driven
grammar learning in general.  

\section{Data-driven learning in machine learning and in linguistics}
Induction has long been a popular learning method in machine 
learning (for example, \cite{Wins75,Mitc78,Mich80,Shap83,Quin86,Quin89}).
Broadly speaking, when sufficient instances of some concept have been seen,
a concept description can then be constructed.  The concept description
can then be used to classify other instances.  For example, having
observed the set of instances $\{a, aa, aaa \}$, the learner might construct
the concept (regular expression) $a^n$.  Clearly, this describes the instances,
along with other, unseen instances, such as $aaaa$.  Inductive 
approaches  have also
been used in linguistics.  Harris \cite{Harr51} and others developed procedures
to carry out 
``distributional analysis''; the procedure,  when applied to a corpus, 
 derived the rules of grammar from the corpus \cite[p.157]{Lyon68}.  This line of research fell by the wayside.  However, more
recently, speech recognition researchers have re-adopted inductive
grammar construction approaches \cite[p.13]{Mage94}.   These 
workers  base their work on 
Shannon's {\it Noisy Channel Model} \cite{Shan48,Chur93}.  This can be
described as follows.  
Suppose
a sequence
of objects  P has been corrupted in some manner, giving the sequence
of objects W.  The most likely sequence of objects $\rho$ can be recovered
from $W$ by hypothesising all possible sequences of P and selecting the
sequence of P that is most likely.  That is,
\begin{eqnarray}
\rho = \mbox{argmax}_{P}Pr(P) Pr(W \mid P)
\end{eqnarray}
$Pr(P)$ is the probability that $P$ will be present in the channel and
$Pr(W \mid P)$ is the probability of $W$ given $P$.  Argmax finds the
argument with the maximal score.     Channel characterisation
is in reality far too difficult
to achieve correctly and so channel approximations 
such as {\it N}-grams and Stochastic
Context Free Grammars are used instead to 
determine the most likely sequence of events.  These 
approximations are inductively
constructed (which constitutes the learning aspect of
using the Noisy Channel approach),  using training material of some form.

Corpus linguists (for example \cite{Gars87,Shar88,Haig88,Sout90,Blac93}),
have, following the success of the speech recognition 
community's use of the Noisy Channel Model, also adopted this model.
A popular reformulation of the model is to use a {\it treebank} and
a matching algorithm as an approximation of the channel \cite{Leec91}.  A 
treebank is
a set of parses for some corpus.  This reformulation, borrowing 
Magerman's terminology, is called {\it treebank 
recognition} \cite[p.59]{Mage94}.  In treebank
recognition, broadly speaking, the task is to 
find a parse $\rho$ for some sentence that is 
sufficiently close to some parse $\phi$ in the treebank.  Alternatively, 
putative parses for some sentence are compared against the treebank, and
the putative parse that is closest to some  parse in the treebank is
deemed to be the `correct' parse.  Because the matching algorithm  
implements a 
total function, any parse can be compared against the treebank.  Hence, 
treebank recognition is `complete': any sentence, no matter how
badly formed, will have a closest parse in the treebank. Accepting 
a parse that is not  present in the treebank 
 makes an inductive leap.  Given this 
characteristic,  a grammar learner could  
 `read-off' the newly 
constructed rules present in the parse that is closest to the 
parses in the treebank.  This is the approach outlined in the next
section.

\section{A data-driven grammar learner}
Here, an example of a data-driven grammar learner, similar in style 
to those used by corpus linguists, is presented.  This is not intended
to be the final statement on inductive grammar learning. Instead, it is
demonstrative of inductive grammar learners and intended to show how
such learners might
benefit from model-based learning.

The first task is to construct the treebank and the second task is
to create a tree matching algorithm.  

Treebanks can be created either manually, or 
automatically \cite{Samp91,Sout92}.  Manual treebank
construction is very labour intensive and error prone.  Automatic 
treebank creation, which does not suffer from these logistical problems, 
is preferable.  Treebanks can be automatically created by selecting a 
corpus, parsing that corpus, and then recording the parses produced. 
To be successful, the grammar used in treebank creation
should be able to generate all of the corpus.  However, undergeneration will
undermine automatic treebank creation.  One way of overcoming the problem of 
undergeneration would be to record, as well as complete parses,
 local trees produced for sentences in the corpus.  Another way of
automatically creating a treebank would be to run a grammar in reverse and 
generate the corpus artificially.\footnote{The ARK Corpus was
constructed in this manner \cite{Sout92}.}  An obvious 
problem with this approach
is that the frequencies of various constructs would be unnatural, and hence
would not be representative of any natural language.  

In this data-driven 
learner, the treebank is created by parsing a corpus and then recording all
local trees generated.  We are not interested in dealing with
syntactic ambiguity and so there is no need to have a parse selection mechanism
that choses the `best' parse for any given sentence in the corpus.  
Following the example of Leech \cite{Leec87,Leec91}, we 
decompose the treebank into a set of
mother-daughter-frequency triples.  This is designed to compress the
treebank and make it more manageable.  The mother of the triple is the
root of a local tree, the daughter is a category immediately dominated by
the mother, and the frequency is the total number of times that that
mother-daughter pair have been seen in the treebank.  For example, 
 a treebank consisting of the two parses:
\begin{center}
 (S (NP Sam) (VP (V laughs)))
\end{center}
 and 
\begin{center}
(S (NP Sam ) (VP (V chases) (NP (Det the) (N cat))))
\end{center}
 would be decomposed into
the set of triples:
\begin{eqnarray*}
\langle \mbox{S},\mbox{NP},2 \rangle \\
\langle \mbox{S},\mbox{VP},2 \rangle \\
\langle \mbox{VP},\mbox{V},2 \rangle \\
\langle \mbox{VP},\mbox{NP},1 \rangle \\
\langle \mbox{NP},\mbox{Det},1 \rangle \\
\langle \mbox{NP},\mbox{N},1 \rangle 
\end{eqnarray*}
Note that decomposing a treebank in this manner throws away some structural
information.  Also, triples are not produced for local trees immediately 
dominating lexical material.  Adding such triples would introduce problems
of sparse statistics (lexical items occur less frequently than syntactic
categories).  When using a unification-based grammar to create the treebank,
the test to determine the frequencies of the triples is category equality.
Category $D$ is equal to category $D'$ if $D \sqsubseteq D'$
and $D' \sqsubseteq D$.

After creating a set of triples, the next task is to create the matching
algorithm.  This should give frequently seen trees in the treebank a high
score, infrequently seen trees a low score, and unobserved trees a small
score.  By giving unobserved trees a small score, the matching algorithm
makes the ergodic assumption,\footnote{The ergodic assumption, roughly speaking,
is that all events have a non-zero probability of
occurring.  Hence, all events might be seen, and none are ruled-out.} and hence is complete.  The algorithm should
also allow frequently observed trees in the treebank (which can be seen as
having a high degree of confidence of being `correct') to support 
other trees.  Conversely, trees with a low degree of `support' in the treebank
should not support other trees.  Given these considerations, the data-driven
learner has the following matching algorithm.  Let $\Upsilon$ be a set of triples 
encoding a treebank.  For some triple $x$ in $\Upsilon$ of the form
$\langle a,b,f \rangle$, let $M(x) = a$, $D(x) = b$, and $F(x) = f$.
Define the function {\it lookup}:$A \times B \mapsto [0,1]$, where $A$ and
$B$ are categories, as follows:
\begin{equation}
\mbox{\it{lookup}}(A,B) = \left\{ \begin{array}{ll}
\sum_{\forall i \in \tau} F(i) / \sum_{ \forall j \in \Upsilon} F(j) & \tau \neq \{\} \\
\delta & \mbox{otherwise}   \end{array}
                   \right.
\end{equation}
Here, $\tau$ is the set of triples drawn from $\Upsilon$, such 
that each triple
$x$ in $\tau$ satisfies $M(x) \sqcup A$ and $D(x) \sqcup B$.  $\delta$ is a
small value that allows the ergodic assumption to be made.
Now, a local tree $t$ of the form
\begin{eqnarray*}
(A_0 (A_1 a_1 \ldots a_n) (A_2 b_1 \ldots b_n) \ldots (A_n c_1 \ldots c_n)))
\end{eqnarray*}
 produced during interleaved parsing and learning
is scored as:
\begin{eqnarray}
\mbox{{\it score}}(t) = \mbox{{\it geo-mean}}(\mbox{{\it lookup}}(A_{0},A_{1}),\
 \mbox{{\it lookup}}(A_{0},A_{2}), \ldots , \mbox{{\it lookup}}(A_{0},A_{n}))
\end{eqnarray}
Here, the categories $A_1 \ldots A_n$ immediately dominate lexical items.
The geometric mean, and not the product, is used to compose scores, thereby
avoiding  penalising local trees with more daughters over local trees with
fewer daughters \cite{Mage91}.

Interior trees of the form 
\begin{eqnarray*}
(A_0 (A_1 (B_1 \ldots )) (A_2 (B_2 \ldots)) \ldots (A_n  (B_n \ldots )))
\end{eqnarray*}
where $A_1 \ldots A_n$ immedediately dominate other nonterminals
are scored as:
\begin{eqnarray}
\mbox{{\it score}}(t) & = \mbox{{\it geo-mean}}&( \mbox{{\it lookup}}(A_{0}, A_{1}) * \mbox{{\it score}}(A_{1}),
 \mbox{{\it lookup}}(A_{0},A_{2}) * \mbox{{\it score}}(A_{2}),  \hspace{0.25in}\nonumber \\  
& & \ldots , \mbox{{\it lookup}}(A_{0},A_{n}) * \mbox{{\it score}}(A_{n}))
\end{eqnarray}
Including the scores of dominated material provides contextual support
for local trees.

If some $A_i$ is a  disjunctive category, the score then is the maximal score
of the local, non-disjunctive trees encoded disjunctively.  Taking the
maximum is equivalent to an interpretation of disjunction used in
fuzzy logic.

Finally, after scoring local trees, a local tree $t$ of the 
form 
\begin{eqnarray*}
(A_0 (A_1 (B_1 \ldots )) (A_2 (B_2 \ldots)) \ldots (A_n  (B_n \ldots )))
\end{eqnarray*}
is judged to be
sufficiently similar to previously seen local trees (encoded in the treebank)
if:
\begin{eqnarray}
\mbox{{\it geo-mean}}(\mbox{{\it score}}(A_0), \mbox{{\it score}}(A_1), \ldots , \mbox{{\it score}}(A_n)) >
\omega
\end{eqnarray}
where $\omega$ is a small value, greater than $\delta$.  This judgement 
forms the inductive component of the Grammar Garden's critic.  Only 
local trees corresponding to instantiations of the super rules are subject
to judgement, but because of the need to provide contextual support, all
local trees are scored.

In summary, a treebank is collapsed into a set of triples, which are then
used by a tree matching algorithm.  The algorithm is then used to score
local trees produced during interleaved learning and parsing and local
trees arising from usage of super rules are subject to inductive
criticism. Our data-driven learner is similar to other approaches
(for example \cite{Leec87,Haig88,Sout90,Pere92})  based
upon treebank recognition.

As an aside, it is sometimes useful to post-process a grammar.  Post-processing
is aims to reduce a grammar's overgeneration.  This is achieved by
refining the grammar.  Grammar refinement consists
of removing disjuncts from disjunctive categories, removing rules if they
have insufficient inductive support and removing rules whose structural
support has been undermined.  

Removing disjuncts is achieved by multiplying-out rules encoded within the
single disjunctive rule and scoring each such rule.  Scoring is the same
as for local, non-interior trees and the rule with the highest score is
selected, replacing the previous disjunctive rule within the grammar.  If
no single highest scoring rule exists, the disjunctive rule remains within
the grammar.  For example, the disjunctive rule $\{A, B\} \rightarrow C~D$
would be expanded into \psr{A}{C~D} and \psr{B}{C~D}.  If 
{\it score}(\psr{A}{C~D}) $>$ {\it score}(\psr{B}{C~D}) then \psr{A}{C~D}
would be used to replace $\{ A,B \} \rightarrow C~D$.

If the score of a rule fails to exceed a small value then that rule is
removed from the grammar.  

Removing rules whose structural support has gone is best shown with an 
example.  Suppose the rule \psr{A}{B~C} was learnt in the context of
the local tree (A (B (E F)) C).  Now, if the rule \psr{B}{E~F} was
no longer in the grammar\footnote{Other options might be to only remove
the rule if it is no longer {\it useful}.  A useful rule is
one that can be eventually re-written as a sequence of terminals.}, then arguably, the structural support for the
rule \psr{A}{B~C} is absent, and so the rule should be removed from the grammar.
To achieve this, the local tree dominated by a learnt rule is recorded and
if any rule within that local tree is no longer within the grammar, the rule
is also  removed from the grammar.  This process is similar to 
ensuring consistency in truth maintenance systems

This concludes the description of the data-driven learner.  It is now time
to look at examples of data-driven learning.
\section{Examples of data-driven grammar learning}

This section demonstrates the capabilities of data-driven learning by 
firstly re-presenting the learning example given in chapter three, 
 showing the effect of gradually increasing the inductive threshold
$\omega$, secondly by giving an
example of rule refinement, and finally, by showing how data-driven learning
can improve on the performance of model-driven learning.  The 
grammar and lexicon used in these examples are the same as in the previous
chapter.  

Prior to learning, the data-driven learner needs a treebank for tree scoring.
To this end, the following sentences were parsed:
\numberSentence{Sam chases the cat.}
\numberSentence{The cat chases Sam.}
\numberSentence{The cat down the road chases Sam.}
\numberSentence{*Sam down the road chases the happy cat.\footnote{This
ungrammatical sentence is introduced to show that the data-driven learner
cannot validate the grammaticality of the training set.}}
Parsing this (tiny) corpus produced $68$ triples.  
Now, suppose the {\it Sam chases the happy cat} sentence 
is processed, using just data-driven
learning, with an $\omega$ value of  $0.35$.  As before, the grammar
cannot generate this sentence, and so learning takes place:
\begin{verbatim}
165 Parse+>> Sam chases the happy cat
learning
8 rule(s) acquired.
63 parse(s)
10083090 msec CPU, 11104000 msec elapsed
\end{verbatim}				
Here, the $\omega$ value is too small, and little data-driven criticism takes
place.  Repeating this experiment, but with a higher $\omega$ value of  $0.40$
produces the following result:
\begin{verbatim}
155 Parse+>> Sam chases the happy cat
learning
4 rule(s) acquired.
7 parse(s)
201300 msec CPU, 232000 msec elapsed
\end{verbatim}
Increasing the threshold has resulted in super rule instantiations being
rejected.  Using atomic symbols as a notational convenience, the 
four rules learnt are:
\begin{center}
\begin{tabular}{l}
$\{A1,A2,N1,N2\} \rightarrow A1~N1$ \\
$\{V0,V1 \} \rightarrow V0~Det$ \\
$\{A0,A1 \} \rightarrow Det~A0$ \\
$\{N2,N3,V0,V1\} \rightarrow N2~V0$
\end{tabular}
\end{center}
As it turns out, these four rules are all equally scored and so are
indistinguishable from each other.  As such, the data-driven learner cannot
select the desired rule from the undesired set of rules.  Still, without
LP rules, types, or a Head Feature Convention, it has managed to
eliminate four other implausible rules.  

This has shown examples of 
how data-driven learning compares with model-based learning.
However, data-driven learning can compensate for incompleteness within the
model of grammaticality.  Consider rule LHSs.  As can be seen from the
previous learnt rules, they all contain disjunctions.  This is because the
rule constructor usually cannot determine what the LHS category should be.
However, data-driven learning is capable of refining rules.  Suppose the
model-based learner acquired the rule:
\begin{center}
$\{A1,N2,A2,N1\} \rightarrow A1~N1$
\end{center}
This rule, being disjunctive, needs to be refined.  The data-driven learner
can refine this rule by monitoring its usage in parse trees.  For example,
when parsing the following sentence:
\begin{verbatim}
167 Parse+>> Sam chases the happy happy cat
1 parse(s)
6100 msec CPU, 7000 msec elapsed
\end{verbatim}
the system could record the triples produced. This  exposure to data  causes an
increase in  the frequency of the mother-daughter pairs
$\langle N1, A1 \rangle$ and $\langle N1, N1 \rangle$ 
 over and above that of 
the frequency of the  mother-daughter pairs
(such as $\langle NP, N1 \rangle$) which support other categories
as candidates for the LHS.   
The system 
could then determine how the rule is used:
\begin{verbatim}
168 Parse+>> !(refine-grammar)
Refining and deleting rules ...
 Refining 4 rules encoded in *binary510746 score: 0.14390989949130545
rule is N1 -> A1 N1
\end{verbatim}
That is, the data-driven learner has refined the 
learnt rule, throwing away
the extra disjuncts, and giving the rule:
\begin{center}
$N1 \rightarrow A1~N1$
\end{center}

This example of rule refinement is something that the model-based learner, through incompleteness,
 could not achieve.

Finally, if we return to the PP-problem in the last chapter, we saw
how the model-based learner acquired more rules than necessary:
\begin{verbatim}
17 Parse+>> Sam chases the cat down the road
learning
8 rule(s) acquired.
13 parse(s)
189290 msec CPU, 226000 msec elapsed
\end{verbatim}
Now, if data-driven learning was also used, then some of these implausible
rules could be rejected.  The following table shows the number of
rules rejected, when using model-based learning and data-driven learning, 
for the PP-problem, against a varying $\omega$ value:
\begin{center}
\begin{tabular}{|l|l|} \hline
$\omega$ value & Number of rules acquired \\ \hline
0.010 & 8 \\
0.040  & 7 \\
0.050 & 6 \\
0.060 & 5 \\
0.062 & 4 \\
0.070 & 3 \\
0.080 & 2 \\
0.090 &  1  \\
0.100 & 0 \\ \hline 
\end{tabular}
\end{center}
Hence, data-driven learning can reduce the number of spurious rules learnt.
However, being inductive, it can also throw away rules that are not
spurious.  For example, when $\omega = 0.070$, the rules learnt are:
\begin{center} \label{ppProblem}
\begin{tabular}{l}
$\{ V0,V1,P3,P2,N3 \} \rightarrow V0~ \{ P3,N3,P2 \} $ \\
$\{ V1,V2,P2,P3 \} \rightarrow V1~P2 $  \\
$\{ N2,N3,P2,P3 \} \rightarrow N2~P2$ \\
\end{tabular}
\end{center}
When $\omega = 0.080$, the first of these rules is thrown away.  Whilst
this rejects a rule such as:
\begin{center}
$\{V1,V0, N3\} \rightarrow V0~N3$
\end{center}
it also rejects a rule that attaches the PP to the verb.  As can be seen, 
the relationship between $\omega$ and the rules rejected is much less clear
than for the model-based learner.  However, it is far easier in data-driven
learning to vary the aggression of the critic than  with model-based
learning.  In data-driven learning, all that needs to be changed is the
$\omega$ value: higher for fewer rules being learnt, lower for more rules
being learnt.  In model-based learning, to increase the
rejection level means extending the model of grammaticality.  This entails
knowledge engineering, which is obviously labour intensive.

In sum, we saw how data-driven learning works, and how it can interact
with model-based learning.  As should be apparent, neither learning style
in isolation is ideal, but using both together helps achieve a better
solution overall.
\section{Discussion}

This chapter explained what is meant by data-driven learning and went on
to show how a grammar could be extended using this learning style.  The
chapter also showed how data-driven learning can be used in conjunction
with model-based learning.

There are two advantages to using data-driven learning:
\begin{center}
\begin{itemize}
\item The learner is complete.
\item There is little logistical effort required, other than
the creation of the treebank.
\end{itemize}
\end{center}
As was previously stated, `completeness' here means that the learner
can always make a decision.  In the context of a learner using a treebank
and tree matching algorithm, this means always returning a non-zero 
score for any local tree being matched.  Advocates of data-driven 
grammar learning often cite this advantage as a rationale for their work.
This is because they value the ability 
of inductively constructed grammars always 
to provide an analysis for some sentence, no matter how ungrammatical 
that sentence might be. As should be clear from the stance taken in this
work, not everyone (including the author) views this as an advantage. Some
situations, such as message understanding, might always require some 
syntactic analysis.  Other applications, such as handwriting recognition\footnote{Grammars have also been used to recognise handwriting (for example \cite{Brue94}).}
of cheques,
might demand that the system at times rejects input that is 
unrecognisable.  

The second advantage means that knowledge engineering is kept to a
minimum.  Linguists are not required
to formulate grammatical models 
within the data-driven approach to grammar learning.  It could therefore be
argued that data-driven approaches are `theory-neutral' and hence 
unaffected by
reformulations (or abandonment)  of linguistic principles.  

Data-driven learners also have weaknesses:
\begin{center}
\begin{itemize}
\item They  cannot identify in the limit natural languages.
\item They usually approximate the Noisy Channel Model
and hence make mistakes over and above their theoretical limitations.
\item There never will be enough data to train upon and hence 
the grammars will be undertrained.
\item Data-driven learners cannot distinguish rare constructs from
ungrammatical sentences.
\end{itemize}
\end{center}
The first weakness follows from the work of Gold and was considered
in chapter two.

Approximate formulations  of the Noisy Channel Model include using inadequate
formalisms,
 using noisy
training data, and
 using decision
theories that do not always have the desired behaviour.

Examples of using inadequate formalisms can be seen in the
work of Garside {\it et al.} \cite{Gars87}, or in the use 
Hidden Markov Models in speech recognisers (for example \cite{Jeli90}).  These
 are  all  finite state technologies, which cannot
 capture all 
 natural language constructs.  Apart from theoretical objections, 
there is empirical support of the inadequacy of finite state technologies.
Lari
and Young, in 
their experiment  of learning the palindrome language (which is context
free), found that using a context
free grammar  produced   better results 
  than when using a Hidden Markov Model \cite{Lari90}.

Noise undermines the ability of the data-driven learner to make
correct decisions.  In the context of  a data-driven learner using a 
manually constructed  treebank, variability of analyses means that
regularities being sought cannot always be found.  This problem has been
noted in the literature  (for example, Brill {\it el al.} comment that
in the Penn Treebank,  on average $3.2\%$ of the words are 
mis-tagged \cite{Bril92}; Black {\it et al.} comment that the treebank
that they use contains $2.5\%$ noise \cite[p.194]{Blac93}).

A {\it decision  theory} is one that helps the learner decide whether
to accept an hypothesis, or reject that hypothesis.  
Inductive grammar learners typically use a statistical decision theory.
However, as 
 Carroll shows, these decision theories do not always give the desired
results \cite[p.133]{Carr93}.  For example, associating probabilities with context free rules
means that the decision theory cannot arbitrate between different
derivation sequences.  In the context of grammar learning, a differing
derivation sequence might correspond to a competing set of rules for
some sentence. Hence, such a decision theory would not always be able
to decide which set of rules to prefer.  Regarding the data-driven learner, 
 using the context of a local tree 
can be seen as  an attempt at overcoming this  problem.  Note that  taking
the geometric mean means that the scores given to local trees do not have
probabilistic interpretations.

Returning to the list of weaknesses of data-driven learners, 
it is clear that, assuming natural languages are infinite, there will
never be enough data available to fully train an inductively constructed
language model.  However, researchers can approximate an infinite size
training set with a training set that is suitably large.  Unfortunately,
`suitably large' can be very large indeed.  For example, Church and Mercer
show that to obtain reliable information about the adjective `strong' 
requires at least 46 million words \cite{Chur93}.  Of course, this is
an upper bound, and less data could be used, assuming  that 
smoothing techniques are employed to deal with the resulting sparse
statistics \cite{Gale90}.  Smoothing approaches, however, 
can only estimate the underestimated
parameters of an inductively constructed language model.  Therefore, in
practise, the size of the 
training sets required will pose such a formidable computational task that the
resulting grammar, even with smoothing, will be undertrained.  

As data-driven
learners base their concept of grammaticality upon frequency, the learner 
 cannot  distinguish between rare constructs and 
ungrammatical constructs. Rare
constructs will have a low probability (by definition).  Ungrammatical
constructs will also (hopefully) have a low probability.   Therefore, if a low probability is taken
to mean an ungrammatical construct, then correction of such a construct
would mean that grammatically well-formed sentences would be incorrectly
corrected.  This is suboptimal behaviour and undermines the performance
of a sentence correcting device.  

All of these weaknesses with data-driven learning can be tackled by
also using model-driven learning.  Chapter seven considers the link
between these two learning styles in detail.

The previous two chapters demonstrated various aspects of the learners.  In
the next chapter, the system is evaluated with naturally occurring language.

\chapter{Evaluation}

\section{Introduction}

The previous two chapters demonstrated capabilities of the learning
system.  This chapter  evaluates the system in terms
of the success criteria introduced in chapter two and principally
aims to demonstrate 
that combining data-driven and model-based learning produces 
qualitatively better grammars than  are produced when using either learning
style in isolation.\footnote{Some of these 
experiments have also been published in
two papers \cite{Osbo94a,Osbo94b}.}

Broadly speaking, there are two 
approaches to evaluating machine learning systems.  They can be either {\it formally} or {\it empirically} evaluated.
A formal evaluation will prove the system meeting the success criteria, whilst 
an empirical evaluation will only show the system meeting the success 
criteria.  In chapter
two, one such formal analysis (Gold's Identification in the 
Limit \cite{Gold67}) was introduced.  Another, more recent formal approach is
{\it Probably-Approximately-Correct Learning} (PAC learning)
\cite{Vali84}.  The has-it
or has-it-not identified the language nature
 of Gold's approach is replaced with a how-well-has-it identified the language
framework.
PAC learnability is therefore better suited than Gold's approach for
many machine learning systems.  Yet another formal approach is computational
complexity theory, which gives characterisations of time and space
requirements of solutions to various classes of 
problems \cite{Berw87}.  Unfortunately,
none of these formal approaches are suitable for evaluating the learning
system.  Gold's approach is too coarse and says nothing about 
approximations \cite[p.3]{Nata91}.
PAC learnability talks about approximations, but the theory is not sufficiently
advanced to deal with learning natural language 
grammars \cite{Bunt89,Kiet94}.  Complexity theory
considers worst case behaviour and does not address domain specific issues
such as parse plausibility  or undergeneration.  Indeed, Kibler and Langley
comment that ``\ldots many learning algorithms remain too complex for 
formal analysis.  In such cases, empirical studies of the behaviour of these
algorithms must retain a central role.'' \cite{Kibl88}.  Hence, the approach taken
when evaluating the system in this thesis is empirical. 

Related work on empirically analysing grammars  tend 
to concentrate upon measuring `correctness' (plausibility) of parses
and/or measuring overgeneration.  Plausibility is usually measured either
qualitatively (by manually inspecting parses, for example 
\cite{Bris91,Bris92}) or
quantitatively (by tree-matching algorithms, for 
example \cite{Harr91,Blac93}).  
Qualitative measurement  is labour intensive and hence 
quantitative methods are (from a logistical perspective) 
preferable.  Overgeneration is usually quantitatively measured (for 
example \cite{Lari90,Carr92,Bris92}) in terms of
how random  the language generated by the grammar is.  
The  more random the language,  the greater the grammar's 
overgeneration.  Few approaches
consider undergeneration, given the ergodic assumption made by most inductively
constructed
grammars,  which eliminates undergeneration.
  An ideal 
approach would consider all three aspects of a grammar (undergeneration,
overgeneration, and plausibility), not just 
any one in
isolation.  This would  counter arguments such as  success at dealing with
 undergeneration being due to excessive overgeneration. Unfortunately, 
few systems are evaluated in terms of all three aspects of grammaticality.
This thesis uses all three aspects of grammaticality.

The experiments in this chapter follow the same  outline:
\begin{center}
\begin{itemize}
\item Run the configured 
learner over the training material and produce a grammar.
\item Evaluate that grammar using a set of metrics.
\end{itemize}
\end{center}
In general, there are many aspects of the learner that could be tested
and so for logistical reasons, only those relating to the success
criteria outlined in chapter two will be considered.  So, the training
material (for example) will be kept constant,  but the learner's 
configuration  will vary.

The rest of this chapter is as follows.  Section \ref{metrics}
 outlines the metrics used;
section \ref{experiments} describes a set of experiments evaluating the learning system
and finally, section \ref{evaldisc}
 discusses what has been revealed about the learning 
system.

\section{Metrics \label{metrics}}

This section presents three quantitative metrics measuring
grammar quality: a metric testing a grammar's undergeneration,
a metric testing a grammar's overgeneration, and finally, a 
metric testing for parse plausibility.

\subsection{Measuring undergeneration}

Since undergeneration is defined as a grammar's inability to generate
a grammatical sentence, an obvious way to measure this would be to construct
a set of grammatical sentences and then determine how many of these
sentences are generated by the grammar.  The more of these 
sentences generated, the
lower the undergeneration of a grammar.  Grammatical sentences can be
found in a corpus of naturally occurring language.

\subsection{Measuring overgeneration}

Testing for overgeneration is similar to testing for undergeneration.
A set of ungrammatical strings should be selected and 
we should determine how many of these
are generated by a grammar.  The fewer strings generated 
by the grammar, the lower
the grammar's overgeneration.  It is harder to find a set of ungrammatical
strings than it is to find a set of grammatical sentences.  One
 possible way to
locate strings  would be by
 concatenating different numbers of randomly chosen  terminals together to
generate a set of strings
of any length.  Such a set would be {\it largely} ungrammatical on the grounds
that natural languages are predictable.  Clearly, random generation of
strings results in a language that is not predictable.    Other 
researchers (for example \cite{Lari90,Carr92,Bris92,Jone93}) 
use this idea indirectly and measure overgeneration as
the {\it entropy}
of the language generated by the grammar.  
Entropy is a measure of how much information is produced by (say) a
word in a text.  If the words are replaced by bits in an optimal manner, then
the entropy $H$ is the average number of bits required
per word: the 
higher the entropy, the greater the ungrammaticality of the set of 
strings.  For example, Shannon estimated the entropy of encoding letters
in English roughly to be  $2.3$ bits per letter \cite{Shan48}.    A problem
with using entropy is that the figure  gives no insight into how a
 grammar overgenerates. 
By measuring overgeneration directly in terms of generating ungrammatical
strings, it is easier both to locate sources of overgeneration and also
to see what the figure for overgeneration means.  This work therefore
uses the random string generation approach to measuring  overgeneration.

\subsection{Plausibility}

It is difficult to determine if a parse for some sentence is plausible.
For example, Harrison {\it et al} reported an experiment involving the
comparison of manually produced parses for $50$ sentences taken from
the Brown Corpus \cite{Harr91}.  The results revealed that there
was little common structure between the
parses, reflecting a diversity of approaches to punctuation, empty categories,
and so on.  Hence, what is plausible for one person is not plausible 
for another.
One  popular avoidance of  this problem with manual 
plausibility determination
is  to define   parse plausibility  as
conformity to a benchmark parse.  The greater the deviation of the test parse
from the benchmark parse of the same sentence, the less plausible the
test parse.

The approach in this work is based loosely upon the work of Harrison {\it et 
al\footnote{The chief difference between our matching algorithm and their
approach is that we match against the entire tree, whilst Harrison
{\it et al} only match against the bracketing of the tree.} }
and the matching algorithm consists of the following steps:
\begin{center}
\begin{enumerate}
\item Normalise the test parse to use the same labelling scheme as the
benchmark parse.
\item Flatten both test parse and benchmark parse into the lists $\tau$ and
$\beta$ by a preorder walk.  
\item Starting from the head of $\tau$, find the longest list that is 
in $\tau$
that is also in $\beta$.  Remove this list from $\tau$.  Repeat this, 
 removing the longest
list again, until either $\tau$ is empty, or no such list is common to
 both $\tau$ and
$\beta$. \label{step3}
\item The match between the test parse and the benchmark parse is the
arithmetic mean of the list lengths of the lists found in step
\ref{step3}  divided by the list
length of $\beta$.  A score of $1$ is a perfect match and a score of $0$ is
a perfect mismatch.
\end{enumerate}
\end{center}
The test parse has to be normalised so that like is compared with
like.  This is particularly true when the grammar formalisms  and the
assumptions behind the analyses differ.  For example, grammars learnt in this
thesis are unification-based and assign steep parses.  The benchmark parses
use atomic labels and are shallow \cite{Gars87,Leec91}.  Normalisation is 
performed by
mapping feature structures to atomic categories.  Categories with bar levels
greater than one are mapped to  phrasal atomic categories, thereby
flattening steep parses.  By matching lists, and not trees, some
structural information is thrown away.  Again, this helps in the normalisation
process, at the cost of making the match only approximate.

For example, if $\tau$ was the list (a b c d) and  $\beta$ the list 
(c a b c), then the first longest sublist common to both would   
be (a b c).  Removing  this list from $\tau$
results in $\tau$ becoming the list (d).  As there are now no lists common
to both $\tau$ and $\beta$, matching halts, with the matching
lists  being $\{$(a b c)$\}$.
The closeness score would then be $3/4$.

Although the matching algorithm is simple, it does give adequate results.  For
example, the University of Pennsylvania Treebank  benchmark parse:
\begin{flushleft}
\verb|(S (S (NP MISS X) (VP WAS BEST))|
\\ \verb|   (SBAR WHEN (S SHE (VP NEED (V BE (ADJP TOO PROBING))))))|
\end{flushleft} 
when matched against the test parses:
\begin{flushleft}
\verb|(S (NP-S MISS X) (VP WAS BEST)|
\\ \verb|   (S WHEN SHE (VP NEED (V BE (ADJP TOO PROBING)))))|
\verb|(S (S (NP MISS X) (VP WAS BEST))|
\\ \verb|   (S (WHEN (S (NP SHE) (VP NEED BE (ADJP TOO PROBING))))))|
\verb|(ADJP (S (ADJP (NP-S MISS (VP X WAS)) BEST) WHEN)|
\\ \verb|      (NP-S SHE (NP-S (NP-S NEED BE TOO) PROBING)))|
\end{flushleft}
gives a highest match with the first test parse.  The lowest match is
with the last parse (which was randomly generated).

In practise, many parses will be produced for a sentence and so the first
$k$ ($k=10$) parses are sampled and matched with the  benchmark parse.  We record the
score for the 
best matching tree.  Note that if all of the parses that a grammar could
generate for some sentence were considered, then the 
plausibility matching would benefit from overgeneration.  That is, if
some grammar generates all possible parses, then there would be one that
matches exactly with the benchmark parse.  However, sampling only $k$ parses
ensures that the matching only looks at a subset of these parses.  This means
that on average, assuming that most sentences have more than $k$ 
possible parses, then the effects of overgeneration can be minimised.  

\subsection{Comments}

Given these metrics, grammar $A$ is of higher quality than grammar $B$ if
$A$ undergenerates and overgenerates less than grammar $B$ and assigns more
plausible parses than $B$.

None of these metrics are exact measurements, due to the  fact that they all
sample finite subsets of the infinite languages that each grammar can
generate.  There is a risk that some unrepresentative subset
of this language is sampled, thus undermining the results.  For example, the
supposed ungrammatical strings  could in reality be sentences.  However,
by selecting naturally occurring sentences  when  measuring undergeneration,
by randomly generating strings when measuring overgeneration, 
 and by using benchmark parses that are manually
produced so as to be plausible when measuring plausibility, the 
representativeness of each set of
 data is enhanced.  This is similar to stratified sampling techniques 
used by statisticians trying to overcome unrepresentativeness when dealing
with large sample spaces.

\section{Experiments \label{experiments}}

This section describes a series of experiments that determines how well 
various configurations of the learner meet the success criteria.   It also
describes an experiment showing the convergence rate of the system.  
Convergence is defined as the system identifying the language; it is 
is worth considering, given the fact that a system that converges
rapidly is preferable to one that converges less rapidly, all things
being equal.

A manually constructed grammar, $G1$, consisting of 97 unification-based rules
was used throughout all experiments and extended by
various styles of learning.\footnote{The grammar
$G1$ was kindly supplied by Ted Briscoe (University of Cambridge).  A listing
of $G1$ appears in appendix \ref{previous}. The lexicon of $G1$ is given in
appendix \ref{lex}.}  A typical
rule from $G1$ is:
\begin{center}
\psr{\fs{N - \\  V + \\ BAR 2 \\ MINOR NONE \\ PLU \fbox{6} \\ VFORM \fbox{17}}}
    {\fs{N + \\ V - \\ BAR 2 \\ MINOR NONE \\ PLU \fbox{6} \\ CONJ -}~
     \fs{N - \\ V + \\ BAR 1 \\  MINOR NONE \\ PLU \fbox{6} \\ VFORM \fbox{17} \\ CONJ -}}
\end{center}
(which can be paraphrased, using atomic symbols, by the rule \psr{S}{NP~VP}).

The learning  of grammars requires data and  
 in keeping with other researchers (for example \cite{Bris92,Carr94,Atwe94}),
the Spoken English Corpus (SEC) was chosen as a source of training and testing 
material \cite{Leec91}.\footnote{Access to the SEC was kindly given by
Eric Atwell (Leeds University).}  The SEC consists of
 {\it c.\ }$50,000$ words of prepared monologues
broadcast over the radio.   An advantage
of using the SEC is that it is lexically tagged (using the CLAWS2 
tagset \cite{Blac93})
and manually parsed (using the UCREL parsing scheme \cite{Gars87}), saving on the need to
construct a suitable lexicon or to construct a set of benchmark parses for
plausibility evaluation.  
A typical entry in the SEC is:
\begin{verbatim}
[N It_PPH1 N]
[V 's_VBZ [N a_AT1 useful_JJ reminder_NN1 
             [Fn that_CST 
                 [N some_DD scientists_NN2 N]
                 [V find_VV0 [N Don_NP1 Cupitt_NP1 N]
                 unscientific_JJ V]Fn]N]V] ._.
\end{verbatim}
Here, the sentence:
\numberSentence{It's a useful reminder that some 
scientists find Don Cupitt unscientific} 
 is shown tagged (for example, the word {\tt reminder} has
a tag NN1) and parsed (indicated by the 
square brackets and their adjacent phrasal 
categories).  

As a resource for evaluation, the SEC is quite good.  
However, in common with all automatically tagged corpora, some of the
words in the SEC are  mistagged. The SEC sentences are mostly grammatical,
given that they have been edited in preparation for broadcast.  However,
the manual
parses in the SEC are uneven in quality.  Because of this quality 
problem, those
 manual parses
used in plausibility evaluation had, in some cases, to be edited.  Mostly this
editing 
consisted of adding a sentence root node  to the SEC parse
trees.  Because of the mistagging and the uneven quality of the parses, the
experiments will only reveal approximate results of the system.  In 
particular, the plausibility results will only allow  relative system
performance, and not allow reliable 
comparison with other systems that use the SEC
as data.  

All punctuation was removed from the SEC sentences used in the 
experiments.  This was
because it was difficult to determine when punctuation was used
syntactically (for example in ``apples, pears and bread baskets'') or
when it was used textually (for example in ``John laughed: Bill tickled
him'').\footnote{The examples on which these are based were supplied 
by Ted Briscoe
(personal communication).} A grammar, if it is to be a theory of syntax, 
needs to distinguish between 
these two uses of punctuation.  There has been work
on computationally dealing with punctuation \cite{Numb90,Carr94}
 and it would be interesting to 
see, as future work, 
 if punctuation has any impact upon the quality of learnt grammars.

Prior to learning, the $60$ shortest\footnote{Short sentences were
used simply to reduce the computation involved with learning.  Other 
learners, such as DACS \cite{Naum93} also use short sentences.}
 sentences were selected
from the SEC without regard to how syntactically well-formed they were
and set aside as the set of sentences {\it Train}. These were used
as training material.  A further
(different)  $60$ sentences
were  set aside, also selected without regard to syntactic 
well-formedness, as the set {\it Test}. These were used to measure
undergeneration.  A small number of sentences (less than
20), also different from {\it Train} or {\it Test},  were also selected 
(called {\it Pretrain}) to pretrain the data-driven learner.  Pretraining is
necessary in order to give an initial estimate of the language 
model  \cite{Shar88}.  $15$ sentences from {\it Test} were selected 
(called {\it Yardstick}) and each sentence was paired with its associated 
manual parse taken from the SEC treebank.  A further $15$ different 
sentences and their
manual parses  (called {\it Plausible}) were also selected from the treebank.
The sentences in {\it Yardstick} were selected such that they could all
be generated by $G1$, whilst the sentences in {\it Plausible} each required
at least one learnt rule in order to be generated.  Both of these sets of 
sentence-parse pairs were used  to determine plausibility.  {\it Yardstick}
served as the base for comparison with {\it Plausible}.  Finally, $100$
strings of length $6$ (called {\it Random}) 
were randomly generated using grammar $G1$'s lexicon.  These strings were
used to measure overgeneration. \footnote{Listing of 
 {\it Test}, {\it  Train}, {\it Bad}, {\it Yarstick} and 
{\it Plausible} appear in appendix \ref{qqq}.}
Before proceeding, it is necessary to justify the size of the data sets 
used in the experiments.  By comparison with other researchers, these
sets are small, and it could  therefore be argued  that system evaluation
using this amount of data is unconvincing.  In other data-driven
learning approaches,  researchers need to use larger amounts of data. As a
consequence of the learning style they use, if they are    to 
achieve a reasonable performance from their technologies, they need to use
large data sets.  However, evaluation in this thesis is not concerned with
measuring how well the data-driven or the model-based   learners perform in the
limit.  Evaluation is concerned instead with determining if using data-driven learning 
and model-based learning together 
is better than using either learning style in isolation.
Hence, all that matters is that a difference is observed, and there is no
need to compare  fully trained learners at all.  In fact, it could equally
be argued that hoping for full convergence 
 is an incoherent idea,  given 
 that natural languages are infinite in size, and training sets, being finite, will never be 
enough.  Clearly however, in evaluation, the 
data sets need to be sufficiently large for any
effects to be noticed.  As it turns out, good estimates, based upon using 
only small 
data sets, can be obtained.  {\it Chernoff bounds} \cite{Cher52}
 tell us the probable rate of
convergence of estimating a variable   to the true value of that variable:
the probability that the estimate is inaccurate goes to 0 exponentially
fast as the size of the data set increases.  That is, more is better, but 
much more is not much better.  Using ever larger data sets brings diminishing
returns.  For example, Brill \cite[p.72]{Brill93} explored the effect of
training set size  on the accuracy of his lexical tagger and found 
the following behaviour:
\begin{center}
\begin{tabular}{|l|l|} \hline
Training size (sentences) & Accuracy ($\%$) \\ \hline 
1000 & 90.5 \\
2000 & 91.7 \\
4000 & 92.2 \\ \hline
\end{tabular}
\end{center}
That is, quadrupling the number of sentences only increases accuracy 
by $1.7\%$.  In sum, small (but not tiny) 
data sets can be used, assuming that any
conclusions made based upon these sets do not relate to absolute system
performance, but only to relative performance.  It is believed that the
size of the data sets used in these experiments, though small, are
sufficient for showing some of the differences between data-driven and
model-based learning. The confidence level results in 
\S \ref{plaus} support this belief.

Learning, as outlined in chapter three, is computationally very expensive
and so resource bounds were placed upon the learners.  These bounds were to
stop learning when either $n$ parses or $m$ edges had been created by the
chart parser for some sentence ($n = 1, m = 3000$).  Increasing $n$ leads to
more  rules being learnt and hence increases the 
likelihood of finding highly plausible parses (i.\ e.\ those that
match the benchmark parses exactly).  The motivation for the edge limit 
follows from
others  who suggest that ungrammaticality
might be related to an excessive number of edges being generated for some 
string \cite{Mage91,Mage92,Chit92}.  In effect, the chart parser spends a 
lot of time fruitlessly searching for
parses that may not exist.  Resource bounds make learning incomplete, but
if they are kept constant across all the learning configurations, any 
incompleteness effects will be factored out.  However, incompleteness does
mean that all of the evaluation results will be systematically 
underestimated.

The learner was configured in three distinct ways:
\begin{center}
\begin{tabular}{|l|l|l|} \hline
Configuration & Parameters & Grammar produced \\ \hline
A & Data-driven learning only & $G2$ \\
B & Model-based learning only & $G3$ \\
C & Data-driven learning and model-based learning & $G4$ \\ \hline
\end{tabular}
\end{center}
The model-based learner's model consisted of $4$ LP rules, $32$ semantic
types and a Head Feature Convention.  In configurations 
A and C, the system uses {\it Pretrain} to get initial 
frequencies of mother-daughter pairs. All configurations of the learner
used X-bar syntax, as this is such a necessary aspect of rule construction.

Each configuration was run over {\it Train}, causing learning to take
place. 
Using {\it Test}, 
the learnt grammars were evaluated with 
respect 
to the three metrics.  The following table shows the size (in terms of the
number of
disjunctive and, after multiplying-out, non-disjunctive  rules) of the various 
grammars learnt:
\begin{center}
\begin{tabular}{|l|l|l|} \hline
Grammar & Size (disjunctive rules) & Size (non-disjunctive rules) \\ \hline
$G2$ & 129 & 392\\
$G3$ & 128 & 345 \\
$G4$ & 129 & 385\\ \hline
\end{tabular}
\end{center}
Appendix \ref{xxx} presents, in paraphrased form, the grammars $G2$, $G3$ and
$G4$.

Each of the grammars was then evaluated as follows.
\subsection{Undergeneration}
The following table shows the percentage of sentences in {\it Test} parsed
by all of the grammars:
\begin{center}
\begin{tabular}{|l|l|} \hline
Grammar & Percentage generated \\ \hline
$G1$ & 26.7 \\
$G2$ & 75.0\\
$G3$ & 65.0\\
$G4$ & 75.0\\ \hline
\end{tabular}
\end{center}
As can be seen, learning reduces $G1$'s undergeneration.  Data-driven learning
(producing $G2$)   and combined learning (producing $G4$) 
jointly  reduce undergeneration most.  
\subsection{Overgeneration}
The following table shows the percentage of sentences in {\it Random} parsed
by all of the grammars:
\begin{center}
\begin{tabular}{|l|l|} \hline
Grammar & Percentage generated \\ \hline
$G1$ & 7.0 \\
$G2$ & 38.0  \\
$G3$ & 30.0 \\
$G4$ & 38.0 \\ \hline
\end{tabular}
\end{center}
This time, the reverse has been found: learning increases overgeneration.
However, the increase in overgeneration is less than the reduction in
undergeneration.
\subsection{Plausibility results \label{plaus}}
The following table shows plausibility results after matching grammar
$G1$ using {\it Yardstick}:\footnote{All figures are now 
quoted to three decimal
places.}
\begin{center}
\begin{tabular}{|l|l|} \hline 
Sentence & Score \\ \hline
Y1 & 0.083\\
Y2 & 0.133 \\
Y3 &  0.092\\
Y4 &  0.075\\
Y5 &  0.191\\
Y6 &  0.079\\ 
Y7 &  0.108\\
Y8 &  0.105\\
Y9 &  0.088\\
Y10 & 0.079 \\
Y11 &  0.102\\
Y12 &  0.174\\
Y13 &  0.094 \\
Y14 &  0.090 \\
Y15 &  0.053 \\ \hline
\end{tabular}
\end{center}
The arithmetic mean and standard deviation 
of the plausibility scores are as follows:
\begin{center}
\begin{tabular}{|l|l|l|}  \hline
Mean & 0.103 \\ \hline
Standard deviation & 0.037 \\ \hline
\end{tabular}
\end{center}

The following table shows plausibility scores for the learnt grammars
using {\it Plausible}:
\begin{center}
\begin{tabular}{|l|l|l|l|} \hline
Sentence         & $G2$ & $G3$ & $G4$ \\ \hline
P1 & 0.100 & 0.075 & 0.100  \\
P2 & 0.111 & 0.156 & 0.156 \\
P3 & 0.111 & 0.156 & 0.156 \\
P4 & 0.110 & 0.100 & 0.110 \\
P5 & 0.083 & 0.083 & 0.083 \\
P6 & 0.111 & 0.102 & 0.111 \\
P7 & 0.054 & 0.048 & 0.054 \\
P8 & 0.093 & 0.093 & 0.093  \\
P9 & 0.071 & 0.071 & 0.071 \\
P10 & 0.100 & 0.111 & 0.100 \\
P11 & 0.104 & 0.104 & 0.104 \\
P12 & 0.070 & 0.070 & 0.070 \\
P13 & 0.095 & 0.048 & 0.095 \\
P14 & 0.069 & 0.069 & 0.069 \\
P15 & 0.103 & 0.082 & 0.103 \\ \hline 
\end{tabular}
\end{center}
The following table gives the arithmetic means and standard deviations
of the  plausibility scores:
\begin{center}
\begin{tabular}{|l|l|l|l|} \hline
 & $G2$ & $G3$ & $G4$ \\ \hline
Mean & 0.092 & 0.091 & 0.098 \\ \hline
Standard deviation & 0.018 & 0.032 & 0.029 \\ \hline
\end{tabular}
\end{center}
If we assume that the sentences used in the experiments are normally
distributed, then we can determine how confident we are in the plausibility
results.   The null hypothesis is that there is no statistically
significant difference between some pair of  grammars.  Given this hypothesis,
the t-test results are as follows:
\begin{center}
\begin{tabular}{|l|l|l|l|l|l|}  \hline
 & $G_2$ and $G_4$ & $G_3$ and $G_4$ & $G_2$ and $G_3$ \\ \hline
t-test & 1.418 & 1.870 & -1.300 \\ \hline
\end{tabular}
\end{center}
The t-tests measure if the null hypothesis holds when
comparing pairs of  grammars. 
From these figures, we can be more than $90\%$ confident that
there is a significant difference between grammars $G_2$ and $G_4$
and  more than
$95\%$ confident that there is a difference between $G_3$ and $G_4$.  There
is no evidence to suggest that there is a significant difference between
$G_2$ and $G_3$.  
Hence, we can conclude that
 $G4$ is more plausible than either $G2$ or $G3$.  $G4$ is 
less plausible than
$G1$.  Interestingly, the combined learning process has acquired a grammar that
is the best of either learning style in isolation.  From the standard 
deviations (which measure
how much a set of samples varies), the data-driven learner is the most
consistently plausible grammar, whilst the manually
 constructed grammar is the least consistently
plausible grammar.  

\subsection{Convergence results}
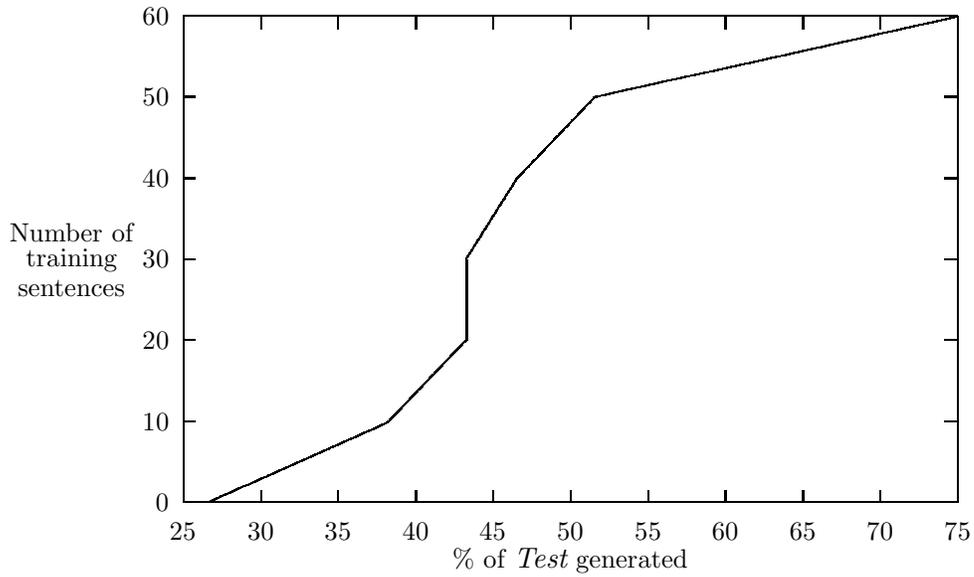
\begin{figure}
\setlength{\unitlength}{0.240900pt}
\ifx\plotpoint\undefined\newsavebox{\plotpoint}\fi
\sbox{\plotpoint}{\rule[-0.200pt]{0.400pt}{0.400pt}}%
\begin{picture}(1500,900)(0,0)
\font\gnuplot=cmr10 at 10pt
\gnuplot
\sbox{\plotpoint}{\rule[-0.200pt]{0.400pt}{0.400pt}}%
\put(220.0,113.0){\rule[-0.200pt]{292.934pt}{0.400pt}}
\put(220.0,113.0){\rule[-0.200pt]{4.818pt}{0.400pt}}
\put(198,113){\makebox(0,0)[r]{0}}
\put(1416.0,113.0){\rule[-0.200pt]{4.818pt}{0.400pt}}
\put(220.0,240.0){\rule[-0.200pt]{4.818pt}{0.400pt}}
\put(198,240){\makebox(0,0)[r]{10}}
\put(1416.0,240.0){\rule[-0.200pt]{4.818pt}{0.400pt}}
\put(220.0,368.0){\rule[-0.200pt]{4.818pt}{0.400pt}}
\put(198,368){\makebox(0,0)[r]{20}}
\put(1416.0,368.0){\rule[-0.200pt]{4.818pt}{0.400pt}}
\put(220.0,495.0){\rule[-0.200pt]{4.818pt}{0.400pt}}
\put(198,495){\makebox(0,0)[r]{30}}
\put(1416.0,495.0){\rule[-0.200pt]{4.818pt}{0.400pt}}
\put(220.0,622.0){\rule[-0.200pt]{4.818pt}{0.400pt}}
\put(198,622){\makebox(0,0)[r]{40}}
\put(1416.0,622.0){\rule[-0.200pt]{4.818pt}{0.400pt}}
\put(220.0,750.0){\rule[-0.200pt]{4.818pt}{0.400pt}}
\put(198,750){\makebox(0,0)[r]{50}}
\put(1416.0,750.0){\rule[-0.200pt]{4.818pt}{0.400pt}}
\put(220.0,877.0){\rule[-0.200pt]{4.818pt}{0.400pt}}
\put(198,877){\makebox(0,0)[r]{60}}
\put(1416.0,877.0){\rule[-0.200pt]{4.818pt}{0.400pt}}
\put(220.0,113.0){\rule[-0.200pt]{0.400pt}{4.818pt}}
\put(220,68){\makebox(0,0){25}}
\put(220.0,857.0){\rule[-0.200pt]{0.400pt}{4.818pt}}
\put(342.0,113.0){\rule[-0.200pt]{0.400pt}{4.818pt}}
\put(342,68){\makebox(0,0){30}}
\put(342.0,857.0){\rule[-0.200pt]{0.400pt}{4.818pt}}
\put(463.0,113.0){\rule[-0.200pt]{0.400pt}{4.818pt}}
\put(463,68){\makebox(0,0){35}}
\put(463.0,857.0){\rule[-0.200pt]{0.400pt}{4.818pt}}
\put(585.0,113.0){\rule[-0.200pt]{0.400pt}{4.818pt}}
\put(585,68){\makebox(0,0){40}}
\put(585.0,857.0){\rule[-0.200pt]{0.400pt}{4.818pt}}
\put(706.0,113.0){\rule[-0.200pt]{0.400pt}{4.818pt}}
\put(706,68){\makebox(0,0){45}}
\put(706.0,857.0){\rule[-0.200pt]{0.400pt}{4.818pt}}
\put(828.0,113.0){\rule[-0.200pt]{0.400pt}{4.818pt}}
\put(828,68){\makebox(0,0){50}}
\put(828.0,857.0){\rule[-0.200pt]{0.400pt}{4.818pt}}
\put(950.0,113.0){\rule[-0.200pt]{0.400pt}{4.818pt}}
\put(950,68){\makebox(0,0){55}}
\put(950.0,857.0){\rule[-0.200pt]{0.400pt}{4.818pt}}
\put(1071.0,113.0){\rule[-0.200pt]{0.400pt}{4.818pt}}
\put(1071,68){\makebox(0,0){60}}
\put(1071.0,857.0){\rule[-0.200pt]{0.400pt}{4.818pt}}
\put(1193.0,113.0){\rule[-0.200pt]{0.400pt}{4.818pt}}
\put(1193,68){\makebox(0,0){65}}
\put(1193.0,857.0){\rule[-0.200pt]{0.400pt}{4.818pt}}
\put(1314.0,113.0){\rule[-0.200pt]{0.400pt}{4.818pt}}
\put(1314,68){\makebox(0,0){70}}
\put(1314.0,857.0){\rule[-0.200pt]{0.400pt}{4.818pt}}
\put(1436.0,113.0){\rule[-0.200pt]{0.400pt}{4.818pt}}
\put(1436,68){\makebox(0,0){75}}
\put(1436.0,857.0){\rule[-0.200pt]{0.400pt}{4.818pt}}
\put(220.0,113.0){\rule[-0.200pt]{292.934pt}{0.400pt}}
\put(1436.0,113.0){\rule[-0.200pt]{0.400pt}{184.048pt}}
\put(220.0,877.0){\rule[-0.200pt]{292.934pt}{0.400pt}}
\put(45,495){\makebox(0,0){\shortstack{Number of \\ training \\ sentences}}}
\put(828,23){\makebox(0,0){\% of {\it Test} generated}}
\put(220.0,113.0){\rule[-0.200pt]{0.400pt}{184.048pt}}
\put(1306,812){\makebox(0,0)[r]{}}
\put(259,113){\usebox{\plotpoint}}
\multiput(259.00,113.58)(1.119,0.499){251}{\rule{0.994pt}{0.120pt}}
\multiput(259.00,112.17)(281.936,127.000){2}{\rule{0.497pt}{0.400pt}}
\multiput(543.58,240.00)(0.499,0.524){241}{\rule{0.120pt}{0.520pt}}
\multiput(542.17,240.00)(122.000,126.921){2}{\rule{0.400pt}{0.260pt}}
\multiput(665.58,495.00)(0.499,0.795){157}{\rule{0.120pt}{0.735pt}}
\multiput(664.17,495.00)(80.000,125.474){2}{\rule{0.400pt}{0.368pt}}
\multiput(745.58,622.00)(0.499,0.524){241}{\rule{0.120pt}{0.520pt}}
\multiput(744.17,622.00)(122.000,126.921){2}{\rule{0.400pt}{0.260pt}}
\multiput(867.00,750.58)(2.245,0.499){251}{\rule{1.892pt}{0.120pt}}
\multiput(867.00,749.17)(565.073,127.000){2}{\rule{0.946pt}{0.400pt}}
\put(665.0,368.0){\rule[-0.200pt]{0.400pt}{30.594pt}}
\end{picture}
\caption{\label{curve} The system's learning curve }
\end{figure}

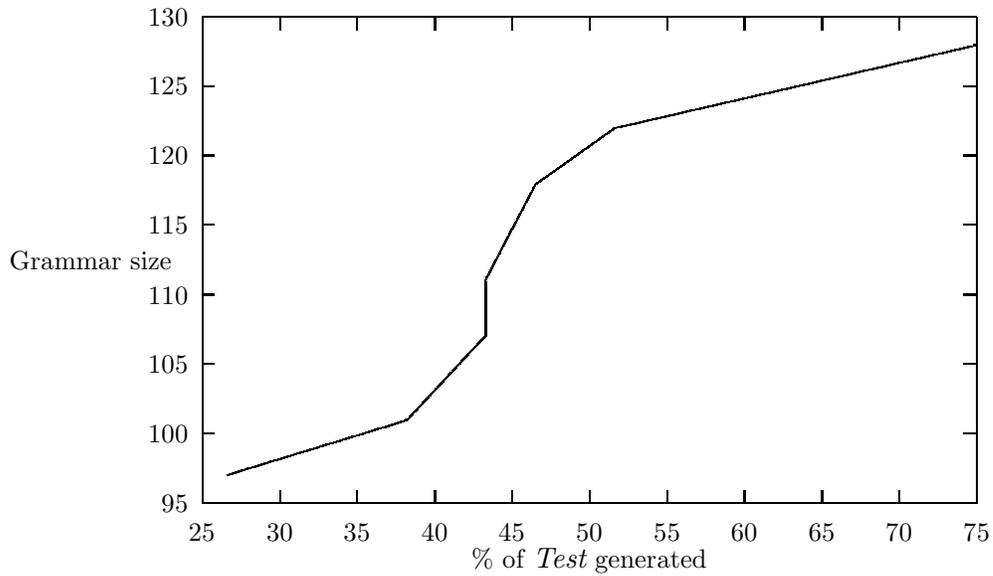
\begin{figure}
\setlength{\unitlength}{0.240900pt}
\ifx\plotpoint\undefined\newsavebox{\plotpoint}\fi
\sbox{\plotpoint}{\rule[-0.200pt]{0.400pt}{0.400pt}}%
\begin{picture}(1500,900)(0,0)
\font\gnuplot=cmr10 at 10pt
\gnuplot
\sbox{\plotpoint}{\rule[-0.200pt]{0.400pt}{0.400pt}}%
\put(220.0,113.0){\rule[-0.200pt]{4.818pt}{0.400pt}}
\put(198,113){\makebox(0,0)[r]{95}}
\put(1416.0,113.0){\rule[-0.200pt]{4.818pt}{0.400pt}}
\put(220.0,222.0){\rule[-0.200pt]{4.818pt}{0.400pt}}
\put(198,222){\makebox(0,0)[r]{100}}
\put(1416.0,222.0){\rule[-0.200pt]{4.818pt}{0.400pt}}
\put(220.0,331.0){\rule[-0.200pt]{4.818pt}{0.400pt}}
\put(198,331){\makebox(0,0)[r]{105}}
\put(1416.0,331.0){\rule[-0.200pt]{4.818pt}{0.400pt}}
\put(220.0,440.0){\rule[-0.200pt]{4.818pt}{0.400pt}}
\put(198,440){\makebox(0,0)[r]{110}}
\put(1416.0,440.0){\rule[-0.200pt]{4.818pt}{0.400pt}}
\put(220.0,550.0){\rule[-0.200pt]{4.818pt}{0.400pt}}
\put(198,550){\makebox(0,0)[r]{115}}
\put(1416.0,550.0){\rule[-0.200pt]{4.818pt}{0.400pt}}
\put(220.0,659.0){\rule[-0.200pt]{4.818pt}{0.400pt}}
\put(198,659){\makebox(0,0)[r]{120}}
\put(1416.0,659.0){\rule[-0.200pt]{4.818pt}{0.400pt}}
\put(220.0,768.0){\rule[-0.200pt]{4.818pt}{0.400pt}}
\put(198,768){\makebox(0,0)[r]{125}}
\put(1416.0,768.0){\rule[-0.200pt]{4.818pt}{0.400pt}}
\put(220.0,877.0){\rule[-0.200pt]{4.818pt}{0.400pt}}
\put(198,877){\makebox(0,0)[r]{130}}
\put(1416.0,877.0){\rule[-0.200pt]{4.818pt}{0.400pt}}
\put(220.0,113.0){\rule[-0.200pt]{0.400pt}{4.818pt}}
\put(220,68){\makebox(0,0){25}}
\put(220.0,857.0){\rule[-0.200pt]{0.400pt}{4.818pt}}
\put(342.0,113.0){\rule[-0.200pt]{0.400pt}{4.818pt}}
\put(342,68){\makebox(0,0){30}}
\put(342.0,857.0){\rule[-0.200pt]{0.400pt}{4.818pt}}
\put(463.0,113.0){\rule[-0.200pt]{0.400pt}{4.818pt}}
\put(463,68){\makebox(0,0){35}}
\put(463.0,857.0){\rule[-0.200pt]{0.400pt}{4.818pt}}
\put(585.0,113.0){\rule[-0.200pt]{0.400pt}{4.818pt}}
\put(585,68){\makebox(0,0){40}}
\put(585.0,857.0){\rule[-0.200pt]{0.400pt}{4.818pt}}
\put(706.0,113.0){\rule[-0.200pt]{0.400pt}{4.818pt}}
\put(706,68){\makebox(0,0){45}}
\put(706.0,857.0){\rule[-0.200pt]{0.400pt}{4.818pt}}
\put(828.0,113.0){\rule[-0.200pt]{0.400pt}{4.818pt}}
\put(828,68){\makebox(0,0){50}}
\put(828.0,857.0){\rule[-0.200pt]{0.400pt}{4.818pt}}
\put(950.0,113.0){\rule[-0.200pt]{0.400pt}{4.818pt}}
\put(950,68){\makebox(0,0){55}}
\put(950.0,857.0){\rule[-0.200pt]{0.400pt}{4.818pt}}
\put(1071.0,113.0){\rule[-0.200pt]{0.400pt}{4.818pt}}
\put(1071,68){\makebox(0,0){60}}
\put(1071.0,857.0){\rule[-0.200pt]{0.400pt}{4.818pt}}
\put(1193.0,113.0){\rule[-0.200pt]{0.400pt}{4.818pt}}
\put(1193,68){\makebox(0,0){65}}
\put(1193.0,857.0){\rule[-0.200pt]{0.400pt}{4.818pt}}
\put(1314.0,113.0){\rule[-0.200pt]{0.400pt}{4.818pt}}
\put(1314,68){\makebox(0,0){70}}
\put(1314.0,857.0){\rule[-0.200pt]{0.400pt}{4.818pt}}
\put(1436.0,113.0){\rule[-0.200pt]{0.400pt}{4.818pt}}
\put(1436,68){\makebox(0,0){75}}
\put(1436.0,857.0){\rule[-0.200pt]{0.400pt}{4.818pt}}
\put(220.0,113.0){\rule[-0.200pt]{292.934pt}{0.400pt}}
\put(1436.0,113.0){\rule[-0.200pt]{0.400pt}{184.048pt}}
\put(220.0,877.0){\rule[-0.200pt]{292.934pt}{0.400pt}}
\put(45,495){\makebox(0,0){Grammar size}}
\put(828,23){\makebox(0,0){\% of {\it Test} generated}}
\put(220.0,113.0){\rule[-0.200pt]{0.400pt}{184.048pt}}
\put(1306,812){\makebox(0,0)[r]{}}
\put(259,157){\usebox{\plotpoint}}
\multiput(259.00,157.58)(1.636,0.499){171}{\rule{1.406pt}{0.120pt}}
\multiput(259.00,156.17)(281.082,87.000){2}{\rule{0.703pt}{0.400pt}}
\multiput(543.58,244.00)(0.499,0.537){241}{\rule{0.120pt}{0.530pt}}
\multiput(542.17,244.00)(122.000,129.901){2}{\rule{0.400pt}{0.265pt}}
\multiput(665.58,462.00)(0.499,0.958){157}{\rule{0.120pt}{0.865pt}}
\multiput(664.17,462.00)(80.000,151.205){2}{\rule{0.400pt}{0.433pt}}
\multiput(745.00,615.58)(0.702,0.499){171}{\rule{0.661pt}{0.120pt}}
\multiput(745.00,614.17)(120.628,87.000){2}{\rule{0.330pt}{0.400pt}}
\multiput(867.00,702.58)(2.176,0.499){259}{\rule{1.837pt}{0.120pt}}
\multiput(867.00,701.17)(565.186,131.000){2}{\rule{0.919pt}{0.400pt}}
\put(665.0,375.0){\rule[-0.200pt]{0.400pt}{20.958pt}}
\end{picture}
\caption{\label{curve2} The increase in sentences generated}
\end{figure}

  Notionally, {\it Train} was partitioned into groups of
10  sentences and  processed
incrementally, using the learning configuration C.  After 
dealing with each group, the resulting grammar  was
saved
for subsequent inspection.  This gave 6 grammars,
$G4_1 \ldots G4_6$, where $L(G4_i) \subseteq L(G4_{i+1})$ and 
$G4_4 = G_4$.
The following table shows the growth in grammar size (measured in number of
rules) with respect to training set size:
\begin{center}
\begin{tabular}{|l|l|}  \hline
Grammar & Size  \\ \hline
$G1$ & 97  \\
$G4_1$ & 101      \\
$G4_2$ & 107 \\
$G4_3$ & 111  \\
$G4_4$ & 118 \\
$G4_5$ & 122 \\
$G4_6$ &  128  \\  \hline
\end{tabular}
\end{center}
The  graph \ref{curve} shows the system's learning curve in
terms of the percentage  of sentences parsed in {\it Test}
with respect to training set size (in sentences).
The graph \ref{curve2} shows the increase in  number of
sentences generated  in terms of the percentage 
 of sentences parsed in {\it Test} and the size of the grammar.

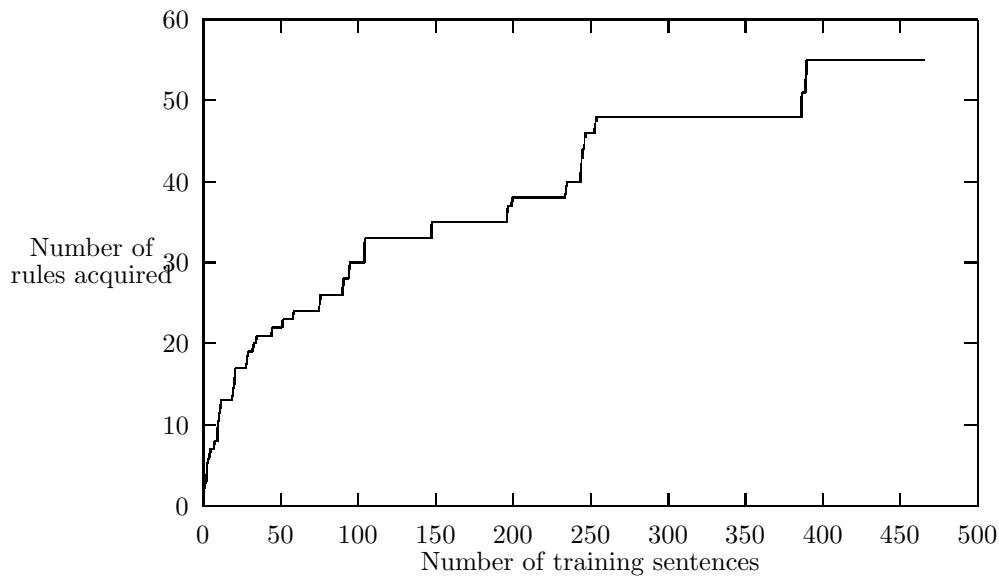
\begin{figure}[htb]
\centering
\setlength{\unitlength}{0.240900pt}
\ifx\plotpoint\undefined\newsavebox{\plotpoint}\fi
\sbox{\plotpoint}{\rule[-0.200pt]{0.400pt}{0.400pt}}%
\begin{picture}(1500,900)(0,0)
\font\gnuplot=cmr10 at 10pt
\gnuplot
\sbox{\plotpoint}{\rule[-0.200pt]{0.400pt}{0.400pt}}%
\put(220.0,113.0){\rule[-0.200pt]{292.934pt}{0.400pt}}
\put(220.0,113.0){\rule[-0.200pt]{0.400pt}{184.048pt}}
\put(220.0,113.0){\rule[-0.200pt]{4.818pt}{0.400pt}}
\put(198,113){\makebox(0,0)[r]{0}}
\put(1416.0,113.0){\rule[-0.200pt]{4.818pt}{0.400pt}}
\put(220.0,240.0){\rule[-0.200pt]{4.818pt}{0.400pt}}
\put(198,240){\makebox(0,0)[r]{10}}
\put(1416.0,240.0){\rule[-0.200pt]{4.818pt}{0.400pt}}
\put(220.0,368.0){\rule[-0.200pt]{4.818pt}{0.400pt}}
\put(198,368){\makebox(0,0)[r]{20}}
\put(1416.0,368.0){\rule[-0.200pt]{4.818pt}{0.400pt}}
\put(220.0,495.0){\rule[-0.200pt]{4.818pt}{0.400pt}}
\put(198,495){\makebox(0,0)[r]{30}}
\put(1416.0,495.0){\rule[-0.200pt]{4.818pt}{0.400pt}}
\put(220.0,622.0){\rule[-0.200pt]{4.818pt}{0.400pt}}
\put(198,622){\makebox(0,0)[r]{40}}
\put(1416.0,622.0){\rule[-0.200pt]{4.818pt}{0.400pt}}
\put(220.0,750.0){\rule[-0.200pt]{4.818pt}{0.400pt}}
\put(198,750){\makebox(0,0)[r]{50}}
\put(1416.0,750.0){\rule[-0.200pt]{4.818pt}{0.400pt}}
\put(220.0,877.0){\rule[-0.200pt]{4.818pt}{0.400pt}}
\put(198,877){\makebox(0,0)[r]{60}}
\put(1416.0,877.0){\rule[-0.200pt]{4.818pt}{0.400pt}}
\put(220.0,113.0){\rule[-0.200pt]{0.400pt}{4.818pt}}
\put(220,68){\makebox(0,0){0}}
\put(220.0,857.0){\rule[-0.200pt]{0.400pt}{4.818pt}}
\put(342.0,113.0){\rule[-0.200pt]{0.400pt}{4.818pt}}
\put(342,68){\makebox(0,0){50}}
\put(342.0,857.0){\rule[-0.200pt]{0.400pt}{4.818pt}}
\put(463.0,113.0){\rule[-0.200pt]{0.400pt}{4.818pt}}
\put(463,68){\makebox(0,0){100}}
\put(463.0,857.0){\rule[-0.200pt]{0.400pt}{4.818pt}}
\put(585.0,113.0){\rule[-0.200pt]{0.400pt}{4.818pt}}
\put(585,68){\makebox(0,0){150}}
\put(585.0,857.0){\rule[-0.200pt]{0.400pt}{4.818pt}}
\put(706.0,113.0){\rule[-0.200pt]{0.400pt}{4.818pt}}
\put(706,68){\makebox(0,0){200}}
\put(706.0,857.0){\rule[-0.200pt]{0.400pt}{4.818pt}}
\put(828.0,113.0){\rule[-0.200pt]{0.400pt}{4.818pt}}
\put(828,68){\makebox(0,0){250}}
\put(828.0,857.0){\rule[-0.200pt]{0.400pt}{4.818pt}}
\put(950.0,113.0){\rule[-0.200pt]{0.400pt}{4.818pt}}
\put(950,68){\makebox(0,0){300}}
\put(950.0,857.0){\rule[-0.200pt]{0.400pt}{4.818pt}}
\put(1071.0,113.0){\rule[-0.200pt]{0.400pt}{4.818pt}}
\put(1071,68){\makebox(0,0){350}}
\put(1071.0,857.0){\rule[-0.200pt]{0.400pt}{4.818pt}}
\put(1193.0,113.0){\rule[-0.200pt]{0.400pt}{4.818pt}}
\put(1193,68){\makebox(0,0){400}}
\put(1193.0,857.0){\rule[-0.200pt]{0.400pt}{4.818pt}}
\put(1314.0,113.0){\rule[-0.200pt]{0.400pt}{4.818pt}}
\put(1314,68){\makebox(0,0){450}}
\put(1314.0,857.0){\rule[-0.200pt]{0.400pt}{4.818pt}}
\put(1436.0,113.0){\rule[-0.200pt]{0.400pt}{4.818pt}}
\put(1436,68){\makebox(0,0){500}}
\put(1436.0,857.0){\rule[-0.200pt]{0.400pt}{4.818pt}}
\put(220.0,113.0){\rule[-0.200pt]{292.934pt}{0.400pt}}
\put(1436.0,113.0){\rule[-0.200pt]{0.400pt}{184.048pt}}
\put(220.0,877.0){\rule[-0.200pt]{292.934pt}{0.400pt}}
\put(45,495){\makebox(0,0){\shortstack{Number of \\ rules  acquired}}}
\put(828,23){\makebox(0,0){Number of training sentences}}
\put(220.0,113.0){\rule[-0.200pt]{0.400pt}{184.048pt}}
\multiput(222.61,138.00)(0.447,2.695){3}{\rule{0.108pt}{1.833pt}}
\multiput(221.17,138.00)(3.000,9.195){2}{\rule{0.400pt}{0.917pt}}
\put(225.17,151){\rule{0.400pt}{5.300pt}}
\multiput(224.17,151.00)(2.000,15.000){2}{\rule{0.400pt}{2.650pt}}
\multiput(227.61,177.00)(0.447,2.472){3}{\rule{0.108pt}{1.700pt}}
\multiput(226.17,177.00)(3.000,8.472){2}{\rule{0.400pt}{0.850pt}}
\put(230.17,189){\rule{0.400pt}{2.700pt}}
\multiput(229.17,189.00)(2.000,7.396){2}{\rule{0.400pt}{1.350pt}}
\put(237.17,202){\rule{0.400pt}{2.700pt}}
\multiput(236.17,202.00)(2.000,7.396){2}{\rule{0.400pt}{1.350pt}}
\put(232.0,202.0){\rule[-0.200pt]{1.204pt}{0.400pt}}
\put(242.17,215){\rule{0.400pt}{5.100pt}}
\multiput(241.17,215.00)(2.000,14.415){2}{\rule{0.400pt}{2.550pt}}
\multiput(244.61,240.00)(0.447,5.597){3}{\rule{0.108pt}{3.567pt}}
\multiput(243.17,240.00)(3.000,18.597){2}{\rule{0.400pt}{1.783pt}}
\put(247.17,266){\rule{0.400pt}{2.700pt}}
\multiput(246.17,266.00)(2.000,7.396){2}{\rule{0.400pt}{1.350pt}}
\put(239.0,215.0){\rule[-0.200pt]{0.723pt}{0.400pt}}
\multiput(266.61,279.00)(0.447,5.374){3}{\rule{0.108pt}{3.433pt}}
\multiput(265.17,279.00)(3.000,17.874){2}{\rule{0.400pt}{1.717pt}}
\put(269.17,304){\rule{0.400pt}{5.100pt}}
\multiput(268.17,304.00)(2.000,14.415){2}{\rule{0.400pt}{2.550pt}}
\put(249.0,279.0){\rule[-0.200pt]{4.095pt}{0.400pt}}
\multiput(288.61,329.00)(0.447,5.597){3}{\rule{0.108pt}{3.567pt}}
\multiput(287.17,329.00)(3.000,18.597){2}{\rule{0.400pt}{1.783pt}}
\put(271.0,329.0){\rule[-0.200pt]{4.095pt}{0.400pt}}
\put(298.17,355){\rule{0.400pt}{2.700pt}}
\multiput(297.17,355.00)(2.000,7.396){2}{\rule{0.400pt}{1.350pt}}
\put(291.0,355.0){\rule[-0.200pt]{1.686pt}{0.400pt}}
\put(303.17,368){\rule{0.400pt}{2.500pt}}
\multiput(302.17,368.00)(2.000,6.811){2}{\rule{0.400pt}{1.250pt}}
\put(300.0,368.0){\rule[-0.200pt]{0.723pt}{0.400pt}}
\put(327.17,380){\rule{0.400pt}{2.700pt}}
\multiput(326.17,380.00)(2.000,7.396){2}{\rule{0.400pt}{1.350pt}}
\put(305.0,380.0){\rule[-0.200pt]{5.300pt}{0.400pt}}
\put(344.17,393){\rule{0.400pt}{2.700pt}}
\multiput(343.17,393.00)(2.000,7.396){2}{\rule{0.400pt}{1.350pt}}
\put(329.0,393.0){\rule[-0.200pt]{3.613pt}{0.400pt}}
\put(361.17,406){\rule{0.400pt}{2.700pt}}
\multiput(360.17,406.00)(2.000,7.396){2}{\rule{0.400pt}{1.350pt}}
\put(346.0,406.0){\rule[-0.200pt]{3.613pt}{0.400pt}}
\multiput(402.61,419.00)(0.447,5.374){3}{\rule{0.108pt}{3.433pt}}
\multiput(401.17,419.00)(3.000,17.874){2}{\rule{0.400pt}{1.717pt}}
\put(363.0,419.0){\rule[-0.200pt]{9.395pt}{0.400pt}}
\put(439.17,444){\rule{0.400pt}{5.300pt}}
\multiput(438.17,444.00)(2.000,15.000){2}{\rule{0.400pt}{2.650pt}}
\put(405.0,444.0){\rule[-0.200pt]{8.191pt}{0.400pt}}
\put(449.17,470){\rule{0.400pt}{5.100pt}}
\multiput(448.17,470.00)(2.000,14.415){2}{\rule{0.400pt}{2.550pt}}
\put(441.0,470.0){\rule[-0.200pt]{1.927pt}{0.400pt}}
\put(473.17,495){\rule{0.400pt}{7.700pt}}
\multiput(472.17,495.00)(2.000,22.018){2}{\rule{0.400pt}{3.850pt}}
\put(451.0,495.0){\rule[-0.200pt]{5.300pt}{0.400pt}}
\put(578.17,533){\rule{0.400pt}{5.300pt}}
\multiput(577.17,533.00)(2.000,15.000){2}{\rule{0.400pt}{2.650pt}}
\put(475.0,533.0){\rule[-0.200pt]{24.813pt}{0.400pt}}
\put(697.17,559){\rule{0.400pt}{5.100pt}}
\multiput(696.17,559.00)(2.000,14.415){2}{\rule{0.400pt}{2.550pt}}
\put(580.0,559.0){\rule[-0.200pt]{28.185pt}{0.400pt}}
\put(704.17,584){\rule{0.400pt}{2.700pt}}
\multiput(703.17,584.00)(2.000,7.396){2}{\rule{0.400pt}{1.350pt}}
\put(699.0,584.0){\rule[-0.200pt]{1.204pt}{0.400pt}}
\multiput(789.61,597.00)(0.447,5.374){3}{\rule{0.108pt}{3.433pt}}
\multiput(788.17,597.00)(3.000,17.874){2}{\rule{0.400pt}{1.717pt}}
\put(706.0,597.0){\rule[-0.200pt]{19.995pt}{0.400pt}}
\multiput(813.61,622.00)(0.447,11.179){3}{\rule{0.108pt}{6.900pt}}
\multiput(812.17,622.00)(3.000,36.679){2}{\rule{0.400pt}{3.450pt}}
\put(792.0,622.0){\rule[-0.200pt]{5.059pt}{0.400pt}}
\multiput(818.61,673.00)(0.447,5.597){3}{\rule{0.108pt}{3.567pt}}
\multiput(817.17,673.00)(3.000,18.597){2}{\rule{0.400pt}{1.783pt}}
\put(816.0,673.0){\rule[-0.200pt]{0.482pt}{0.400pt}}
\multiput(835.61,699.00)(0.447,5.374){3}{\rule{0.108pt}{3.433pt}}
\multiput(834.17,699.00)(3.000,17.874){2}{\rule{0.400pt}{1.717pt}}
\put(821.0,699.0){\rule[-0.200pt]{3.373pt}{0.400pt}}
\put(1159.17,724){\rule{0.400pt}{7.700pt}}
\multiput(1158.17,724.00)(2.000,22.018){2}{\rule{0.400pt}{3.850pt}}
\put(838.0,724.0){\rule[-0.200pt]{77.329pt}{0.400pt}}
\put(1166.17,762){\rule{0.400pt}{10.300pt}}
\multiput(1165.17,762.00)(2.000,29.622){2}{\rule{0.400pt}{5.150pt}}
\put(1161.0,762.0){\rule[-0.200pt]{1.204pt}{0.400pt}}
\put(1168.0,813.0){\rule[-0.200pt]{44.566pt}{0.400pt}}
\end{picture}
\caption{\label{big} The learning curve when using more training sentences}
\end{figure}

As can be seen, convergence increases   at a varying rate.
Because of the small size of training material, nothing can be said
about how much material is required for the system to converge.\footnote{Subsequent to the experiments reported in this thesis, the learner was configured to
use model-based learning and was trained on 466 sentences taken from the SEC.  This grammar parsed $98.3\%$ of {\it Test}, $31\%$ of {\it Bad} and 
atttained a mean plausibility rating (against 48 sentences) of $0.102$.  The
convergence garph is shown in figure \ref{big}.  These results indicate that
the results obtained from testing and training using 
small amounts of data are not that far from those
obtained with larger testing and training sets.}
  However,
it is clear that only a moderate amount of material is necessary to
allow a large increase in coverage.

This concludes the experiments.  It is now time to draw some conclusions.
\section{Discussion \label{evaldisc}}
The hypothesis, we tested by the experiments, 
 was that combining both learning styles together would produce
a quantitatively better grammar than would be
 produced when using either learning
style in isolation.

From the plausibility results, we saw that $G4$ was the most plausible,
ranked equal first for the undergeneration test, and equal last
for the overgeneration test.  Hence, at least
for plausibility, using both learning styles together produced
a grammar that was  quantitatively better than was produced when using
either learning style in isolation.  For the other tests, the outcome
was less clear.  $G4$ certainly did well at dealing with undergeneration,
but this was equalled by the performance of $G2$.  $G4$ also faired
equally badly as $G2$ for the overgeneration test.  Both of these
results  show that, at
least for the experiments reported here, the model, through incompleteness, 
 made little impact upon the
string sets generated by the  grammars learnt.  The results also 
show that 
 data-driven learning has compensated for
model-based incompleteness.  That is, $G4$ performance
was equal to $G2$, despite the fact that $G4$ also used a model as part
of the learning process.  
  Interestingly enough, there is a correlation
between the size of the learnt grammars, in terms of
disjunctive rules,  and the degree of undergeneration
and overgeneration: the larger the grammar, the larger the string set
generated by that grammar.  All the learnt grammars are larger than
grammar $G1$, and indeed, they all  undergenerate less than $G1$.
From the convergence results, little can be said, other than that they 
hint that the
system does not require much training material in order to allow a wide
covering grammar to be learnt.

The conclusion from these experiments is that grammar learning does
 meet the success criteria for natural language grammars, but that this
success is mainly due to data-driven learning.  When model-based learning does have
 a role to play, it
outperforms data-driven learning.  Overall, using data-driven learning and model-based learning together
is better than using data-driven learning or model-based learning in isolation.

\chapter{Conclusions}

\section{Introduction}

This thesis described work on the machine learning of plausible
unification-based grammars.  In particular, the  motivation
of the work was to
 deal with the problem of undergeneration.  Empirically, it was found that
machine learning could deal with undergeneration, and in particular, that
the combined use of data-driven learning and model-based learning produced more plausible 
grammars than when using either learning style in isolation.  Other advances
reported in this thesis include:
\begin{center}
\begin{itemize}
\item  a detailed discussion of what undergeneration is, why it occurs
and how it can be successfully dealt with.
\item a system that is capable of correctly treating undergeneration, is
noise resistant, can learn about ambiguity, uses a parsimonious grammar
formalism, is capable of being used in conjunction with a sentence
corrector and  allows experimentation between model-based and
data-driven learning.
\item an evaluation methodology for measuring grammar quality that is
stricter than other approaches reported in the literature.
\end{itemize}
\end{center}

In order to achieve these results, numerous assumptions have been made.  
Section \ref{ass} discusses these and considers which assumptions can be
weakened, and which must be held.  Section \ref{further} outlines ways in which
grammar learning can be enhanced as a solution to the problem of dealing
with undergeneration.  Finally, section \ref{rant} ends the thesis with some
general comments about the field of grammar learning.  
\section{Assumptions \label{ass}}

The assumptions made in this thesis can be divided into two groups:
those made for practical reasons, and those made for theoretical
reasons.  The former group have a conjectural status, in that they can
be given-up or altered, whilst the latter group have an axiomatic status, 
in that they cannot be given up so easily.

Assumptions in the former group include:
\begin{center}
\begin{itemize}
\item Having a complete lexicon.
\item Using unary and binary rules.
\end{itemize}
\end{center}

The system presupposes that each word encountered was present within
the lexicon.  In practise, this meant using a stochastic tagger to
assign a part-of-speech to any word that the system encountered.  
What is 
unsatisfactory about using a stochastic tagger 
is the coarse nature of the tag sets used.  For example, 
the CLAWS2a tagset \cite{Blac93}, which is used to label words in the Spoken 
English Corpus, lacks verbal subcategorisation and so grammars learnt 
will  overgenerate more than they need to.   Furthermore, lexical taggers
tend to suffer degraded performance when tagging corpora other than the
corpus used to train the tagger, thereby limiting the choice of 
language that can be successfully processed \cite{Mete91}.  An
avoidance of this coarseness problem and the lack of transportability 
 would be to use a richer
lexical representation (for example that used by 
the ACQUILEX Project \cite{Sanf91}).  This still runs the risk that
the lexicon is incomplete.  
A method of ensuring that
the lexicon was both complete, and also sufficiently rich, would be to learn
the necessary entries.  There has been some work in this area (for example
Russell's thesis \cite{Russ93}) which could be used to weaken the 
complete lexicon assumption.  

There are a variety of ways of weakening the assumption that rules
are either unary or binary.  A first approach might be to fold
 rules that are frequently used together.\footnote{Samuelsson and Rayner's
use of EBL is similar to rule folding \cite{Samu91}.}   For example, assuming
only using binary super rules, the system might learn rules 
for  ditransitives such as 
 \psr{VP}{V1~NP} and \psr{V1}{V0~NP}.  These could then
be collapsed into the single rule \psr{VP}{V0~NP~NP}.  This has the
advantage that the number of super rules are minimised (i.\ e.\ there
is no need for a super rule with three RHS categories),  which reduces
the search space.  However, it makes the rule construction
less declarative.  The number of categories in the learnt
RHS's will no longer exactly relate to the number of categories
in the instantiated super rule. A second approach might be to expand the
super rules to include rules with more categories in their RHS.  Learning 
would then proceed as before, except that this time, the learner might try to
relate the length of rule RHSs with plausibility.  This information would then
be used to reduce the super rule set.  However, expanding the
super rules  would 
increase the search space.

To learn about gapping, the system could use a 0-ary super rule.  This
would need to be refined or rejected by appropriate principles of 
grammaticality.  As should be apparent, gap learning is a hard problem.  
One reason for this difficulty is that a sentence with gaps can be viewed as
a longer version of that same sentence without gaps:
\numberSentence{What did Tony have accepted today? \label{xx}}
\numberSentence{What did Tony have  accepted today e? \label{xx2}}
Here, sentence \ref{xx} has a length of 6 words, whilst sentence \ref{xx2}
has a length of 7 words (including the  gap).  Since gaps
 can appear almost anywhere in the sentence, and  from
 section \ref{complexity}, we saw that grammar learning is at least exponential
with respect to sentence length, gap learning will introduce a 
dramatic expansion of the search space.  Another reason for the difficulty
in learning about unbounded dependencies  is that  gaps are hypothesised in
relation to 
non-local, potentially erroneous information in the parse tree.  This 
means that the system will wastefully construct gaps 
based upon structures that are 
later thrown away.  The only way to deal with both of these sources of
intractability involved with gap 
learning would be to use a strong model of grammaticality.

Assumptions in the latter group include:
\begin{center}
\begin{itemize}
\item Using a formalism that is (at least) context free.
\item Syntax is important for NLP.
\item Learning competence, and not performance grammars.
\end{itemize}
\end{center}

The system does not learn nonterminal symbols, and hence cannot 
be used to learn the formal  complexity of the language being learnt.
So, the system cannot learn if the language is finite state, context
free, or any of the other classes of language in the Chomsky hierarchy.
Hence, it is necessary to fix the 
computational power of the grammar formalism prior to learning.  Fixing the
grammar formalism's power is a claim on the power of the language being learnt.
In this work, natural languages are stated  as being (at least) context free.

Another assumption is that
syntax is important in its own right, and has its own role to play.  Quite
apart from contemporary theories of semantic interpretation (which require a
distinct, separate grammar), other aspects of applied linguistics 
require grammars.   For example, corpora can be parsed and used 
as linguistic databases for exploration.  Robust grammar checkers, 
which are useful
when teaching first and second languages to students, require a grammar.  
Grammars can therefore be assumed as being important.  

The final assumption is the most problematic.  As was stated in various parts
of this thesis, the learner tries to learn competence and not performance
grammars.  As should be clear, making this distinction is controversial.
The defence of learning a competence grammar is that a distinction can be
made between ungrammatical and grammatical sentences.  We contend that
a grammar that fails to make this distinction becomes a vacuous theory
and of little value.  However, it is not clear if such a distinction
can always be made, and hence, it could be said that the learner at times makes
an arbitrary decision regarding the grammaticality/ungrammaticality 
distinction.  Whilst this may be so, the converse, of allowing the learner
to acquire performance grammars, is even less satisfactory. For example,
a performance grammar trained using Scottish speakers would reflect
Scottish performance.  Such a grammar would need to be retrained if it
were subsequently to be used in  processing the language of (say) 
Welsh speakers.
This would not be the case for a competence grammar.   Hence, the task
of  acquisition is assumed to be one of learning competence, and not
performance  grammars.

\section{Further work \label{further}}
There are numerous ways of extending this work.  As was clear from the
experiments in chapter six, the model of grammaticality had little impact
upon the quality of the grammars.  Hence, an obvious step would be to
increase the contribution that the model can make.  Areas of incompleteness
could be identified by analysing the contributions that each of the principles
of grammaticality had to make, and noting areas where they were deficient.
Another step would be to use other principles of grammaticality, drawn
from theories such as Government and Binding Theory \cite{Chom81} .  For 
example, the learner does not deductively 
learn about
long distance dependencies.  Using GB principles such as subjacency and
Move-$\alpha$ would be a way of achieving this.  

The learner could try to use textuality as a constraining
device upon the quality of grammars.  It is a known fact that  the ability 
to predict which
sentence is to come next increases with the number of prior
sentences considered \cite{Shan51}.  Hence, unexpected  analyses could be 
detected
and rejected, thereby preventing the learner from acquiring implausible
grammars.  As far as is known,  considering textuality when learning
grammars is completely novel.  Related to the use of textuality would be
using punctuation as another constraining device.  Jones shows how punctuation
can help reduce the syntactic ambiguity in long sentences \cite{Jone94}, and
 so punctuation
would help to identify constituents when learning.  

Turning to 
data-driven learning, the language model would benefit from using a better
approximation of the Noisy Channel Model.  This could be achieved by using
lexical co-occurrence statistics, by trying to reduce the entropy of the
language generated by the learnt grammar, by using a more principled
decision theory, and so on.  

So far, the further work mentioned has only considered the two learning styles.
It would also be possible to exploit the grammar formalism and try to
generalise the rules.  As it stands, the learner constructs rules that, being
the result of unification, might at times be too specific.  Interesting work
would be to try to determine which of the features in the rules could be 
uninstantiated, and which of these features should be re-entrant.

The learner used a {\it rule}-based grammar formalism.  An interesting possibility
would be to consider using instead a {\it lexically}-based formalism.  This
would have  the advantage of  allowing  syntactic,
semantic, and pragmatic constraints to be expressed within the single
framework \cite{Pere92}.  

Finally, it will be worthwhile to consider how the
language learner relates to theories of human language acquisition.  For
example, one could view the data-driven component as corresponding to
a set of parameters that need to be set, and the model-based component as
corresponding to a set of principles.  The experiments might therefore
shed light on the relationship between these two aspects of language
learning.

\section{General conclusions \label{rant}}

When this research started just under three years ago, not many 
grammar learning researchers were considering what linguistics had to
offer.  Now, this is no longer true, and  the
grammar learning community are tentatively beginning to  
 use linguistic theories of universal
grammar.   One of the contributions of this thesis has been to show
that a whole-hearted use of model-based learning can overcome the problems 
associated with data-driven grammar learning.  Hence, data-driven approaches
such as the Inside-Outside algorithm will benefit greatly from using 
model-based learning.  Completeness can be combined with quality, thereby allowing
wide covering grammars to be constructed that lend themselves to semantic
interpretation.  Hopefully, gone will be the days when researchers 
use just data-driven learning, coupled with inadequate formalisms, to try to acquire
grammars that cannot be used for any task other than for language
recognition purposes.

\appendix                 

\chapter{The Lexicon \label{lex}}

The experiments reported in chapter seven used the SEC as a source of
 training
and testing material.  However, in common with other researchers, the
Grammar Garden parsed SEC tag sequences, and not SEC sentences.  This is for 
logistical reasons: it is far easier to create a lexicon of a few
hundred tags than it is to create a lexicon of millions of words.  So,
a  SEC entry:
\begin{verbatim}
[N It_PPH1 N]
[V 's_VBZ [N a_AT1 useful_JJ reminder_NN1 
             [Fn that_CST 
                 [N some_DD scientists_NN2 N]
                 [V find_VV0 [N Don_NP1 Cupitt_NP1 N]
                 unscientific_JJ V]Fn]N]V] ._.
\end{verbatim}
 would be preprocessed into the tag sequence:
\numberSentence{PPH1 VBZ AT1 JJ NN1  CST DD NN2 VV0 NP1 NP1 JJ}
for use by the Grammar Garden.

Grammar $G1$ contained the following lexicon: \\
\lexical{\$}{\fs{N +, V -, BAR 0, MINOR NONE, POSS +, NTYPE POSS, \\ WH -, CONJ -}} \\
\lexical{APP\$}  {\fs{MINOR DET, POSS +,  WH -}} \\ 
\lexical{AT}  {\fs{MINOR DET, POSS -,  WH -}} \\ 
\lexical{AT1}  {\fs{MINOR DET, PLU -, POSS -,  WH -}} \\ 
\lexical{BCS}  {\fs{N -, V -, BAR 0, MINOR NONE, SUBCAT SCOMP}} \\ 
\lexical{BTO}  {\fs{N -, V -, BAR 0, MINOR NONE, SUBCAT VPINF}} \\ 
\lexical{CC}  {\fs{MINOR CONJ, CJTYPE END}} \\ 
\lexical{CCB}  {\fs{MINOR CONJ, CJTYPE END}} \\ 
\lexical{CF}  {\fs{N -, V -, BAR 0, MINOR NONE, SUBCAT SFIN, CONJ -}} \\ 
\lexical{CS}  {\fs{N -, V -, BAR 0, MINOR NONE, SUBCAT SFIN,  CONJ -}} \\ 
\lexical{CSA}  {\fs{N -, V -, BAR 0, MINOR NONE,  PFORM AS, CONJ -}} \\ 
\lexical{CSN}  {\fs{N -, V -, BAR 0, MINOR NONE,  PFORM THAN, CONJ -}} \\ 
\lexical{CST}  {\fs{N -, V -, BAR 0, MINOR NONE, SUBCAT SFIN, \\ PFORM THAT, CONJ -}} \\ 
\lexical{CSW}  {\fs{N -, V -, BAR 0, MINOR NONE, SUBCAT SFIN, \\ PFORM WH, CONJ -}} \\
\lexical{CSW}   {\fs{N -, V -, BAR 0, MINOR NONE, SUBCAT VPINF, \\ PFORM WH, CONJ -}} \\ 
\lexical{DA}  {\fs{N +, V +, BAR 0, MINOR NONE, ATYPE ATT, \\ AFORM NONE, ADV -, CONJ -}} \\
\lexical{DA}   {\fs{N +, V -, BAR 0, MINOR NONE, POSS -, NTYPE PRO, \\ WH -, CONJ -}} \\
\lexical{DA}   {\fs{N +, V -, BAR 2, MINOR NONE, POSS -, NTYPE PRO, \\ WH -, CONJ -}} \\ 
\lexical{DA}   {\fs{MINOR DET, POSS -, WH -}} \\ 
\lexical{DA1}  {\fs{N +, V +, BAR 0, MINOR NONE, ATYPE ATT, \\ AFORM NONE,  ADV -, CONJ -}} \\
\lexical{DA1}   {\fs{N +, V -, BAR 0, MINOR NONE, PLU -, POSS -, NTYPE PRO, \\ WH -, CONJ -}} \\ 
\lexical{DA1}   {\fs{N +, V -, BAR 2, MINOR NONE, PLU -, POSS -, NTYPE PRO,\\  WH -, CONJ - }} \\
\lexical{DA1}   {\fs{MINOR DET, PLU -, POSS -, \\ WH -}} \\ 
\lexical{DA2}  {\fs{N +, V +, BAR 0, MINOR NONE, ATYPE ATT, \\ AFORM NONE, ADV -, CONJ -}} \\
\lexical{DA2}   {\fs{N +, V -, BAR 0, MINOR NONE, PLU +, POSS -, NTYPE PRO, \\ WH -, CONJ -}} \\
\lexical{DA2}   {\fs{N +, V -, BAR 2, MINOR NONE, PLU +, POSS -, NTYPE PRO, \\ WH -, CONJ -}} \\
\lexical{DA2}   {\fs{MINOR DET, PLU +, POSS -,  WH -}} \\ 
\lexical{DA2R}  {\fs{N +, V +, BAR 0, MINOR NONE, AFORM ER, ADV -, CONJ -}} \\
\lexical{DA2R}   {\fs{N +, V -, BAR 0, MINOR NONE, PLU +, POSS -, \\ NTYPE PRO,  WH -, CONJ -}} \\
\lexical{DA2R}  {\fs{N +, V -, BAR 2, MINOR NONE, PLU +, POSS -, \\ NTYPE PRO,  WH -, CONJ -}} \\
\lexical{DA2R}   {\fs{MINOR DET, PLU +, POSS -,  WH -}} \\ 
\lexical{DAR}  {\fs{N +, V +, BAR 0, MINOR NONE, AFORM ER, ADV -, CONJ -}} \\ 
 \lexical{DAR}  {\fs{N +, V -, BAR 0, MINOR NONE, PLU +, POSS -, NTYPE PRO, \\ WH -, CONJ -}} \\ 
 \lexical{DAR}  {\fs{N +, V -, BAR 2, MINOR NONE, PLU +, POSS -, NTYPE PRO, \\ WH -, CONJ -}} \\ 
\lexical{DAR}   {\fs{MINOR DET, PLU +, POSS -,  WH -}} \\ 
\lexical{DAT}  {\fs{N +, V +, BAR 0, MINOR NONE, AFORM EST, ADV -, CONJ -}} \\ 
\lexical{DAT}   {\fs{N +, V -, BAR 0, MINOR NONE, PLU +, POSS -,  NTYPE PRO, \\ WH -, CONJ -}} \\ 
 \lexical{DAT}  {\fs{N +, V -, BAR 2, MINOR NONE, PLU +, POSS -,  NTYPE PRO, \\ WH -, CONJ -}} \\ 
\lexical{DAT}   {\fs{MINOR DET, PLU +, POSS -,  WH -}} \\ 
\lexical{DB}  {\fs{N +, V -, BAR 0, MINOR NONE, POSS -, NTYPE PART, \\ WH -, CONJ -}} \\ 
\lexical{DB}   {\fs{N +, V -, BAR 2, MINOR NONE, POSS -, NTYPE PRO, \\ WH -, CONJ -}} \\ 
\lexical{DB2}  {\fs{N +, V -, BAR 0, MINOR NONE, PLU +, POSS -, NTYPE PART, \\ WH -, CONJ -}} \\ 
 \lexical{DB2}  {\fs{N +, V -, BAR 2, MINOR NONE, PLU +, POSS -, NTYPE PRO, \\ WH -, CONJ -}} \\ 
\lexical{DD}  {\fs{MINOR DET, POSS -,  WH -}} \\ 
 \lexical{DD}  {\fs{N +, V -, BAR 2, MINOR NONE, PLU +, POSS -, NTYPE PRO, \\ WH -, CONJ -}} \\ 
\lexical{DD1}  {\fs{MINOR DET, PLU -, POSS -,  WH -}} \\ 
 \lexical{DD1}  {\fs{N +, V -, BAR 2, MINOR NONE, PLU -, POSS -, NTYPE PRO, \\ WH -, CONJ -}} \\ 
\lexical{DD2}  {\fs{MINOR DET, PLU +, POSS -,  WH -}} \\ 
 \lexical{DD2}  {\fs{N +, V -, BAR 2, MINOR NONE, PLU +, POSS -, NTYPE PRO, \\ WH -, CONJ -}} \\ 
\lexical{DDQ}  {\fs{MINOR DET, POSS -,  WH +}} \\ 
\lexical{DDQ\$}  {\fs{MINOR DET, POSS -,  WH +}} \\ 
\lexical{DDQV}  {\fs{MINOR DET, POSS -,  WH +}} \\ 
\lexical{EX}  {\fs{N +, V -, BAR 2, MINOR NONE, POSS -, NTYPE THERE, \\ WH -, CONJ -}} \\ 
\lexical{ICS}  {\fs{N -, V -, BAR 0, MINOR NONE, SUBCAT SFIN}} \\ 
\lexical{ICS}   {\fs{N -, V -, BAR 0, MINOR NONE, SUBCAT VPING, CONJ -}} \\ 
\lexical{IF}  {\fs{N -, V -, BAR 0, MINOR NONE, \\ SUBCAT SINF, PFORM FOR, CONJ -}} \\ 
\lexical{IF}   {\fs{N -, V -, BAR 0, MINOR NONE, \\ SUBCAT NP, PFORM FOR, CONJ -}} \\ 
\lexical{II}  {\fs{N -, V -, BAR 0, MINOR NONE,  CONJ -}} \\ 
\lexical{IO}  {\fs{N -, V -, BAR 0, MINOR NONE, SUBCAT NP, PFORM OF, CONJ -}} \\ 
\lexical{IO}   {\fs{N -, V -, BAR 0, MINOR NONE, SUBCAT VPING, \\ PFORM OF, CONJ -}} \\ 
\lexical{IW}  {\fs{N -, V -, BAR 0, MINOR NONE, SUBCAT NP, \\ PFORM WITH, CONJ -}} \\ 
 \lexical{IW}  {\fs{N -, V -, BAR 0, MINOR NONE, SUBCAT VPING, \\ PFORM WITHOUT, CONJ -}} \\ 
\lexical{JA}  {\fs{N +, V +, BAR 0, MINOR NONE, ATYPE PRD, AFORM NONE, \\ADV -, CONJ -}} \\ 
 \lexical{JA}  {\fs{N +, V +, BAR 1, MINOR NONE, ATYPE PRD, AFORM NONE, \\ ADV -, CONJ -}} \\ 
\lexical{JB}  {\fs{N +, V +, BAR 0, MINOR NONE, ATYPE ATT, AFORM NONE, \\ ADV -, CONJ -}} \\ 
\lexical{JB}   {\fs{N +, V +, BAR 1, MINOR NONE, ATYPE ATT, AFORM NONE, \\ ADV -, CONJ -}} \\ 
\lexical{JBR}  {\fs{N +, V +, BAR 1, MINOR NONE, ATYPE ATT, AFORM ER, \\ ADV -, CONJ -}} \\ 
\lexical{JBT}  {\fs{N +, V +, BAR 1, MINOR NONE, ATYPE ATT, AFORM EST, \\ ADV -, CONJ -}} \\ 
\lexical{JJ}  {\fs{N +, V +, BAR 0, MINOR NONE, AFORM NONE, ADV -, CONJ -}} \\
\lexical{JJ}   {\fs{N +, V +, BAR 1, MINOR NONE, AFORM NONE, ADV -, CONJ -}} \\ 
\lexical{JJR}  {\fs{N +, V +, BAR 0, MINOR NONE, AFORM ER, ADV -, CONJ -}} \\ 
\lexical{JJR}   {\fs{N +, V +, BAR 1, MINOR NONE, AFORM ER, ADV -, CONJ -}} \\
\lexical{JJT}  {\fs{N +, V +, BAR 0, MINOR NONE, AFORM EST, ADV -, CONJ -}} \\ 
 \lexical{JJT}   {\fs{N +, V +, BAR 1, MINOR NONE, AFORM EST, ADV -, CONJ -}} \\ 
\lexical{JK}  {\fs{N +, V +, BAR 0, MINOR NONE, ATYPE CAT, \\ AFORM NONE, ADV -, CONJ -}} \\ 
\lexical{LE}  {\fs{MINOR CONJ, CJTYPE BEGIN}} \\ 
\lexical{MC}  {\fs{N +, V -, BAR 2, MINOR NONE, POSS -, NTYPE NUM,  WH -, CONJ -}} \\ 
\lexical{MC}   {\fs{MINOR DET, PLU +, POSS -,  WH -}} \\ 
\lexical{MC\$}  {\fs{N +, V -, BAR 2, MINOR NONE, PLU +,  POSS +,  NTYPE NUM,  WH -, CONJ -}} \\ 
\lexical{MC\$}   {\fs{MINOR DET, PLU +, POSS +, \\ WH -}} \\ 
\lexical{MC-MC } 
   {\fs{N +, V -, BAR 2, MINOR NONE, PLU +, POSS -, NTYPE NUM, \\ WH -, CONJ -}} \\ 
 \lexical{MC-MC }   {\fs{MINOR DET, PLU +, POSS +,  WH -}} \\ 
\lexical{MC1}  {\fs{N +, V -, BAR 2, MINOR NONE, PLU -, POSS -, \\ NTYPE NUM,  WH -, CONJ -}} \\ 
\lexical{MC1}   {\fs{MINOR DET, PLU -, POSS +,  WH -}} \\ 
\lexical{MC2}  {\fs{N +, V -, BAR 2, MINOR NONE, PLU +, POSS -, \\ NTYPE NUM,  WH -, CONJ -}} \\ 
\lexical{MC2}   {\fs{MINOR DET, PLU +, POSS +,  WH -}} \\ 
\lexical{MD}  {\fs{N +, V +, BAR 0, MINOR NONE, ATYPE NUM, AFORM NONE, \\ ADV -, CONJ -}} \\ 
 \lexical{MD}  {\fs{N +, V -, BAR 2, MINOR NONE, PLU -, POSS -, NTYPE NUM,\\  WH -, CONJ -}} \\ 
\lexical{MF}  {\fs{N +, V -, BAR 2, MINOR NONE, POSS -, NTYPE NUM, \\ WH -, CONJ -}} \\ 
\lexical{NC2}  {\fs{N +, V -, BAR 0, MINOR NONE, PLU +, \\ POSS -, NTYPE NORM,  WH -, CONJ -}} \\ 
\lexical{ND1}  {\fs{N +, V -, BAR 0, MINOR NONE, PLU -, \\ POSS -, NTYPE DIR, WH -, CONJ -}} \\ 
\lexical{ND1}   {\fs{N +, V -, BAR 2, MINOR NONE, PLU -, \\ POSS -, NTYPE DIR,  WH -, CONJ -}} \\ 
\lexical{NN}  {\fs{N +, V -, BAR 0, MINOR NONE, \\ POSS -, NTYPE NORM,  WH -, \\ CONJ -}} \\ 
\lexical{NN}   {\fs{N +, V -, BAR 2, MINOR NONE, \\ POSS -, NTYPE NORM,  WH -, \\ CONJ -}} \\ 
\lexical{NN1}  {\fs{N +, V -, BAR 0, MINOR NONE, PLU -, \\ POSS -, NTYPE NORM, \\ WH -, CONJ -}} \\ 
\lexical{NN1\$}  {\fs{N +, V -, BAR 0, MINOR NONE, PLU +, \\ POSS +, NTYPE NORM,  WH -, CONJ -}} \\ 
\lexical{NN2}  {\fs{N +, V -, BAR 0, MINOR NONE, PLU +, \\ POSS -, NTYPE NORM,  WH -, CONJ -}} \\ 
\lexical{NN2}   {\fs{N +, V -, BAR 2, MINOR NONE, PLU +, \\ POSS -, NTYPE NORM,  WH -, CONJ -}} \\ 
\lexical{NNJ}  {\fs{N +, V -, BAR 0, MINOR NONE, \\ POSS -, NTYPE NORM, \\ WH -,  CONJ -}} \\ 
\lexical{NNJ1}  {\fs{N +, V -, BAR 0, MINOR NONE, PLU -, \\ POSS -, NTYPE NORM,  WH -, CONJ -}} \\ 
\lexical{NNJ2}  
   {\fs{N +, V -, BAR 0, MINOR NONE, PLU +, \\ POSS -, NTYPE NORM,  WH -, CONJ -}} \\ 
 \lexical{NNJ2}  {\fs{N +, V -, BAR 2, MINOR NONE, PLU +, \\ POSS -, NTYPE NORM,  WH -, CONJ -}} \\ 
\lexical{NNL}  {\fs{N +, V -, BAR 0, MINOR NONE, \\ POSS -, NTYPE NORM,  WH -, CONJ -}} \\ 
\lexical{NNL1}  {\fs{N +, V -, BAR 0, MINOR NONE, PLU -, \\ POSS -, NTYPE NORM,  WH -, CONJ -}} \\ 
\lexical{NNL2} 
   {\fs{N +, V -, BAR 0, MINOR NONE, PLU +, \\ POSS -, NTYPE NORM,  WH -, CONJ -}} \\ 
\lexical{NNL2}   {\fs{N +, V -, BAR 2, MINOR NONE, PLU +, \\ POSS -, NTYPE NORM,  WH -, CONJ -}} \\ 
\lexical{NNO}  {\fs{N +, V -, BAR 0, MINOR NONE, \\ POSS -, NTYPE NORM,  WH -, CONJ -}} \\ 
\lexical{NNO1}  {\fs{N +, V -, BAR 0, MINOR NONE, PLU -, \\ POSS -,  NTYPE NORM,  WH -, CONJ -}} \\ 
\lexical{NNO2}  
   {\fs{N +, V -, BAR 0, MINOR NONE, PLU +, \\ POSS -, NTYPE NORM,  WH -, CONJ -}} \\ 
 \lexical{NN02}  {\fs{N +, V -, BAR 2, MINOR NONE, PLU +, \\ POSS -, NTYPE NORM,  WH -, CONJ -}} \\ 
\lexical{NNS}  {\fs{N +, V -, BAR 0, MINOR NONE, \\ POSS -, NTYPE NORM,  WH -, CONJ -}} \\ 
\lexical{NNS1}  {\fs{N +, V -, BAR 0, MINOR NONE, PLU -, \\ POSS -, NTYPE TIT,  WH -, CONJ -}} \\ 
\lexical{NNS2}  {\fs{N +, V -, BAR 0, MINOR NONE, PLU +, \\ POSS -, NTYPE TIT,  WH -, CONJ -}} \\ 
\lexical{NNS2}   {\fs{N +, V -, BAR 2, MINOR NONE, PLU +, \\ POSS -, NTYPE TIT,  WH -, CONJ -}} \\ 
\lexical{NNSA1}  
   {\fs{N +, V -, BAR 0, MINOR NONE, PLU -, \\ POSS -, NTYPE POSTTIT,  WH -, CONJ -}} \\ 
\lexical{NNSA2} 
   {\fs{N +, V -, BAR 0, MINOR NONE, PLU +, \\ POSS -, NTYPE POSTTIT,  WH -, CONJ -}} \\ 
\lexical{NNSB} 
   {\fs{N +, V -, BAR 0, MINOR NONE, \\ POSS -, NTYPE PRETIT,  WH -, CONJ -}} \\ 
\lexical{NNSB1} 
   {\fs{N +, V -, BAR 0, MINOR NONE, PLU -, \\ POSS -, NTYPE PRETIT,  WH -, CONJ -}} \\ 
\lexical{NNSB2} 
   {\fs{N +, V -, BAR 0, MINOR NONE, PLU +, \\ POSS -, NTYPE PRETIT,  WH -, CONJ -}} \\ 
\lexical{NNSB2}   {\fs{N +, V -, BAR 2, MINOR NONE, PLU +, \\ POSS -, NTYPE PRETIT,  WH -, CONJ -}} \\ 
\lexical{NNT}  {\fs{N +, V -, BAR 0, MINOR NONE, \\ POSS -, NTYPE TEMP,  WH -, CONJ -}} \\ 
\lexical{NNT1}  {\fs{N +, V -, BAR 0, MINOR NONE, PLU -, \\ POSS -, NTYPE TEMP,  WH -, CONJ -}} \\ 
\lexical{NNT2}  
   {\fs{N +, V -, BAR 0, MINOR NONE, PLU +, \\ POSS -, NTYPE TEMP,  WH -, CONJ -}} \\ 
\lexical{NNT2}   {\fs{N +, V -, BAR 2, MINOR NONE, PLU +, \\ POSS -, NTYPE TEMP,  WH -, CONJ -}} \\ 
\lexical{NNU}  {\fs{N +, V -, BAR 0, MINOR NONE, \\ POSS -, NTYPE MEAS,  WH -, CONJ -}} \\ 
\lexical{NNU1}  {\fs{N +, V -, BAR 0, MINOR NONE, PLU -, \\ POSS -, NTYPE MEAS,  WH -, CONJ -}} \\ 
\lexical{NNU2}  
   {\fs{N +, V -, BAR 0, MINOR NONE, PLU +, \\ POSS -, NTYPE MEAS,  WH -, CONJ -}} \\ 
 \lexical{NNU2}  {\fs{N +, V -, BAR 2, MINOR NONE, PLU +, \\ POSS -, NTYPE MEAS,  WH -, CONJ -}} \\ 
\lexical{NP}  {\fs{N +, V -, BAR 0, MINOR NONE, \\ POSS -, NTYPE NAME,  WH -, CONJ -}} \\ 
\lexical{NP1}  {\fs{N +, V -, BAR 0, MINOR NONE, PLU -, \\ POSS -, NTYPE NAME,  WH -, CONJ -}} \\ 
\lexical{NP1}   {\fs{N +, V -, BAR 2, MINOR NONE, PLU -, \\ POSS -, NTYPE NAME,  WH -, CONJ -}} \\ 
\lexical{NP2}  {\fs{N +, V -, BAR 0, MINOR NONE, PLU +, \\ POSS -, NTYPE NAME,  WH -, CONJ -}} \\ 
\lexical{NPD1}  
   {\fs{N +, V -, BAR 0, MINOR NONE, PLU -, \\ POSS -, NTYPE TEMP,  WH -, CONJ -}} \\ 
\lexical{NPD1}   {\fs{N +, V -, BAR 2, MINOR NONE, PLU -, \\ POSS -, NTYPE TEMP,  WH -, CONJ -}} \\ 
\lexical{NPD2}  
   {\fs{N +, V -, BAR 0, MINOR NONE, PLU +, \\ POSS -, NTYPE TEMP,  WH -, CONJ -}} \\ 
\lexical{NPD2}   {\fs{N +, V -, BAR 2, MINOR NONE, PLU +, \\ POSS -, NTYPE MEAS,  WH -, CONJ -}} \\ 
\lexical{NPM1}  
   {\fs{N +, V -, BAR 0, MINOR NONE, PLU -, \\ POSS -, NTYPE TEMP,  WH -, CONJ -}} \\ 
\lexical{NPM1}   {\fs{N +, V -, BAR 2, MINOR NONE, PLU -, \\ POSS -, NTYPE MEAS,  WH -, CONJ -}} \\ 
\lexical{NPM2}  
   {\fs{N +, V -, BAR 0, MINOR NONE, PLU +, \\ POSS -, NTYPE TEMP,  WH -, CONJ -}} \\ 
\lexical{NPM2}   {\fs{N +, V -, BAR 2, MINOR NONE, PLU +, \\ POSS -, NTYPE MEAS,  WH -, CONJ -}} \\ 
\lexical{PN}  {\fs{N +, V -, BAR 2, MINOR NONE, \\ POSS -, NTYPE PRO,  WH -, CONJ -}} \\ 
\lexical{PN1}  {\fs{N +, V -, BAR 2, MINOR NONE, PLU -, \\ POSS -, NTYPE PRO,  WH -, CONJ -}} \\ 
\lexical{PNQO}  {\fs{N +, V -, BAR 2, MINOR NONE, PLU -, \\ POSS -, NTYPE PRO,  WH +, CONJ -}} \\ 
\lexical{PNQS}  {\fs{N +, V -, BAR 2, MINOR NONE, PLU -, \\ POSS -, NTYPE PRO,  WH +, CONJ -}} \\ 
\lexical{PNQV\$}  {\fs{N +, V -, BAR 2, MINOR NONE, PLU -, \\ POSS -, NTYPE PRO,  WH +, CONJ -}} \\ 
\lexical{PNQVO}  {\fs{N +, V -, BAR 2, MINOR NONE, PLU -, \\ POSS -, NTYPE PRO,  WH +, CONJ -}} \\ 
\lexical{PNQVS}  {\fs{N +, V -, BAR 2, MINOR NONE, PLU -, \\ POSS -, NTYPE PRO,  WH +, CONJ -}} \\ 
\lexical{PNX1}  {\fs{N +, V -, BAR 2, MINOR NONE, PLU -, \\ POSS -, NTYPE PRO,  WH -, CONJ -}} \\ 
\lexical{PP\$}  {\fs{N +, V -, BAR 2, MINOR NONE, PLU -, \\ POSS +, NTYPE PRO,  WH -, CONJ -}} \\ 
\lexical{PPH1}  {\fs{N +, V -, BAR 2, MINOR NONE, PLU -, \\ POSS -, NTYPE PRO,  WH -, CONJ -}} \\ 
\lexical{PPHO1}  {\fs{N +, V -, BAR 2, MINOR NONE, PLU -, \\ POSS -, NTYPE PRO,  WH -, CONJ -}} \\ 
\lexical{PPHO2}  {\fs{N +, V -, BAR 2, MINOR NONE, PLU +, \\ POSS -, NTYPE PRO,  WH -, CONJ -}} \\ 
\lexical{PPHS1}  
   {\fs{N +, V -, BAR 2, MINOR NONE, PLU -, \\ POSS -, \\ NTYPE PRONOM, WH -, CONJ -}} \\ 
\lexical{PPHS2}  
   {\fs{N +, V -, BAR 2, MINOR NONE, PLU +, \\ POSS -, \\ NTYPE PRONOM,   WH -, CONJ -}} \\ 
\lexical{PPIO1}  {\fs{N +, V -, BAR 2, MINOR NONE, PLU -, \\ POSS -, NTYPE PRO,  WH -, CONJ -}} \\ 
\lexical{PPIO2}  {\fs{N +, V -, BAR 2, MINOR NONE, PLU +, \\ POSS -, NTYPE PRO,  WH -, CONJ -}} \\ 
\lexical{PPIS1}  
   {\fs{N +, V -, BAR 2, MINOR NONE, PLU -, \\ POSS -, NTYPE PRONOM,  WH -, CONJ -}} \\ 
\lexical{PPIS2}  
   {\fs{N +, V -, BAR 2, MINOR NONE, PLU +, \\ POSS -, \\ NTYPE PRONOM,  WH -, CONJ -}} \\ 
\lexical{PPX1}  {\fs{N +, V -, BAR 2, MINOR NONE, PLU -, \\ POSS -, NTYPE PRO,  WH -, CONJ -}} \\ 
\lexical{PPX2}  {\fs{N +, V -, BAR 2, MINOR NONE, PLU +, \\ POSS -, NTYPE PRO,  WH -, CONJ -}} \\ 
\lexical{PPY} 
   {\fs{N +, V -, BAR 2, MINOR NONE, PLU -, \\ POSS -, NTYPE PRONOM, WH -, CONJ -}} \\ 
\lexical{RA}  {\fs{N +, V +, BAR 1, MINOR NONE, ATYPE POST, AFORM NONE, \\ ADV +, CONJ -}} \\ 
\lexical{REX}  {\fs{N +, V +, BAR 0, MINOR NONE, ATYPE XCOMP, AFORM NONE, \\ ADV +, CONJ -}} \\ 
\lexical{RG}  {\fs{MINOR DEG}} \\ 
\lexical{RGA}  {\fs{N +, V +, BAR 1, MINOR NONE, ATYPE POST, AFORM NONE, \\ ADV +, CONJ -}} \\ 
\lexical{RGQ}  {\fs{N +, V +, BAR 1, MINOR NONE, ATYPE HOW, AFORM NONE, \\ ADV +, CONJ -}} \\ 
\lexical{RGQV}  {\fs{N +, V +, BAR 1, MINOR NONE, ATYPE HOW, AFORM NONE, \\ ADV +, CONJ -}} \\ 
\lexical{RGR}  {\fs{N +, V +, BAR 0, MINOR NONE, AFORM ER, \\ ADV +, CONJ -}} \\ 
 \lexical{RGR}  {\fs{N +, V +, BAR 1, MINOR NONE, AFORM ER,\\  ADV +, CONJ -}} \\
\lexical{RGT}  {\fs{N +, V +, BAR 0, MINOR NONE, AFORM EST, \\ ADV +, CONJ -}} \\ 
 \lexical{RGT}  {\fs{N +, V +, BAR 1, MINOR NONE, AFORM EST, \\ ADV +, CONJ -}} \\ 
\lexical{RL}  {\fs{N +, V +, BAR 1, MINOR NONE, AFORM NONE, \\ ADV +, CONJ -}} \\ 
\lexical{RP}  {\fs{MINOR PRT}} \\ 
\lexical{RPK}  {\fs{N +, V +, BAR 0, MINOR NONE, ATYPE CAT, AFORM NONE, \\ ADV +, CONJ -}} \\ 
\lexical{RR}  {\fs{N +, V +, BAR 0, MINOR NONE, AFORM NONE, \\ ADV +, CONJ -}} \\
\lexical{RR}   {\fs{N +, V +, BAR 1, MINOR NONE, AFORM NONE, \\ ADV +, CONJ -}} \\ 
\lexical{RRQ} {\fs{N -, V -, BAR 1, MINOR NONE, SUBCAT NONE, \\ PFORM WH, CONJ -}} \\ 
\lexical{RRQV}  {\fs{N -, V -, BAR 1, MINOR NONE, SUBCAT NONE, \\ PFORM WH, CONJ -}} \\ 
\lexical{RRR}  {\fs{N +, V +, BAR 0, MINOR NONE, AFORM ER, ADV +, CONJ -}} \\ 
 \lexical{RRR}  {\fs{N +, V +, BAR 1, MINOR NONE, AFORM ER, ADV +, CONJ -}} \\ 
\lexical{RRT}  {\fs{N +, V +, BAR 0, MINOR NONE, AFORM EST, ADV +, CONJ -}} \\ 
 \lexical{RRT}  {\fs{N +, V +, BAR 0, MINOR NONE, AFORM EST, ADV +, CONJ -}} \\ 
 \lexical{RT}  {\fs{N +, V -, BAR 2, MINOR NONE, PLU -, \\ POSS -, NTYPE TEMP,  WH -, CONJ -}} \\ 
 \lexical{TO}  {\fs{N -, V +, BAR 0, MINOR NONE, AUX TO, VFORM INF, CONJ -}} \\ 
 \lexical{UH}  {\fs{MINOR INTERJ}} \\ 
 \lexical{VB0}  {\fs{N -, V +, BAR 0, MINOR NONE, AUX BE, VFORM BSE, CONJ -}} \\ 
 \lexical{VBDR}  {\fs{N -, V +, BAR 0, MINOR NONE, AUX BE, VFORM PAST, CONJ -}} \\ 
 \lexical{VBDZ}  {\fs{N -, V +, BAR 0, MINOR NONE, \\ PLU -, AUX BE, VFORM PAST, CONJ -}} \\ 
 \lexical{VBG}  {\fs{N -, V +, BAR 0, MINOR NONE, \\ AUX BE, VFORM ING, CONJ -}} \\ 
 \lexical{VBM}  {\fs{N -, V +, BAR 0, MINOR NONE, \\ PLU -, AUX BE, VFORM PRES, CONJ -}} \\ 
 \lexical{VBN}  {\fs{N -, V +, BAR 0, MINOR NONE, \\ AUX BE, VFORM PPART, CONJ -}} \\ 
 \lexical{VBR}  {\fs{N -, V +, BAR 0, MINOR NONE, \\ PLU +, AUX BE, VFORM PRES, CONJ -}} \\ 
 \lexical{VBZ}  {\fs{N -, V +, BAR 0, MINOR NONE, \\ PLU -, AUX BE, VFORM PRES, CONJ -}} \\ 
 \lexical{VD0}  {\fs{N -, V +, BAR 0, MINOR NONE, \\ AUX DO, VFORM BSE, CONJ -}} \\ 
 \lexical{VDD}  {\fs{N -, V +, BAR 0, MINOR NONE, \\ AUX DO, VFORM PAST, CONJ -}} \\ 
 \lexical{VDG}  {\fs{N -, V +, BAR 0, MINOR NONE, \\ AUX DO, VFORM ING, CONJ -}} \\ 
 \lexical{VDN}  {\fs{N -, V +, BAR 0, MINOR NONE, \\ AUX DO, VFORM PPART, CONJ -}} \\ 
 \lexical{VDZ}  {\fs{N -, V +, BAR 0, MINOR NONE, \\ PLU -, AUX DO, VFORM PRES, CONJ -}} \\ 
 \lexical{VH0}  {\fs{N -, V +, BAR 0, MINOR NONE, \\ AUX HAVE, VFORM BSE, CONJ -}} \\ 
 \lexical{VHD}  {\fs{N -, V +, BAR 0, MINOR NONE, \\ AUX HAVE, VFORM PAST, CONJ -}} \\ 
 \lexical{VHG}  {\fs{N -, V +, BAR 0, MINOR NONE, \\ AUX HAVE, VFORM ING, CONJ -}} \\ 
 \lexical{VHN}  {\fs{N -, V +, BAR 0, MINOR NONE, \\ AUX HAVE, VFORM PPART, CONJ -}} \\ 
 \lexical{VHZ}  {\fs{N -, V +, BAR 0, MINOR NONE, \\ PLU -, AUX HAVE, VFORM PRES, CONJ -}} \\ 
 \lexical{VM}  {\fs{N -, V +, BAR 0, MINOR NONE, \\ AUX MODAL,  CONJ -}} \\ 
 \lexical{VMK}  {\fs{N -, V +, BAR 0, MINOR NONE, \\ AUX CAT,  CONJ -}} \\ 
 \lexical{VV0}  {\fs{N -, V +, BAR 0, MINOR NONE, \\ VFORM BSE, CONJ -}} \\ 
 \lexical{VVD}  {\fs{N -, V +, BAR 0, MINOR NONE, \\ VFORM PAST, CONJ -}} \\ 
 \lexical{VVG}  {\fs{N -, V +, BAR 0, MINOR NONE, \\VFORM ING, CONJ -}} \\ 
 \lexical{VVGK}  {\fs{N -, V +, BAR 0, MINOR NONE, \\VFORM ING, CONJ -}} \\ 
 \lexical{VVN}  {\fs{N -, V +, BAR 0, MINOR NONE, \\VFORM PPART, CONJ -}} \\ 
 \lexical{VVNK}  {\fs{N -, V +, BAR 0, MINOR NONE, \\VFORM PPART, CONJ -}} \\ 
 \lexical{VVZ}  {\fs{N -, V +, BAR 0, MINOR NONE, \\ PLU -, VFORM PRES, CONJ -}} \\ 
 \lexical{XX}  {\fs{MINOR NOT}} \\ 
 \lexical{ZZ1}  {\fs{N +, V -, BAR 0, MINOR NONE, \\ PLU -, \\ POSS -, NTYPE NAME,  WH -, CONJ -}} \\ 
  \lexical{ZZ1}  {\fs{N +, V -, BAR 2, MINOR NONE,  PLU -, \\ POSS -, NTYPE NAME,  WH -, CONJ -}} \\ 
 \lexical{ZZ2}  {\fs{N +, V -, BAR 0, MINOR NONE,  PLU +, \\ POSS -, NTYPE NAME,  WH -, CONJ -}} \\ 
 \lexical{ZZ2}  {\fs{N +, V -, BAR 2, MINOR NONE,  PLU +, \\ POSS -, NTYPE NAME,  WH -, CONJ -}} \\

\chapter{The Original Grammar \label{previous}}
Here we present, in a paraphrased form, grammar $G1$.  The symbols used
are notational abbreviations for feature-structures and are of the form
{\it Xn}, where {\it X} is a phrase, and {\it n} is the bar level.  Minor
categories are represented by idiosyncratic symbols.  The correspondences
between symbols and feature-structures are as follows:
\begin{center}
\begin{tabular}{|l|l|} \hline 
Symbol & Feature-structure \\ \hline 
Nn     & \fs{N +, V -, BAR n, MINOR NONE } \\
Vn     & \fs{N -, V +, BAR n, MINOR NONE} \\
An     & \fs{N +, V +, BAR n, MINOR NONE } \\
Pn     & \fs{N -, V -, BAR n, MINOR NONE } \\
INTERJ & \fs{MINOR INTERJ} \\
DEG & \fs{MINOR DEG} \\
NOT  & \fs{MINOR NOT} \\
CONJ & \fs{MINOR CONJ} \\
DT & \fs{MINOR DET} \\ \hline 
\end{tabular}
\end{center}
Note that the grammars are paraphrased: the grammars in feature-structure
form contain categories with  $18$ features.  This 
means that some of the
paraphrased rules contain apparent redundancies (such as \psr{\{N1,~N1\}}{N1~N1}).  In
reality, such rules, in full,  do have distinct categories, but these distinctions
are not preserved when paraphrased.  If interested, the reader should
contact the author (miles@minster.york.ac.uk) for full details.

Grammar $G1$ is as follows:
\begin{center}
$A1$ $ \rightarrow $ $A1$ ~ $A1$ ~ \\  
$A1$ $ \rightarrow $ $DEG$ ~ $A1$ \\  
$A0$ $ \rightarrow $ $A0$ ~ $A0$ \\  
$A1$ $ \rightarrow $ $A0$ ~ $V1$ \\  
$A1$ $ \rightarrow $ $A0$ ~ $V2$ \\  
$A1$ $ \rightarrow $ $A0$ ~ $P1$ \\  
$A1$ $ \rightarrow $ $A0$ \\  
$P0$ $ \rightarrow $ $P0$ ~ $P0$ \\  
$P2$ $ \rightarrow $ $A2$ ~ $P1$ \\  
$P1$ $ \rightarrow $ $P0$ \\  
$P1$ $ \rightarrow $ $P0$ ~ $V1$ \\  
$P1$ $ \rightarrow $ $P0$ ~ $V1$ \\  
$P1$ $ \rightarrow $ $P0$ ~ $V2$ \\  
$P1$ $ \rightarrow $ $P0$ ~ $V2$ \\  
$P1$ $ \rightarrow $ $P0$ ~ $N2$ \\  
$N1$ $ \rightarrow $ $A2$ ~ $N1$ \\  
$N1$ $ \rightarrow $ $A1$ ~ $N1$ \\  
$N1$ $ \rightarrow $ $N1$ ~ $N1$ \\  
$N1$ $ \rightarrow $ $N1$ ~ $P1$ \\  
$N0$ $ \rightarrow $ $N0$ ~ $N0$ \\  
$N0$ $ \rightarrow $ $N0$ ~ $N0$ \\  
$N1$ $ \rightarrow $ $N0$ ~ $P1$ \\  
$N1$ $ \rightarrow $ $N0$ ~ $V1$ \\  
$N1$ $ \rightarrow $ $N0$ ~ $V2$ \\  
$N1$ $ \rightarrow $ $N0$ \\  
$N2$ $ \rightarrow $ $N2$ ~ $N1$ \\  
$N2$ $ \rightarrow $ $DT$ ~ $N1$ \\  
$N2$ $ \rightarrow $ $DT$ ~ $N1$ \\  
$N2$ $ \rightarrow $ $N1$ \\  
$V1$ $ \rightarrow $ $NOT$ ~ $V1$ \\  
$V1$ $ \rightarrow $ $V0$ ~ $V1$ \\  
$V1$ $ \rightarrow $ $V0$ ~ $V1$ \\  
$V1$ $ \rightarrow $ $V0$ ~ $V1$ \\  
$V1$ $ \rightarrow $ $V0$ ~ $V1$ \\  
$V1$ $ \rightarrow $ $V0$ ~ $N2$ \\  
$V1$ $ \rightarrow $ $V0$ ~ $A2$ \\  
$V1$ $ \rightarrow $ $V0$ ~ $A1$ \\  
$V1$ $ \rightarrow $ $V0$ ~ $P1$ \\  
$V1$ $ \rightarrow $ $V0$ ~ $P2$ \\  
$V1$ $ \rightarrow $ $V0$ ~ $V1$ \\  
$V1$ $ \rightarrow $ $V0$ ~ $V1$ \\  
$V1$ $ \rightarrow $ $V1$ ~ $V1$ \\  
$V1$ $ \rightarrow $ $V1$ ~ $N2$ \\  
$V1$ $ \rightarrow $ $V1$ ~ $P2$ \\  
$V1$ $ \rightarrow $ $V1$ ~ $A2$ \\  
$V1$ $ \rightarrow $ $V1$ ~ $A1$ \\  
$V0$ $ \rightarrow $ $V0$ ~ $V0$ \\  
$V1$ $ \rightarrow $ $V0$ ~ $N2$ ~ $N2$ ~ $V2$ \\  
$V1$ $ \rightarrow $ $V0$ ~ $N2$ ~ $A2$ \\  
$V1$ $ \rightarrow $ $V0$ ~ $N2$ ~ $A1$ \\  
$V1$ $ \rightarrow $ $V0$ ~ $N2$ ~ $V2$ \\  
$V1$ $ \rightarrow $ $V0$ ~ $N2$ ~ $P1$ \\  
$V1$ $ \rightarrow $ $V0$ ~ $N2$ ~ $V1$ \\  
$V1$ $ \rightarrow $ $V0$ ~ $N2$ ~ $V1$ \\  
$V1$ $ \rightarrow $ $V0$ ~ $N2$ ~ $N2$ \\  
$V1$ $ \rightarrow $ $V0$ ~ $A2$ \\  
$V1$ $ \rightarrow $ $V0$ ~ $A1$ \\  
$V1$ $ \rightarrow $ $V0$ ~ $V2$ \\  
$V1$ $ \rightarrow $ $V0$ ~ $V1$ \\  
$V1$ $ \rightarrow $ $V0$ ~ $V1$ \\  
$V1$ $ \rightarrow $ $V0$ ~ $P1$ \\  
$V1$ $ \rightarrow $ $V0$ ~ $N2$ \\  
$V0$ $ \rightarrow $ $V0$ ~ $NOT$ \\  
$A0$ $ \rightarrow $ $CONJ$ ~ $A0$ \\  
$A1$ $ \rightarrow $ $CONJ$ ~ $A1$ \\  
$P0$ $ \rightarrow $ $CONJ$ ~ $P0$ \\  
$P1$ $ \rightarrow $ $CONJ$ ~ $P1$ \\  
$N0$ $ \rightarrow $ $CONJ$ ~ $N0$ \\  
$N1$ $ \rightarrow $ $CONJ$ ~ $N1$ \\  
$N2$ $ \rightarrow $ $CONJ$ ~ $N2$ \\  
$V0$ $ \rightarrow $ $CONJ$ ~ $V0$ \\  
$V1$ $ \rightarrow $ $CONJ$ ~ $V1$ \\  
$V2$ $ \rightarrow $ $CONJ$ ~ $V2$ \\  
$A1$ $ \rightarrow $ $CONJ$ ~ $A1$ \\  
$P1$ $ \rightarrow $ $CONJ$ ~ $P1$ \\  
$N2$ $ \rightarrow $ $CONJ$ ~ $N2$ \\  
$V1$ $ \rightarrow $ $CONJ$ ~ $V1$ \\  
$V2$ $ \rightarrow $ $CONJ$ ~ $V2$ \\  
$V2$ $ \rightarrow $ $COMP$ ~ $V2$ \\  
$A2$ $ \rightarrow $ $A1$ ~ $A1$ \\  
$P2$ $ \rightarrow $ $A1$ ~ $P1$ \\  
$V1$ $ \rightarrow $ $A1$ ~ $V1$ \\  
$V2$ $ \rightarrow $ $A1$ ~ $V2$ \\  
$V2$ $ \rightarrow $ $N2$ ~ $V2$ \\  
$P1$ $ \rightarrow $ $P1$ ~ $P1$ \\  
$V2$ $ \rightarrow $ $P1$ ~ $V2$ \\  
$V2$ $ \rightarrow $ $P1$ ~ $V2$ \\  
$V2$ $ \rightarrow $ $A1$ ~ $V2$ \\  
$V2$ $ \rightarrow $ $N2$ ~ $V2$ \\  
$V2$ $ \rightarrow $ $N2$ ~ $V2$ \\  
$V2$ $ \rightarrow $ $V0$ ~ $V2$ \\  
$N2$ $ \rightarrow $ $N2$ ~ $N2$ \\  
$N2$ $ \rightarrow $ $N2$ ~ $N0$ \\  
$N2$ $ \rightarrow $ $N2$ ~ $A1$ \\  
$V2$ $ \rightarrow $ $N2$ ~ $V1$ \\  
$V2$ $ \rightarrow $ $V2$ ~ $V2$ \\  
$S3$ $ \rightarrow $ $V2$ ~ 
\end{center}

\chapter{The Learnt Grammars \label{xxx}}
Here we present the grammars learnt in chapter seven.  These listings
use the same paraphrasing as in appendix \ref{previous} and only show
the rules learnt.  They do not repeat those rules in grammar $G1$.

Grammar $G2$ is:
\begin{center}
\{$V1$, $N3$, $N2$, $V0$\}$ \rightarrow $ $N2$~$V0$~ \\  
\{$N2$, $N1$, $N0$, $N1$\}$ \rightarrow $ $N0$~ $N1$~ \\  
\{$N2$, $P1$, $P0$, $N3$ $N1$, $N0$\}$ \rightarrow $ $P0$~ \{$N0$, $N1$, $N3$\}~ \\  
\{$P2$, $S3$, $P1$\}$ \rightarrow $ $S3$~ $P1$  \\  
\{$V1$, $P2$, $P1$, $V0$\}$ \rightarrow $ $P1$ ~ $V0$~ \\  
\{$N1$, $N3$, $N2$, $N0$\}$ \rightarrow $ $N2$ ~ $N0$ ~ \\  
\{$V1$, $V1$, $V0$, $V0$\}$ \rightarrow $ $V0$ ~ $V0$ ~ \\  
\{$V2$, $V2$, $N2$, $N1$, $V1$, $V1$, $V0$, $V0$\}$ \rightarrow $ $N1$~ \{$V0$, $V0$, $V1$, $V1$\}~ \\  
\{$N2$, $N1$, $N0$, $N1$\}$ \rightarrow $ $N0$~ $N1$ ~ \\  
\{$N2$, $V1$, $V0$, $N1$\}$ \rightarrow $ $V0$ ~ $N1$ ~ \\  
\{$V1$, $N3$, $N2$, $V0$\}$ \rightarrow $ $N2$ ~ $V0$ ~ \\  
\{$N2$, $V1$, $N3$, $N2$, $V0$, $N1$\}$ \rightarrow $ $N2$ ~ \{$N1$, $V0$\}~ \\  
\{$V2$, $V1$, $N3$, $N2$, $V0$\}$ \rightarrow $ $INTERJ$ ~ \{$V0$, $N2$, $N3$, $V1$\} ~ \\  
\{$V1$, $V0$\}$ \rightarrow $ $INTERJ$~ $V0$ ~ \\  
\{$A2$, $A2$, $A1$, $A1$\}$ \rightarrow $ $A1$ ~ $A1$ ~ \\  
\{$V1$, $V1$, $V0$, $V0$\}$ \rightarrow $ $V0$ ~ $V0$ ~ \\  
\{$V1$, $V1$, $N3$, $N2$, $V0$, $V0$\}$ \rightarrow $ $N2$ ~ \{$V0$, $V0$\}~ \\  
\{$A1$, $V1$, $V0$, $A0$\}$ \rightarrow $ $V0$ ~ $A0$ ~ \\  
\{$V1$, $N2$, $N1$, $V0$\}$ \rightarrow $ $N1$ ~ $V0$ ~ \\  
\{$N3$, $N1$, $N0$, $N2$\}$ \rightarrow $ $N0$~ $N2$ ~ \\  
\{$P1$, $N1$, $N0$, $P0$, $N3$\}$ \rightarrow $ $N0$~ \{$N3$, $P0$\}~ \\  
\{$V2$, $A2$, $N1$, $N0$, $A1$, $V1$, $V0$, $A0$\}$ \rightarrow $ $N0$ ~ \{$A0$, $V0$, $V1$, $A1$\}~ \\  
\{$N2$, $N1$, $N0$, $N1$\}$ \rightarrow $ $N0$ ~ $N1$ ~ \\  
\{$N2$, $N2$, $N3$, $N2$, $N1$, $N1$\}$ \rightarrow $ $N2$ ~ \{$N1$, $N1$\}~ \\  
\{$N2$, $N3$, $N2$, $N1$\}$ \rightarrow $ $N2$ ~ $N1$ ~ \\  
\{$V1$, $N3$, $N2$, $V0$\}$ \rightarrow $ $N2$ ~ $V0$ ~ \\  
\{$V1$, $V2$, $V1$, $V0$\}$ \rightarrow $ $V1$ ~ $V0$ ~ \\  
\{$S3$, $V2$, $V1$, $V2$, $V1$, $V0$\}$ \rightarrow $ $DT$ ~ \{$V0$, $V1$, $V2$, $V1$\}~ \\  
\{$V1$, $V0$\}$ \rightarrow $ $DT$ ~ $V0$ ~ \\  
\{$N3$, $N1$, $N0$, $N2$\}$ \rightarrow $ $N0$ ~ $N2$ ~ \\  
\{$N1$, $N3$, $N0$, $N2$\}$ \rightarrow $ $DT$ ~ \{$N2$, $N0$, $N3$\}~ \\  
\{$N3$, $N2$, $A1$, $A0$, $N1$, $N3$, $N0$, $N2$\}$ \rightarrow $ $A0$ ~ \{$N2$, $N0$, $N3$, $N1$\}~ 
\end{center}

Grammar $G3$ is:
\begin{center}
\{$N2$, $N1$, $N0$, $N1$\}$ \rightarrow $ $N0$ ~ $N1$ ~ \\  
\{$N2$, $P1$, $P0$, $N3$, $N1$, $N0$\}$ \rightarrow $ $P0$ ~ \{$N0$, $N1$, $N3$\}~ \\  
\{$A2$, $V1$, $V0$, $A1$\}$ \rightarrow $ $V0$ ~ $A1$ ~ \\  
\{$P2$, $S3$, $P1$\}$ \rightarrow $ $S3$ ~ $P1$ ~ \\  
\{$V1$, $P2$, $P1$, $V0$\}$ \rightarrow $ $P1$ ~ $V0$ ~ \\  
\{$N1$, $N3$, $N2$, $N0$\}$ \rightarrow $ $N2$ ~ $N0$ ~ \\  
\{$V1$, $V1$, $V0$, $V0$\}$ \rightarrow $ $V0$ ~ $V0$ ~ \\  
\{$V2$, $V2$, $N2$, $N1$, $V1$, $V1$, $V0$, $V0$\}$ \rightarrow $ $N1$ ~ \{$V0$, $V0$, $V1$, $V1$\}~ \\  
\{$N2$, $N1$, $N0$, $N1$\}$ \rightarrow $ $N0$ ~ $N1$ ~ \\  
\{$N2$, $V1$, $V0$, $N1$\}$ \rightarrow $ $V0$ ~ $N1$ ~ \\  
\{$V1$, $N3$, $N2$, $V0$\}$ \rightarrow $ $N2$ ~ $V0$ ~ \\  
\{$N2$, $V1$, $N3$, $N2$, $V0$, $N1$\}$ \rightarrow $ $N2$ ~ \{$N1$, $V0$\}~ \\  
\{$V2$, $V1$, $N3$, $N2$, $V0$\}$ \rightarrow $ $INTERJ$ ~ \{$V0$, $N2$, $N3$, $V1$\}~ \\  
\{$V1$, $V0$\}$ \rightarrow $ $INTERJ$ ~ $V0$ ~ \\  
\{$A2$, $A2$, $A1$, $A1$\}$ \rightarrow $ $A1$ ~ $A1$ ~ \\  
\{$V1$, $V1$, $V0$, $V0$\}$ \rightarrow $ $V0$ ~ $V0$ ~ \\  
\{$V1$, $V1$, $N3$, $N2$, $V0$, $V0$\}$ \rightarrow $ $N2$ ~ \{$V0$, $V0$\}~ \\  
\{$V1$, $N2$, $N1$, $V0$\}$ \rightarrow $ $N1$ ~ $V0$ ~ \\  
\{$N3$, $N1$, $N0$, $N2$\}$ \rightarrow $ $N0$ ~ $N2$ ~ \\  
\{$N1$, $N0$, $N3$\}$ \rightarrow $ $N0$ ~ $N3$ ~ \\  
\{$V2$, $N1$, $N0$, $V1$\}$ \rightarrow $ $N0$ ~ $V1$ ~ \\  
\{$N2$, $N1$, $N0$, $N1$\}$ \rightarrow $ $N0$ ~ $N1$ ~ \\  
\{$N2$, $N2$, $N3$, $N2$, $N1$, $N1$\}$ \rightarrow $ $N2$ ~ \{$N1$, $N1$\} ~ \\  
\{$N2$, $N3$, $N2$, $N1$\}$ \rightarrow $ $N2$ ~ $N1$ ~ \\  
\{$V1$, $A3$, $N3$, $N2$, $A2$, $V0$\}$ \rightarrow $ $N2$ ~ \{$V0$, $A2$\}~ \\  
\{$V1$, $N3$, $N2$, $V0$\}$ \rightarrow $ $N2$ ~ $V0$ ~ \\  
\{$V1$, $V2$, $V1$, $V0$\}$ \rightarrow $ $V1$ ~ $V0$ ~ \\  
\{$S3$, $V2$, $V1$, $V2$, $V1$, $V0$\}$ \rightarrow $ $DT$ ~ \{$V0$, $V1$, $V2$, $V1$\}~ \\  
\{$V1$, $V0$\}$ \rightarrow $ $DT$ ~ $V0$ ~ \\  
\{$N3$, $N3$, $N2$, $N2$\}$ \rightarrow $ $N2$ ~ $N2$ ~ \\  
\{$A1$, $A0$, $N3$, $N3$, $N2$, $N2$\}$ \rightarrow $ $A0$~ \{$N2$, $N2$, $N3$, $N3$\}~ 
\end{center}

Finally, grammar $G4$ is:
\begin{center}
\{$V1$, $N3$, $N2$, $V0$\}$ \rightarrow $ $N2$~$V0$~ \\  
\{$N2$, $N1$, $N0$, $N1$\}$ \rightarrow $ $N0$~ $N1$ ~ \\  
\{$N2$, $P1$, $P0$, $N3$, $N1$, $N0$\}$ \rightarrow $ $P0$ ~ \{$N0$, $N1$, $N3$\}~ \\  
\{$P2$, $S3$, $P1$\}$ \rightarrow $ $S3$ ~ $P1$~ \\  
\{$V1$, $P2$, $P1$, $V0$\}$ \rightarrow $ $P1$ ~ $V0$ ~ \\  
\{$N1$, $N3$, $N2$, $N0$\}$ \rightarrow $ $N2$ ~ $N0$ ~ \\  
\{$V1$, $V1$, $V0$, $V0$\}$ \rightarrow $ $V0$ ~ $V0$ ~ \\  
\{$V2$, $V2$, $N2$, $N1$, $V1$, $V1$, $V0$, $V0$\}$ \rightarrow $ $N1$ ~ \{$V0$, $V0$, $V1$, $V1$\}~ \\  
\{$N2$, $N1$, $N0$, $N1$\}$ \rightarrow $ $N0$ ~ $N1$ ~ \\  
\{$N2$, $V1$, $V0$, $N1$\}$ \rightarrow $ $V0$ ~ $N1$ ~ \\  
\{$V1$, $N3$, $N2$, $V0$\}$ \rightarrow $ $N2$ ~ \{$V0$\}~ \\  
\{$N2$, $V1$, $N3$, $N2$, $V0$, $N1$\}$ \rightarrow $ $N2$ ~ \{$N1$, $V0$\}~ \\  
\{$V2$, $V1$, $N3$, $N2$, $V0$\}$ \rightarrow $ $INTERJ$ ~ \{$V0$, $N2$, $N3$, $V1$\}~ \\  
\{$V1$, $V0$\}$ \rightarrow $ $INTERJ$ ~ $V0$ ~ \\  
\{$A2$, $A2$, $A1$, $A1$\}$ \rightarrow $ $A1$ ~ $A1$ ~ \\  
\{$V1$, $V1$, $V0$, $V0$\}$ \rightarrow $ $V0$ ~ $V0$ ~ \\  
\{$V1$, $V1$, $N3$, $N2$, $V0$, $V0$\}$ \rightarrow $ $N2$ ~ \{$V0$, $V0$\}~ \\  
\{$A1$, $V1$, $V0$, $A0$\}$ \rightarrow $ $V0$ ~ $A0$ ~ \\  
\{$V1$, $N2$, $N1$, $V0$\}$ \rightarrow $ $N1$ ~ $V0$ ~ \\  
\{$N3$, $N1$, $N0$, $N2$\}$ \rightarrow $ $N0$ ~ $N2$ ~ \\  
\{$N1$, $N0$, $N3$\}$ \rightarrow $ $N0$ ~ $N3$ ~ \\  
\{$V2$, $A2$, $N1$, $N0$, $A1$, $V1$, $V0$, $A0$\}$ \rightarrow $ $N0$ ~ \{$A0$, $V0$, $V1$, $A1$\}~ \\  
\{$N2$, $N1$, $N0$, $N1$\}$ \rightarrow $ $N0$ ~ $N1$ ~ \\  
\{$N2$, $N2$, $N3$, $N2$, $N1$, $N1$\}$ \rightarrow $ $N2$ ~ \{$N1$, $N1$\}~ \\  
\{$N2$, $N3$, $N2$, $N1$\}$ \rightarrow $ $N2$ ~ $N1$ ~ \\  
\{$V1$, $N3$, $N2$, $V0$\}$ \rightarrow $ $N2$ ~ $V0$ ~ \\  
\{$V1$, $V2$, $V1$, $V0$\}$ \rightarrow $ $V1$ ~ $V0$ ~ \\  
\{$S3$, $V2$, $V1$, $V2$, $V1$, $V0$\}$ \rightarrow $ $DT$~ \{$V0$, $V1$, $V2$, $V1$\}~ \\  
\{$V1$, $V0$\}$ \rightarrow $ $DT$ ~ \{$V0$\}~ \\  
\{$N3$, $N1$, $N0$, $N2$\}$ \rightarrow $ $N0$ ~ $N2$ ~ \\  
\{$N1$, $N3$, $N0$, $N2$\}$ \rightarrow $ $DT0$ ~ \{$N2$, $N0$, $N3$\}~ \\  
\{$N3$, $N2$, $A1$, $A0$, $N1$, $N3$, $N0$, $N2$\}$ \rightarrow $ $A0$ ~ \{$N2$, $N0$, $N3$, $N1$\}~ 
\end{center}

\chapter{The Corpus Material Used in Evaluation \label{qqq}}
In this appendix we present the tag sequences  used to train the grammars
({\it Train}), the tag sequences used to evaluate overgeneration ({\it Bad}),
the tag sequences  used to test for undergeneration ({\it Test}), and the
 parses for both 
{\it Yardstick} and {\it Plausible}.  For brevity, the actual SEC sentences
corresponding to the SEC tag sequences have been omitted.  The interested reader
should contact the author (miles@minster.york.uk.ac) for details of these sentences.

{\it Train} consisted of the tag sequences: \\
RR AT NN2 RT \\
II AT JJ NNJ \\
MC1 NN1 VBZ JJ \\
DD1 VBZ NP1 NP1 \\
DD1 VBZ NP1 NP1 \\
DDQ VDD PPY VV0 \\
PNQS VDD PPY VV0 \\
PPIS1 VH0 AT1 NN1 \\
AT NN1 VVZ RP \\
AT JJ NN1 \\
RT NN1 NN1 \\
NNJ NN1 NN1 \\
NP1 NP1 VVZ \\
NNJ NN1 NN1 \\
NN1 VVD JJ \\
CC APP\$ NN1 \\
PPHS2 VBDR VVN \\
RGQ RR RL \\
UH VV0 RP \\
PPY VH0 NN1 \\
UH PPIS1 VVD \\
PPIS2 VVD RP \\
RR PPHS1 VVD \\
DA1 NN1 RL \\
NNSB1 NP1 \\
NNSB1 NP1 \\
AT NN1 \\
PPHS1 VVD \\
JJ NN1 \\
RR RR  \\
AT NN1  \\
NP1 NNL2  \\
JJ NN1  \\
PN1 RA  \\
NP1 NP1  \\
UH UH  \\
VDD PPHS1  \\
VV0 RP  \\
NP1 VVD  \\
PPIS1 VV0  \\
RRQ XX  \\
CC NN1  \\
AT NN1  \\
PPIS1 VM  \\
VV0 PPY  \\
CC PPX1  \\
JJ NN1  \\
UH VDN  \\
PPH1 VBZ  \\
PPHS1 VVD  \\
NN1 NN2  \\
VV0 RL  \\
VM PPY  \\
PPHS1 VVD  \\
AT1 NNT1  \\
NP1 NP1  \\
NNSB1 NP1  \\
AT NN1  \\
PPHS1 VVD  

{\it Bad} consisted of the tag sequences: \\
LE UH MC1 VHN RRT DA1  \\
JK MD NNT2 DA2 ZZ2 NNL1   \\
CST CSA RRR MC-MC NNL2 RGQV   \\
NNU1 VVN MC-MC AT1 VVNK JJ   \\
DDQ NNU2 CCB VDG RGT NNT1   \\
NNJ2 VBN DAR CC PNQV\$ VB0   \\
RGR ZZ2 RRR CSA DB RGQV   \\
CSA PNQVO NNJ2 DAT XX VBDR   \\
UH PNQS PNQVS PPY DAR DA2R   \\
IW NNJ1 NNL2 TO JJR NPD1   \\
CCB NN2 NNS1 NNO1 NNS2 DD1  \\ 
RGA NNSB1 VHD PNQS JBR \$   \\
IF VM PNQS NP2 VDN RRQ   \\
PPHS2 DDQV VB0 VDN NNU1 VVZ   \\
ICS VHN NP1 MD JJ RGA   \\
AT NNSB2 MC-MC NNO1 VBN MC1   \\
VDN NNS DDQ IF \$ JJT   \\
PPX2 NNL1 \$ NNL UH NPD1   \\
JA BTO VD0 VBN VHN DDQ\$   \\
RRQV PNQVO NP2 PN1 NPD2 MF   \\
VH0 DA2 REX NNT1 NNS NNJ2   \\
NC2 XX VB0 PNQV\$ PNQV\$ MC   \\
NNL VDN NN1\$ NNS1 PNQV\$ DDQ\$   \\
II VBZ PPX1 NNJ1 VMK LE   \\
MD NNL DDQV MF VBN PPHO1   \\
DA1 PN II MC\$ VBM MC-MC   \\
CSA NNS1 RL VDD JJR UH   \\
VB0 VMK DA2R RL PPY JK   \\
PPHS1 VB0 NNL2 RG IW PPIS2   \\
NN1 VBDZ VBZ VHD PPHO2 VBDR   \\
NPM2 CSW JB DB RGQ APP\$   \\
DDQ\$ PNQS NPM2 DDQV DDQ\$ MC2   \\
VM RG VDZ MF VMK PN1   \\
MC-MC VVGK NPM2 MC\$ JA II   \\
NP1 JJT RGR VBG DB2 PN   \\
JJT RGQ RGQV NNO2 DB2 DA1   \\
PPX2 NNSB PPHO1 VM VDN PPY   \\
VBZ PNQO JJ MF BCS NNU1   \\
MC NNU VVNK PNQO NNO1 XX   \\
RA NNS VMK VHZ VVZ NPD2   \\
NNL2 PPH1 NNL PN1 DD NNT   \\
IW NNS2 APP\$ JBT MC-MC VDZ   \\
RR NNS2 JK VDG MC2 DAR   \\
PPHS1 NNU2 JK DB RGQV APP\$  \\ 
XX RP APP\$ NNJ1 NNU2 VD0   \\
CSA DAR PNQO RR NNU VBN   \\
LE PNQV\$ CS CSN MF NNS2   \\
NNT1 VVD PNQV\$ NN1\$ APP\$ NNL2  \\ 
VDN BCS MC1 MC2 RG NP2   \\
RGQV EX RGR VD0 PN1 \$  \\
VVZ RGQV MC RGA MC-MC DD1  \\ 
MC APP\$ JK NNU2 NNS2 VD0   \\
VHG NPD1 PPIS2 NNU NNO1 RT \\  
VHG IO NNSB NNO1 RT RRT  \\ 
CSW DD NN VHN RRR CCB  \\ 
VM ICS RGQ PNQVO \$ NNL   \\
VDG NN1\$ PP\$ NNO1 VVN NNO2  \\ 
MC1 PP\$ RR NNSB NP2 VH0 \\  
ZZ2 PNQV\$ VVNK NN1\$ EX CF  \\ 
DA1 UH PNQVS VDZ DAR XX   \\
XX NNSA1 VDD RPK VDG VVNK  \\ 
VBM DAR ND1 NNJ1 NPM2 NN2   \\
VDD VVZ TO CCB PN1 PNQVO  \\ 
VHG NNSA2 TO VM NNO1 RT  \\ 
VHD EX NNJ1 DAR DA1 DAR   \\
TO CSN VBN PPHS2 NN NNSB2   \\
JBR VBM PNQS VDD ND1 PPIS1  \\ 
NN1\$ CSW JBR IF LE VBM    \\ 
PPIS2 RGT VDD NNJ VVN NPD2  \\
PNQS NNSA2 DD2 NNT1 VM NNSA1 \\ 
DD2 PPH1 MF VBN PNQS PPX1  \\
NNT APP\$ JK PNQV\$ IF NNU2 \\ 
JK DA2R NNSA1 RRT VM PPHS1 \\ 
JBT PNX1 DAR DD PPHS2 VB0 \\ 
MC NNU1 CF DD1 DDQ CC  \\
ND1 NNS AT PNX1 RRT VD0  \\
CS \$ PNX1 AT1 DD BCS  \\
VBR VBM II CC NNSB PNQS  \\
NPM2 IW NNO1 VDN PPIS1 NNU1  \\
PPHS1 VHD MC-MC NNJ2 PNX1 PPX2  \\
VBG NNO1 MC PPIO2 PNX1 RL \\ 
NP PPIS1 VDN NPD2 JBT CSA  \\
RGA NNSB1 VH0 PPHO2 VDN JA \\ 
ICS NNU2 NNL MD PPY DDQ\$  \\
RGA VHZ NNSB2 PPIS2 EX DD2  \\
DD2 RR RR PNQO PPY PPIO2  \\
NPD2 DAT VVNK PPIS2 NC2 PPH1  \\
DA PPIS2 NP1 PNQO VDD NPM2  \\
RPK VBM PPIO2 PN1 NNJ2 RL  \\
DB2 VDG NNL2 VVGK VHZ CSW \\ 
VBG JK VBZ CS JBR VMK \\ 
PPHS1 VDG DB MC1 VD0 MC-MC  \\
NC2 APP\$ BTO JJT RRT NP \\ 
NNU1 JBT VVD NPM2 BTO PPY  \\
ICS RRQ CF NNJ2 NNU1 PPHS2  \\
\$ CSN VBDZ PN1 AT DD2  \\
IW NNJ2 NNO1 RL NNJ1 JB  \\
CCB VBZ NPD1 VMK DD2 JB  \\
NNSA1 VDD RA JJR JJT NNO1 \\ 
RA RL DD2 CST VDN VVN  

{\it Test} consisted of the tag sequences: \\
AT NN1 VVD II AT NNL1 IO NP1 NP1 II NP1 \\  
NNJ NN1 II MC RA II NPD1 AT MD IO NP1 \\
NNJ NN1 II MC RA II NPD1 AT MD IO NPM1 \\ 
CCB RRQ VDZ PPHS1 VV0 NN1 IW PP\$ RR JJ NNS2 \\ 
AT JJ JJ NN1 II NP1 NNL1 VVD RP NNT2 RA \\ 
CCB AT JJ NN1 VVZ VBZ PPH1 II VVG AT NN2 \\ 
AT JJ NN1 AT NP1 VBZ AT NN1 IO DA2 NN2 \\ 
II NP1 AT NN1 II NN1 CC NN1 VBZ RR JJ \\ 
AT NNS1 VVD II PPIO1 II JJ NN2 PPY VH0 NN1 \\ 
RRR RP AT1 NNL1 RRQ EX VBDZ AT1 NN1 VVG RP \\ 
PPIS2 VVD II AT NN1 CC VVD ICS JJ IF NP1 \\ 
AT NN2 AT NN1 VBDZ VVN RG RR VVD AT NNL1 \\ 
PPHS2 VBDR RR VVN IW JJ NN2 CC JJ JJ NN2 \\ 
CCB CS DD NN2 VV0 AT1 JJ NN1 VVZ AT DA \\ 
CF RR RL PPH1 VBZ CC II NP1 IO DB NN2 \\ 
PPHS1 VBZ VVN TO VV0 II NNL1 IF MC NNT2 \\ 
AT NN1 NN2 VH0 VVN AT1 JJ NNT1 II NPM1 \\ 
NNJ NN2 VH0 VVN AT NNJ IO VVG PPHO2 NN2 \\ 
EX VBZ RR NN1 IF NN2 NN2 II JJ NNT2 \\ 
AT1 NN1 VVD MD NNT1 II NP1 II NNL1 NP1 \\ 
AT NN1 VBDZ VVN RR MC NNU2 II AT NNL1 \\ 
PPHS1 VHD AT1 JJ NN1 NN1 NN1 RL II NPD1 \\ 
PPIS1 VH0 RR RR VVN DDQ PPHS1 VDD IW PPHO2 \\ 
RR JJ CST NP1 VBZ AT1 JJ NNL1 IF NN2 \\ 
PPH1 VVZ PPH1 VM VB0 RG JJ TO VV0 PPHO2 \\ 
DB JJ NN2 VH0 AT1 II AT NN1 NN1 NN1 \\ 
PPHS1 RT VVD RP TO VV0 AT1 JJ NNS1 RP  \\ 
MC CC MC NNT2 PPIS2 VVD JJ NN1 NNU2 RL \\ 
PPHS2 VBDR JJ CSA PPHS2 RR VBR IW AT NN \\ 
AT NN1 II DD1 NPM1 NNT1 II NP1 VBDZ JJ \\ 
DD1 VBZ AT1 NN1 IO RR II MC NNO II AT NPM1 NN1 \\ 
CC AT JJ NNJ IO NN2 VV0 AT1 NN1 II AT NN1 NNJ \\ 
PPH1 VBZ VVN CST DB NN2 VM VB0 RP II JJ RRR RT \\ 
EX VBZ AT NN1 II AT NP1 NN1 II NP1 CC NP1 NNL1 \\ 
APP\$ NN1 NP1 NP1 VVD II AT NN1 RR ICS AT NN1 VVD \\ 
IF DD1 PPIS2 VH0 RR TO VV0 II AT JJ NP1 NN1 NN1 \\ 
AT NN1 VBZ VVN AT NP1 CC NP1 NNL1 II APP\$ JJ NN1 \\ 
PPIS1 VH0 RR VVN RRQ PPY VV0 AT1 NN IW NN1 VVG NN2 \\ 
AT JJ JJ NN1 II NN1 NN1 VBZ AT NN1 IO AT NNT1 \\ 
NN1 CC NN1 RT II MC II NP1 VHZ VVN NN AT NNL1 \\ 
ICS NN1 NN1 NN1 VBZ AT MD RGT JJ NN1 IF JJ NN2 \\ 
CCB RL NN2 VVD AT1 JJ NN1 PPY VBR RL IF AT NNJ \\ 
 DDQ VBZ RRR PPHS2 VVD RR JJ JJ CC JJ VVN II NN1\\  
ICS DB PNQS VVZ NN1 NN1 CS PPY VM VV0 AT JJ NN1 \\  
RR MC1 IO AT NN2 CC NN2 PPIS1 VV0 IO VVG AT1 NN1 \\ 
II APP\$ NN1 PPHS1 VVD PPX1 RR II AT1 NN1 II NN1\\  
AT NN2 II AT NN1 VBDR RR VVN II NP1 JJ NNL1 \\ 
AT NP1 NN1 NN1 VVZ RP VVG JJ NN1 II NNS1 NP1 \\ 
PPH1 VBDZ APP\$ NN1 PPHS1 VVD TO VV0 NN2 II AT NN1 \\ 
AT NN1 VBZ CST PPHS2 VV0 CC APP\$ NN2 RR VV0 RL \\ 
CF RG RR CSA NN1 VBZ VVN NP1 VBZ AT1 JJ NN1 \\ 
PPHS1 RR VVZ TO VB0 VVG AT1 NN1 II AT NN1 NN2 \\ 
AT JJ NN1 VBDZ CST AT NN1 VBDZ JJ TO VV0 NN2 \\ 
PPH1 VBZ VVN II DD NN1 VVG NN2 ICS AT RGT JJ \\ 
PPHS2 VV0 JJ NN2 JJ NN2 CC AT NN1 IO AT NN1 \\ 
CSA PPIS2 VVD TO VV0 NNL1 AT1 JJ NNJ NN1 VVD RP \\ 
II NNT1 AT NP1 NN1 VBDZ RP MC II MC RR \\ 
CCB NN2 VVD RR ICS AT NNJ1 IO NP1 \$ NN1 \\ 
DD1 VBZ AT JJT NN1 IF DD NNT1 II MC MC \\ 
AT JJ NP1 NN1 II NNT1 VBDZ RP MC II MC \\

{\it Yardstick} consisted of the following parse trees:
\begin{flushleft}
\verb|(S (N AT JJ JJ NN1 |
\\ \verb|      (P II (N NP1 NNL1))) |
\\ \verb|   (V (R VVD RP) (NP NNT2 RA)))|
\\ \verb|(S (N AT JJ NN1 |
\\ \verb|      (N AT NP1)) |
\\ \verb|   (V VBZ (N AT NN1 |
\\ \verb|             (P IO (N DA2 NN2)))))|
\\ \verb|(S (P II (N NP1))|
\\ \verb|   (S (N AT NN1 |
\\ \verb|         (P II (N NN1 CC NN1))) |
\\ \verb|      (V VBZ (J RR JJ))))|
\\ \verb|(S (N PPHS2|
\\ \verb|      (V VBDR (J RR VVN) |
\\ \verb|         (P IW (N (N& JJ NN2) CC |
\\ \verb|                  (N+ JJ JJ NN2))))))|
\\ \verb|(S (N AT NN1 (P II |
\\ \verb|                (N DD1 NPM1 NNT1 |
\\ \verb|                   (P II (N NP1))))) |
\\ \verb|   (V VBDZ JJ))|
\\ \verb|(S (FA CCB |
\\ \verb|       (FA CS (N DD NN2) |
\\ \verb|           (V VV0)))|
\\ \verb|   (N AT1 JJ NN1 |
\\ \verb|      (V VVZ (N AT DA))))|
\\ \verb|(S (J RR JJ (FN CST |
\\ \verb|                (N NP1) |
\\ \verb|                (V VBZ (N AT1 JJ NNL1 |
\\ \verb|                          (P IF NN2))))))|
\\ \verb|(S (N PPHS2) |
\\ \verb|   (V VBDR (J JJ))|
\\ \verb|   (FA CSA (N PPHS2) |
\\ \verb|       RR (V VBR (P IW (N AT NN)))))|
\\ \verb|(S CC (N AT JJ NNJ |
\\ \verb|         (P IO (N NN2)))|
\\ \verb|   (V VV0 (N AT1 NN1 |
\\ \verb|             (P II (N AT NN1 NNJ)))))|
\\ \verb|(S EX|
\\ \verb|   (V VBZ (N AT NN1) |
\\ \verb|      (P II (N AT NP1 NN1))|
\\ \verb|      (P II (N (N& NP1) |
\\ \verb|               CC (N+ NP1 NNL1)))))|
\\ \verb|(S (N AT JJ JJ NN1 |
\\ \verb|      (P II (N NN1 NN1)))|
\\ \verb|   (V VBZ (N AT NN1 |
\\ \verb|             (P IO (N AT NNT1)))))|
\\ \verb|(S (P II (N APP$ NN1))|
\\ \verb|   (S (N PPHS) (V VVD (N PPX1) RR |
\\ \verb|                  (P II (N AT1 NN1 (P II (N NN1)))))))|
\\ \verb|(S (N PPH1|
\\ \verb|      (V VBDZ (N APP$ NN1) (SI (N PPHS1) (V VVD))|
\\ \verb|         (TI TO VV0 (N NN2 (P II (N AT NN1)))))))|
\\ \verb|(S (N PPHS) RR|
\\ \verb|   (V VVZ (TI TO VB0 VVG |
\\ \verb|              (N AT1 NN1) (P II (N AT NN1 NN2)))))|
\\ \verb|(S (N AT JJ NN1)|
\\ \verb|   (V VBDZ (FN CST (N AT NN1) |
\\ \verb|               (V VBDZ (J JJ (TI TO VV0 (N NN2)))))))|
\end{flushleft}

Finally, {\it Plausible} consisted of the following parse trees:

\begin{flushleft}
\verb|(S (N AT NN1)|
\\ \verb|   (V VVD (P II (N AT NNL1 |
\\ \verb|                   (P IO (N NP1 NP1 |
\\ \verb|                            (P II (N NP1))))))))|
\\ \verb|(N NNJ NN1 (P II MC RA) |
\\ \verb|   (P II (N NPD1 (N AT MD |
\\ \verb|                    (P IO (N NP1))))))|
\\ \verb|(N (NNJ NN1 (P II MC RA) |
\\ \verb|        (P II (N NPD1 (N AT MD |
\\ \verb|                         (P IO (N NPM1)))))))|
\\ \verb|(S CCB|
\\ \verb|   (S (N AT JJ NN1) |
\\ \verb|      (V VVZ (V VBZ (N PPH) |
\\ \verb|                (P II (TG VVG |
\\ \verb|                          (N AT NN2)))))))|
\\ \verb|(S (N AT NNS1)|
\\ \verb|   (V VVD (P II (N PPIO1))|
\\ \verb|      (P II (N JJ NN2))|
\\ \verb|      (S (N PPY) (V VH0 (N NN1)))))|
\\ \verb|(S (N AT NN2 (S (N AT NN1) |
\\ \verb|                (V VBDZ VVN RG RR))) |
\\ \verb|   (V VVD (N AT NNL1)))|
\\ \verb|(S (S& CF (S (S RR (S RL (N PPH1) |
\\ \verb|                      (V VBZ)))))|
\\ \verb|   (S+ CC (P (P II (N NP1)) |
\\ \verb|             (P IO (N DB NN2)))))|
\\ \verb|(S (N PPHS1)|
\\ \verb|   (V VBZ VVN (TI TO VV0 |
\\ \verb|                  (P II (N NNL1)) |
\\ \verb|                  (P IF (N MC NNT2)))))|
\\ \verb|(S (N AT1 NN1) (V VVD (NR MD NNT1) |
\\ \verb|                  (P II (N NP1 (P II |
\\ \verb|                                  (N NNL1 NP1))))))|
\\ \verb|(S (N AT NN1) (V VBDZ VVN RR MC NNU2 |
\\ \verb|                 (P II (N AT NNL1))))|
\\ \verb|(S (N PPHS1) (V VHD (N AT1 JJ NN1 NN1 NN1) |
\\ \verb|                RL (P II (N NPD1))))|
\\ \verb|(S (N PPIS1)|
\\ \verb|   (V VH0 RR RR VVN (FN (N DDQ) |
\\ \verb|                        (N PPHS1) |
\\ \verb|                        (V VDD |
\\ \verb|                           (P IW (N PPHO2))))))|
\\ \verb|(S (N DB JJ NN2) (V VH0 |
\\ \verb|                    (N AT1 II |
\\ \verb|                       (N AT NN1 NN1) NN1)))|
\\ \verb|(S EX|
\\ \verb|   (V VBZ (N AT NN1) (P II (N AT NP1 NN1))|
\\ \verb|      (N (N& NP1) CC (N+ NP1 NNL1))))|
\\ \verb|(S (N AT NN1)|
\\ \verb|   (V VBZ VVN (N (N& AT NP1) |
\\ \verb|                 CC (N+ NP1 NNL1 |
\\ \verb|                        (P II |
\\ \verb|                           (N APP$ JJ NN1))))))|
\end{flushleft}


\end{document}